\PassOptionsToPackage{margin=1in, top = 1in, bottom = 0.8in, headheight = 0.3in}{geometry}

\documentclass[sn-basic]{sn-jnl}

\usepackage[utf8]{inputenc}
\usepackage{ifthen}
\usepackage{hyperref}

\PassOptionsToPackage{prologue,dvipsnames}{xcolor}
\usepackage{soul}
\usepackage[normalem]{ulem}
\usepackage{graphicx}

\usepackage{epsf,picinpar}
\usepackage{varioref}
\usepackage{fdsymbol}
\usepackage{harveyballs}

\usepackage{enumerate}
\usepackage[framemethod=TikZ]{mdframed}

\usepackage{pifont}
\usepackage[pscoord]{eso-pic}
\usepackage{listings}
\usepackage{courier}
\usepackage{subcaption}
\usepackage{multicol}
\usepackage{eqparbox,collcell}

\usepackage{booktabs}
\usepackage{multirow}
\usepackage{float}
\usepackage{array}
\usepackage{harveyballs}
\usepackage{makecell}
\usepackage{wrapfig}
\usepackage{svg}
\usepackage{orcidlink}

\usepackage{algorithm}
\usepackage{algpseudocode}

\usepackage{amsthm}
\newcommand{\secref}[1]{Sec.~\ref{#1}}
\newcommand{\figref}[1]{Fig.~\ref{#1}}
\newcommand{\tabref}[1]{Tab.~\ref{#1}}
\newcommand{\algoref}[1]{Algorithm~\ref{#1}}
\newcommand{\equref}[1]{Equation~\ref{#1}}
\newcommand{\lstref}[1]{Listing~\ref{#1}}
\newcommand{\appref}[1]{Appendix~\ref{#1}}
\newcommand{\theoref}[1]{Theorem~\ref{#1}}
\newcommand{\defref}[1]{Definition~\ref{#1}}

\newtheorem{definition}{Definition}%
\newtheorem{theorem}{Theorem}%

\newcommand{\todo}[1]{
    \ifthenelse{\boolean{showannotations}}%
    {\ifthenelse{\equal{#1}{}}{\textcolor{red}{TODO}}{\textcolor{red}{TODO:~{#1}}}}%
    {}%
}

\newcommand{\assignedto}[1]{%
    \ifthenelse{\boolean{showannotations}}%
    {\textbf{\noindent\ding{46}\textcolor{white}{~\colorbox{\assignementcolor}{Assigned to:}}~\textcolor{\assignementcolor}{#1}\\}%
    }
    {}
}

\renewcommand{\nb}[4]{
    \fcolorbox{gray}{#2}{\bfseries\sffamily\scriptsize{#1}}
	{\sf\small$\blacktriangleright$\textcolor{#4}{\textit{#3}}$\blacktriangleleft$}
}

\newcommand\personmarker[3]{\noindent\nb{#1}{yellow}{#2}{#3}}
\newcommand\id[1]{\noindent\personmarker{ID}{#1}{VioletRed}}
\newcommand\kd[1]{\noindent\personmarker{KD}{#1}{Orange}}

\newcommand\addblockbegin{%
    \ifthenelse{\boolean{showannotations}}%
    {\color{\newtextcolor}}%
    {}%
}

\newcommand\addblockend{%
    \ifthenelse{\boolean{showannotations}}%
    {\color{black}}%
    {}%
}

\newcommand\addblockbeginkyanna{%
    \ifthenelse{\boolean{showannotations}}%
    {\color{Mulberry}}%
    {}%
}

\newcommand\addblockendkyanna{%
    \ifthenelse{\boolean{showannotations}}%
    {\color{black}}%
    {}%
}

\newcommand{\newversion}[1]{%
    \ifthenelse{\boolean{showv2annotations}}%
    {\nb{v2}{yellow}{#1}{blue}}%
    {}%
}

\newenvironment{conclusionframe}[1]
  {\mdfsetup{
    frametitle={\colorbox{white}{\space#1\space}},
    innertopmargin=2pt,
    frametitleaboveskip=-\ht\strutbox,
    frametitlealignment=\center
    }
  \begin{mdframed}%
  }
  {\end{mdframed}}

\newcolumntype{P}[1]{>{\centering\arraybackslash}p{#1}}

\newlength{\dpcircle}
\newlength{\rcircle}
\newlength{\dcircle}
\newcommand{\docircle}[4]{%
  \setlength{\dpcircle}{\dp\strutbox}%
  \setlength{\dcircle}{\dpcircle}%
  \addtolength{\dcircle}{\ht\strutbox}%
  \setlength{\rcircle}{0.5\dcircle}%
  \setlength{\unitlength}{1sp}%
  \begin{picture}(\number\dcircle,0)
    \color{#1}
    \put(\number\rcircle,\number\dpcircle){\circle*{\number\dcircle}}
    \color{#2}
    \put(\number\rcircle,\number\dpcircle){\circle{\number\dcircle}}
    \put(\number\rcircle,0){\makebox[0pt]{\textcolor{#3}{#4}}}
  \end{picture}%
}

\definecolor{circleyellow}{rgb}{0.99, 0.95, 0.8}
\definecolor{examplecolor}{RGB}{66, 135, 245}

\newcommand{\examplebox}{%
    \fcolorbox{white}{examplecolor}{\textcolor{white}{\textbf{Example}}}%
}

\newcommand{\exampleheader}[1]{%
    \paragraph{{\normalfont \examplebox{}}~{#1}}
}

\newcommand{\colornumber}[1]{%
  {\scriptsize\docircle{circleyellow}{Tan}{black}{\textbf{#1}}}%
}

\newcommand\bcf[1]{{#1}^{\odot}}

\newcommand{\hlred}[1]{%
    {\sethlcolor{redhighlightcolor}\hl{#1}\textcolor{redarrowcolor}{$\downarrow$}}%
}

\newcommand{\hlgreen}[1]{%
    {\sethlcolor{greenhighlightcolor}\hl{#1}\textcolor{greenarrowcolor}{$\uparrow$}}%
}

\newcommand{\cell}[2]{%
    \texttt{[{#1},{#2}]}%
}

\newcommand{\parameter}[3]{%
    \texttt{\cell{#1}{#2}, {#3}}%
}

\newcommand{\advice}[3]{%
    \texttt{\cell{#1}{#2}$\rightarrow${#3}}%
}
\newboolean{showannotations}
\setboolean{showannotations}{true} 

\newboolean{showv2annotations}
\setboolean{showv2annotations}{true} 

\newcommand{\newtextcolor}{green!75!black}

\newcommand{\assignementcolor}{orange}
\definecolor{highlightcolor}{rgb}{.99, 1, .0}
\sethlcolor{highlightcolor}
\definecolor{redhighlightcolor}{HTML}{fcd2d4}
\definecolor{greenhighlightcolor}{HTML}{88fc92}
\definecolor{redarrowcolor}{HTML}{ff0000}
\definecolor{greenarrowcolor}{HTML}{018c0d}
\definecolor{orcidlogocol}{HTML}{A6CE39}

\mdfsetup{%
   backgroundcolor=gray!3,
   roundcorner=3pt%
}

\definecolor{darkgreen}{rgb}{0,0.6,0}
\definecolor{gray}{rgb}{0.5,0.5,0.5}
\definecolor{mauve}{rgb}{0.58,0,0.82}
\definecolor{gray}{rgb}{0.4,0.4,0.4}
\definecolor{darkblue}{rgb}{0.0,0.0,0.6}
\definecolor{lightblue}{rgb}{0.0,0.0,0.9}
\definecolor{cyan}{rgb}{0.0,0.6,0.6}
\definecolor{darkred}{rgb}{0.6,0.0,0.0}
\definecolor{armygreen}{rgb}{0.29, 0.33, 0.13}
\definecolor{antiquefuchsia}{rgb}{0.57, 0.36, 0.51}
\definecolor{coolblack}{rgb}{0.0, 0.18, 0.39}

\lstdefinelanguage{gridworld-dsl}
{
    comment=[l]{--},
    morekeywords={READ},
    keywordstyle=\color{white},
    morestring=[s][\color{darkred}\bfseries]{\{}{\}},
    morestring=[s][\color{darkred}\bfseries]{`}{`},
    morestring=[s][\color{mauve}\bfseries]{<}{>},
    extendedchars=true
}

\begin{document}

\title{Opinion-Guided Reinforcement Learning}

\author*[1]{\fnm{Kyanna} \sur{Dagenais}}\email{dagenaik@mcmaster.ca}

\author*[1]{\fnm{Istvan} \sur{David}}\email{istvan.david@mcmaster.ca}

\affil[1]{\orgdiv{Department of Computing and Software}, \orgname{McMaster University}, \orgaddress{\city{Hamilton}, \state{ON}, \country{Canada}}}

\date{Received: date / Accepted: date}

\abstract{
%
Human guidance is often desired in reinforcement learning to improve the performance of the learning agent.
%
However, human insights are often mere opinions and educated guesses rather than well-formulated arguments. While opinions are subject to uncertainty, e.g., due to partial informedness or ignorance about a problem, they also emerge earlier than hard evidence can be produced.
Thus, guiding reinforcement learning agents by way of opinions offers the potential for more performant learning processes, but comes with the challenge of modeling and managing opinions in a formal way.
In this article, we present a method to guide reinforcement learning agents through opinions. To this end, we provide an end-to-end method to model and manage advisors' opinions. To assess the utility of the approach, we evaluate it with synthetic (oracle) and human advisors, at different levels of uncertainty, and under multiple advice strategies. Our results indicate that opinions, even if uncertain, improve the performance of reinforcement learning agents, resulting in higher rewards, more efficient exploration, and a better reinforced policy.
Although we demonstrate our approach through a two-dimensional topological running example,
our approach is applicable to complex problems with higher dimensions as well.
}


\maketitle

\section{Introduction}\label{sec:introduction}

Reinforcement learning (RL) \citep{sutton2018reinforcement} is a machine learning paradigm in which an autonomous agent explores its surroundings and learns optimal behavior
through trial and error.
Because of the continuous learning process,
RL is particularly well-suited to deal with open-ended problems that feature unforeseen situations, tasks, and environments.
Thanks to these benefits, the popularity of RL has been steadily increasing in domains where dealing with complexity is essential, such as fine-tuning the control of manufacturing processes at runtime~\citep{cronrath2019enhancing}, inferring simulation components of digital twins~\citep{david2023automated}, and generating repair actions for conceptual domain models~\citep{barriga2022parmorel}.

However, autonomous behavior does not obviate the need for human agency~\citep{bradshaw2013seven}. Human input to autonomous systems often augments machine intelligence with higher-level strategic directives that require creative problem-solving skills~\citep{najar2021reinforcement}. RL is no exception to this rule either.
The body of knowledge on guided RL methods
is substantial and rapidly growing.

General guidance-based techniques allow humans to inform agents about future aspects of the task, e.g., interesting regions to explore~\citep{subramanian2016exploration} or a trajectory (i.e., a sequence of actions) to follow~\citep{thomaz2006reinforcement}. A severe limitation of such techniques is the inconsistent correctness of human advice. As \citet{scherf2022learning} report, the majority of interactive or guided RL approaches assume the human input to be useful and correct~\citep{li2019human}. However, this is not always true in real applications~\citep{kessler_faulkner2021interactive,koert2020multi,arakawa2018dqn}. Informed advisors, such as experts, often express advice in the form of \textit{opinions}
---cognitive constructs that are subject to \textit{epistemic uncertainty}, i.e., uncertainty that is an artifact of a lack of knowledge, partial informedness, and ignorance. Opinions about the solutions or constraints of a problem emerge earlier than hard evidence can be produced. Thus, relying on opinions to provide advice to RL agents results in more agile and efficient RL methods. For this, epistemic uncertainty has to be approached in a formal way in order to avoid unsound advice that would adversely affect the agent.
Unfortunately, despite the fact that the importance of dealing with uncertainty in machine learning has been well-recognized~\citep{hullermeier2021aleatoric}, there is a lack of RL methods that treat epistemic uncertainty as a first-class citizen in guided RL. Specifically, methods for managing epistemic uncertainty are missing~\citep{hullermeier2021aleatoric}.

Traditional probability cannot handle constructs such as ignorance or partial knowledge, and can lead to unsound formalization of the advice. For example, when the advisor lacks knowledge about a particular situation, the only way to model the advice is to assign a 0.5 probability to it. A probability of 0.5 means that given a statement \textit{x} captured in the advice (e.g., ``there is high reward at location [2, 3]'' in a grid world), \textit{x} and \textit{not (x)} are equally likely. This clearly does not represent ignorance and, thus, could mislead the RL agent. In general, forcing users to express their opinions by traditional probability could lead to unreliable conclusions. The RL agent needs to know that the advice is based on lacking knowledge in order to consider it with proper weight. It would be, thus, preferable to be able to say ``\textit{I don’t know exactly}'' or ``\textit{I'm not sure}'' within a piece of advice, and couple these modifiers with a particular statement (which might exhibit traits of traditional probability).

Subjective logic~\citep{josang2016subjective} is an extension of probabilistic logic~\citep{adams1996primer}, in which users can express opinions by quantified parameters of belief and certainty.
Opinions are formed from a belief component and an uncertainty component. In statistics and economics, the uncertainty component of subjective logic is often called second-order probability~\citep{gardenfors2005unreliable}, and is represented in terms of a probability density function over first-order probabilities.
Unfortunately, the complex framework of subjective logic limits its applicability. Most techniques that rely on it resort to simplifications, e.g., assuming that humans can express their opinions in terms of mathematical abstractions~\citep{burgueno2023dealing}; or emulating human opinion by objectively measured metrics, taking away the core subjective element of the approach~\citep{walkinshaw2020reasoning}.

\phantom{}

\subparagraph{Goal and research questions}
Our goal is to evaluate the feasibility and utility of guiding RL agents by opinions---i.e., through advice that is subject to belief uncertainty---captured in constructs of subjective logic. To meet our goal, we formulate the following research questions.

\phantom{}

\begin{enumerate}[\bfseries{RQ}1.]
  \item \textit{How can one use opinions to guide reinforcement learning agents?}\\
  By answering this research question, we aim to identify methods to formally externalize uncertain opinions and introduce them into the agent's policy as guiding pieces of information.

  \phantom{}
  
  \item \textit{How does opinion-based guidance affect the performance of reinforcement learning agents?}\\
  By answering this research question, we aim to identify relevant changes in key performance metrics of opinion-guided RL agents.
\end{enumerate}

\phantom{}

We develop a method in which opinions are formulated through the sound mathematical foundations of subjective logic, and fused into the policy of RL agents. Subsequently, we evaluate the performance of the advised agent through a series of experiments. 
Our results indicate that opinions, even if uncertain, improve the performance of RL agents, resulting in higher rewards, more efficient exploration, and a better reinforced policy. Since opinions emerge earlier and more easily than hard evidence can be produced~\citep{dagenais2024driving}, opinion-guided RL offers a more efficient and economical alternative to traditional guided RL approaches.

\phantom{}

\subparagraph{Contributions} The main contribution of this work is a modeling approach for guiding RL through opinions, i.e., advice that is subject to belief uncertainty. We investigate various flavors of RL algorithms and show where and how to augment them with advice expressed via subjective logic. Since the human-in-the-loop can quickly become a bottleneck in human-machine collaboration due to its expensive reward function~\citep{christiano2017deep}, we also recommend architectural patterns to implement opinion-guided RL.
Through detailed experiments, we show that the performance of RL agents improves with added human input and improves again with added uncertainty and disbelief information.
Our approach fosters more efficient collaboration between human and machine agents~\citep{li2021survey} while improving the performance of RL agents.

\phantom{}

\phantom{}

\subparagraph{Structure}
The rest of this paper is structured as follows.
In \secref{sec:example}, we present the running example we use throughout the paper to illustrate key concepts, and to drive the evaluation of our approach.
In \secref{sec:background}, we briefly review the background topics relevant to our work and review the related work.
In \secref{sec:approach}, we outline our approach to modeling human belief and certainty in RL.
In \secref{sec:evaluation}, we evaluate our approach under various advice types, sources, and learning parameters.
In \secref{sec:discussion}, we discuss the results and their implications and outline a set of open challenges.
In \secref{sec:threats}, we reflect on the threats to the validity of this study and discuss the mitigation strategies we applied.
In \secref{sec:conclusion}, we draw the conclusions and outline future work.
\section{Running example}\label{sec:example}

To illustrate the principles of our approach, we rely on the following running example of a self-driving car through a frozen lake. The example is analogous to Open AI’s Gym's Frozen Lake environment.\footnote{\url{https://gymnasium.farama.org/environments/toy_text/frozen_lake/}}

\begin{figure}[htb]
    \centering
    \includegraphics[width=0.33\linewidth]{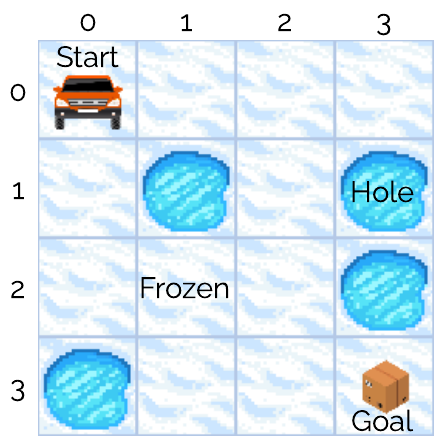}
    \caption{The Frozen Lake running example}
    \label{fig:frozenlake}
\end{figure}

\figref{fig:frozenlake} visualizes the Frozen Lake; an ordinary grid world, on which an agent (a self-driving car) moves one square at a time. The agent begins its exploration at the top leftmost tile (\textit{Start}). Its aim is to reach the bottom rightmost tile (\textit{Goal}).
The agent traverses the Frozen Lake one tile at a time by choosing to move up, down, left, or right. There are three types of tiles the agent may encounter: several \textit{frozen} tiles and \textit{holes}, and one \textit{goal} tile. Encountering these tiles has different consequences, as shown in \tabref{tab:rewards}.

\begin{table}[h]
\centering
\caption{Reward structure}
\label{tab:rewards}
\begin{tabular}{lp{6.5cm}r}
\toprule
\multicolumn{1}{c}{\textbf{Event}} & \multicolumn{1}{c}{\textbf{Consequence}} & \multicolumn{1}{c}{\textbf{Reward}} \\ \midrule
Encountering a frozen tile& Nothing happens; the agent is allowed to continue the traversal of the environment. & 0 \\
Encountering a hole& The task ends unsuccessfully. & 0\\
Encountering the goal & The task is completed successfully. & 1 \\\bottomrule
\end{tabular}
\end{table}



The goal of the agent is to learn the optimal path from the top leftmost tile to the bottom rightmost tile under the given reward structure. Through trial-and-error, the agent gradually learns which actions are beneficial in a specific situation.
For example, in cell \cell{2}{2}, the agent should not move to the right, because this action would result in stepping into a hole and terminating the exploration unsuccessfully, with a reward of 0. Instead, in cell \cell{2}{2}, the agent should move down, because cell \cell{3}{2} is a frozen tile, and it it allows the agent to move closer to the goal, which is in cell \cell{3}{3}.

Eventually, the agent will develop a policy, which will allow it to find the goal faster and in a more reliable fashion (i.e., not falling into holes).

\section{Background and related work}\label{sec:background}

Here, we provide a brief overview of the background of our work and review the related approaches.

\subsection{Reinforcement learning}

\begin{figure}[htb]
    \centering
    \includegraphics[width=0.5\linewidth]{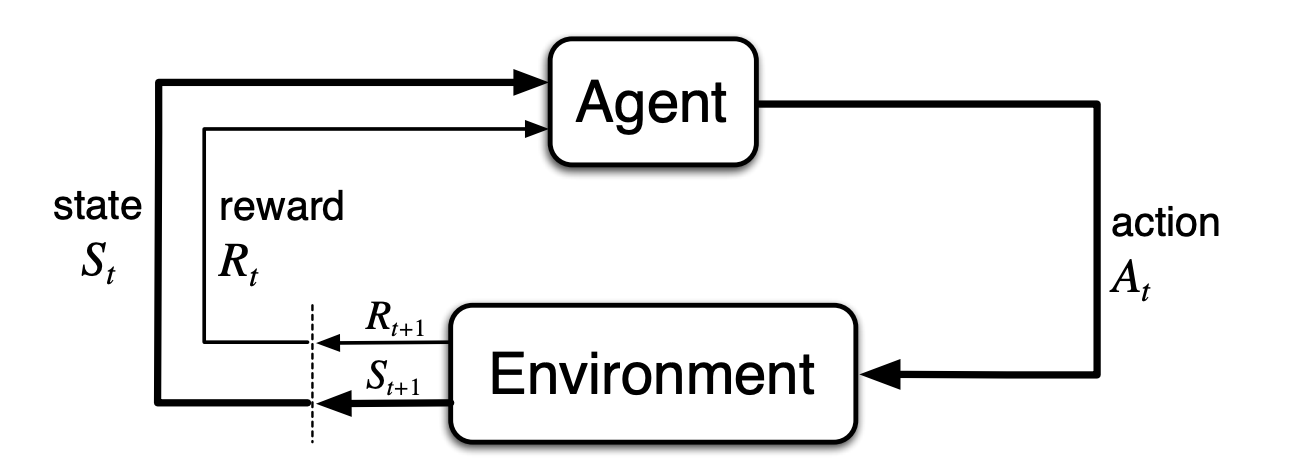}
    \caption{Reinforcement learning -- conceptual overview \citep{sutton2018reinforcement}}
    \label{fig:rl}
\end{figure}

The RL setting (\figref{fig:rl}) as described by \citet{sutton2018reinforcement}, involves an \textit{agent} that sequentially explores its \textit{environment}, and uses feedback in the form of \textit{rewards} to learn the optimal control of said environment. Typically, the RL process is formalized by finite horizon Markov decision processes~\citep{puterman1990markov}, defined as a four-tuple $\langle \mathcal{S}, \mathcal{A}, P, r_t(s, a)\rangle$. Here, $\mathcal{S}$ denotes the set of states the agent may observe from the environment. $\mathcal{A}$ is the set of actions the agent may take in the environment. $\mathcal{R} \subset \mathbb{R}$ is the set of rewards, and $r_t(s, a) \in \mathcal{R}$ is the numeric reward received for choosing action $a \in \mathcal{A}$ while in state $s \in \mathcal{S}$ at time $t$. $P = p(s', r|s,a)$ defines the dynamics of the environment, that is, the probability of the environment transitioning to state $s' \in \mathcal{S}$ and producing reward $r \in \mathcal{R}$ given the agent is in state $s \in \mathcal{S}$ and performs action $a \in \mathcal{A}$. At each time step $t$, the agent observes a state $S_t \in \mathcal{S}$, and carries out an action $A_t \in \mathcal{A}$. At the next time step, $t+1$, the environment produces a new state $S_{t+1}$ as well as a reward $R_{t+1}\in \mathcal{R}$ based on the environment dynamics defined by $P$.

The agent uses a mapping from states to actions, called the policy $\pi(a|s)$, which gives the probability of the agent taking action $a \in \mathcal{A}$ given state $s \in \mathcal{S}$. The agent's sequence of actions, as well as the states and rewards produced by the environment define the \textit{trajectory} $\tau = (S_0, A_0, R_1,..., S_{T-1}, A_{T-1}, R_T)$ where $T$ is the time of termination. By sampling many trajectories through acting in the environment, the goal of the agent is to act according to an optimal policy $\pi_*$, that maximizes the sum of rewards, called the \textit{expected return}, defined as $G_t = R_{t+1} + R_{t+2}+...+ R_T$.

\subsubsection{Various flavors of RL}
While all RL algorithms aim to maximize expected return, various methods achieve this goal differently. \textit{Value-based methods} aim to maximize either the state-value function $v_\pi(s)= \mathbb{E}_\pi [G_t |s]$, the action-value function $q_\pi(s,a)= \mathbb{E}_\pi [G_t |s,a]$, or both. Value-based approaches are deterministic because they select actions greedily when maximizing the value function, which might lead to under-exploration. In contrast, the goal of \textit{policy-based} methods is to learn a parameterized policy that maximizes the reward function in every state without using value functions. Policy-based methods explore the state space more thoroughly than value-based methods, but produce estimates with more noise, potentially resulting in unstable learning processes. Finally, \textit{actor-critic} methods provide a trade-off between value-based and policy-based methods, in which the actor implements a policy-based strategy, and the critic criticizes the actions made by the actor based on a value function. Different methods offer various benefits and perform with unique utility in specific learning problems. 

In particular, \textit{policy gradient} is a subset of the aforementioned policy-based methods. In policy gradient, the policy is parameterized through a parameter vector $\boldsymbol{\theta}$. For reasonably sized action/state spaces, the policy may be parameterized as numerical preferences for all state-action pairs $h(s, a, \boldsymbol{\theta}) \in \mathbb{R}$, and actions can then be chosen using a soft-max distribution. This is called \textit{discrete policy gradient}, and often, a matrix-like representation of a look-up table is used. In contrast, the policy may be parameterized via a neural network such that $\boldsymbol{\theta}$ is a vector of network connection weights or linear in-features, which defines \textit{deep policy gradient}. In either case, the policy is defined as $\pi(a|s, \boldsymbol{\theta})$, the probability of taking action $a$ in state $s$ with parameter $\boldsymbol{\theta}$. The policy may be parameterized in any manner, subject to the condition that it is differentiable with respect to $\boldsymbol{\theta}$. This way, $\boldsymbol{\theta}$ may be updated to maximize some scalar performance metric $J(\boldsymbol{\theta})$. 

\subsection{Subjective logic}\label{sec:sl}
To formalize human guidance in RL, we rely on subjective logic. Subjective logic~\citep{josang2016subjective} is an extension of probabilistic logic~\citep{adams1996primer}, in which users can express opinions by quantified parameters of belief and certainty. Quantifying opinions is an improvement over traditional logic systems as it promotes opinions to first-class citizens in a systematic fashion. Well-informed opinions can be early indicators of emerging new knowledge and often, opinions are the only available insights in engineering. Subjective logic has seen growing applications, such as in model-driven software engineering, e.g., by \citet{burgueno2023dealing}, where it is used to model \textit{belief uncertainty}~\citep{barquero2021improving} in domain models. As well, it has been used in knowledge graphs as per \citet{navarrete2021introducing} to allow reasoning about graph databases enriched with uncertainty.

\subsubsection{Formal underpinnings}
Given a boolean predicate $x$, a binomial opinion regarding the truth of $x$ is given by $\omega_x = (b_x, d_x, u_x, a_x)$, where $b_x$ represents the belief in the truthfulness of $x$; $d_x$ represents the disbelief in the truthfulness of $x$; $u_x$ quantifies the degree of epistemic uncertainty or uncommitted belief concerning the truthfulness of $x$; and $a_x$ represents the base rate, i.e., the prior probability of $x$ being true in the absence of (dis)belief. For each parameter, $0 \leq b_x, d_x, u_x, a_x \leq 1$.
These parameters satisfy the following conditions.
\begin{align}
    b_x + d_x + u_x = 1\label{eq:bdu}\\
    P_x = b_x + a_x u_x.\label{eq:probbau}
\end{align}

\equref{eq:probbau} expresses the \textit{projected probability} of an opinion, effectively transforming opinions into the probability domain, where $P(x)$ means the probability that boolean predicate $x$ holds. It is clear that as uncertainty $u_x$ increases, projected probability $P_x$ is closer to base rate $a_x$. In contrast, as uncertainty $u_x$ decreases, projected probability $P_x$ is closer to that of the belief parameter $b_x$.
Both Equations \ref{eq:bdu} and \ref{eq:probbau} are important invariants that we rely on throughout the paper, particularly in Sections \ref{sec:approach} and \ref{sec:evaluation}.

A traditional probability $p$ corresponds with $u_x =0$, and can thus be transformed into a binomial opinion as follows. 
\begin{align}
    \omega_x = (p, 1-p, 0, p)\label{eq:prob2opinion}
\end{align}

Informed opinions work well at scale. That is, the more experts that express their opinions, the higher the credibility of a collective opinion. Collective opinions can be inferred by semantically sound fusion operators~\citep{josang2013determining}. A fusion operator is a function $f: \Omega \times \Omega \to \Omega$ that maps a pair of opinions onto a new, fused opinion. Various fusion semantics exist, and the right operator must be chosen based on its fit for purpose. A detailed account of fusion operators is given in \secref{sec:fusion}.




\exampleheader{Formulating opinion about the Frozen Lake}
Let $x$ be a boolean predicate regarding frozen lake as follows; $x$: action $a$ is beneficial when in state $s$. Now, consider moving right while in cell $[1,0]$ of \figref{fig:frozenlake}. This would cause the agent to move to a hole tile, resulting in the task to ending unsuccessfully.  In this case, a reasonable opinion would be that the action of moving right is \textit{likely not beneficial} while in cell $[1,0]$, corresponding to $\omega = (0.0, 1.0, 0.0, 0.25)$, meaning there is absolute disbelief that moving right while in cell $[1,0]$ is beneficial. 

\subsubsection{Challenges in employing subjective logic}
The high expressive power of subjective logic comes with the added challenge of calibrating its parameters, particularly uncertainty $u$ and base rate $a$. There are two schools of thought to address this problem. In permissive techniques, setting uncertainty and belief-disbelief are deferred to the user~\citep{burgueno2023dealing,jongeling2023uncertainty}, and it is assumed that these parameters will become available at some point. However, it is usually not explained when, how, and who has to set these parameters, leading to a limited applicability of subjective logic.
%
In restrictive techniques, users are not asked to parameterize the framework. Instead, parameters are controlled by external measures~\citep{josang2000legal,margoni2023subjective}. Unfortunately, this takes away subjective elements from subjective logic. \citet{margoni2023subjective} and \citet{walkinshaw2020reasoning} use statistical evidence to set values, employing subjective logic to analyze empirical studies and experimental results, respectively.
In regards to base rate $a_x$, authors often opt to set this value via statistical evidence~\citep{josang2000legal,margoni2023subjective}. \citet{navarrete2021introducing} allow users to directly indicate the base rate, and \citet{walkinshaw2020reasoning} assume $a_x = 0.5$.
%
%

In our approach, we alleviate to cognitive load on the advisor by deriving base rate $a$ from the structure of the problem; and calibrate uncertainty $u$ by appropriate distance metrics between advisor and advised state (\secref{sec:advice-to-opinion}).


\subsection{Related work}

The closest ours is the work fo \citet{guo2022mtirl}, who combine feedback from several human trainers to create a reliable reward for the RL agent. Subjective logic is used to model the agent's opinion or trust of the human trainer. Converting this opinion into a probability yields a measure of the trainer's \textit{trustworthiness}, and after the agent utilizes human feedback, the agent's opinion about the trainers is updated. In contrast, our work uses subjective logic the other way around: to model the \textit{advisor's opinion about the environment}, and uses that opinion to drive policy shaping and improve the agent's performance. 

Other related works focus either on guiding RL agents but without the use of subjective logic (\secref{sec:guided-rl}) and employing subjective logic in RL but not for guidance (\secref{sec:sl-in-rl}).

\subsubsection{Subjective logic in reinforcement learning (but not for guidance)}\label{sec:sl-in-rl}
The combination of RL and subjective logic is present in a number of publications. Notable work includes the one by \citet{zennaro2020using}, an initial inquiry into using subjective logic in multi-armed bandits, a simplified RL problem. In this work, the agent forms an opinion over the available actions, and if the agent receives a positive reward for an action, the opinion about said action can be updated. Subjective logic to capture uncertainty in multi-armed bandit like problems is extended by \citet{wang2022proactive}. Similarly, \citet{guo2023uncertainty} use a Deep Reinforcement Learning (DRL) multi-agent  in smart farm monitoring to collect data from sensors on cattle. Opinions about each animal's condition are computed via this sensor data. \cite{zhou2021learning} use RL to learn a low dimensional representation of safe regions of complex dynamical systems. Safety of states is estimated using DSAF (Discretized Safety Assessment Function), which is described with subjective logic.
\citet{guo2023uncertainty} use subjective logic to model and compute uncertainty in a DRL framework used to identify the intents of tweets.
\citet{zhao2019uncertainty} use opinions to represent edges in a graph network, and for unknown edges, a set of best paths is determined via DRL. This work proposes 3 different DRL models, each with reward given based on different types of uncertainty that are computed via subjective logic. Similarly, \citet{zhao2019uncertainty} use subjective logic to represent users within an Online Social Network. Different types of users are initialized with particular opinions, which can be updated via subjective logic fusion operators to enhance DRL based Competitive Influence Maximization. While these works combine RL and subjective logic, our work differs by modeling both RL policy and user advice in subjective logic, and using this model to guide the subjective logic agent via human opinion.

\subsubsection{Guidance in reinforcement learning (without Subjective Logic)}\label{sec:guided-rl}
While RL algorithms are self-sufficient-that is, they can run without external intervention-there is a large body of work related to incorporating external human knowledge into these algorithms to improve performance. There are a number of differing approaches for incorporating human advice into RL without the use of subjective logic. \citet{najar2021reinforcement} survey the field of RL with human advice, which they define as ``\textit{teaching signals that can be communicated by the teacher to the learning system without executing the task}''. They define a taxonomy of how advice is provided, first differentiating general advice from contextual advice. General advice is characterized by general constraints and construction, such as if-then rules~\citep{maclin1996creating}, or action plans~\citep{vogel2010learning}. In contrast, contextual advice depends on the context in which it is given, and is subdivided into feedback, and guidance. Feedback methods aim to be evaluative and corrective~\citep{cruz2015interactive,dai2023safe}. 
In preference based RL~\citep{f_urnkranz2012preference}, trajectories are labeled with human preferences. A set of preferences is a partial ordering of trajectories. In this manner, one trajectory can be labeled as preferred over the other.
Preferences allow for limited guidance of RL agents, but fall short of representing uncertainty.
%
Recent work \citep{zhang2024learning} has aimed to improve interpretability of agents trained with this framework.

Guidance involves using human input to bias the exploration strategy, such as probabilistic, early, and importance advising as detailed in \citet{cruz2017agent}. Contextual instructions are a subset of guidance, where advice is given about one action in a particular situation. This is present in the work of \citet{grizou2013robot}, where a robot is given guidance about the next action to take via spoken instruction.

\citet{najar2021reinforcement} also describe strategies to incorporate advice into RL algorithms as shaping methods. These methods are defined based on what point in the learning process they integrated, and include reward shaping, value shaping, policy shaping, and decision biasing. Reward shaping is a particular form of advice that is given directly and translated into numerical rewards. This makes rewards provided by the advisor analogous to the rewards provided by the environment. Often, these rewards are given in the form of feedback, e.g., in the work of \citet{tenoriogonzalez2010dynamic}, where verbal feedback is used to provide additional reward to the agent. In value shaping, advice is considered as a function of action preference. This advice can be used to update the agent's value function~\citep{najar2016training}. In policy shaping, advice is incorporated into the agents policy to bias the exploration strategy. This has been used in several works~\citep{griffith2013policy,knox2009shaping,kessler_faulkner2021interactive,brawer2023interactive,cederborg2015policy}. In decision biasing, advice is used to directly influence the policy output. This advice is not incorporated into any structures of the RL algorithm, and the agent learns from the outcomes of following the advice. This biasing may restrict the actions available to the agent when in a particular state~\citep{thomaz2006reinforcement}. Following this taxonomy, our work falls under the category of guidance, and specifically, policy shaping, as we aim to bias the agent's exploration by infusing the policy with human opinion by way of subjective logic. We are confident, however, with alteration, that our method may be applicable to other advisement mechanisms. 

Other methods not explicitly mentioned in the survey include demonstration or imitation based learning~\citep{celemin2022interactive,pertsch2022demonstration,wu2023human}. Such training protocols have limitations, however, as the human must know how to properly demonstrate the task, and may become the bottleneck in learning leading to increased time to train the agent. Our work facilitates quicker human-machine interaction as the agent does not require continuous human interaction through the process in order to learn. As an alternative, \citet{alshiekh2018safe} propose the use of \textit{shielding}, in which a shield is placed pre-exploration, listing safe actions for the agent to take if it tries to traverse an unsafe region. The shield may instead be placed during exploration if the agent decides to take an unsafe action. Through the use of subjective logic, our work instead provides more semantically rich information to the agent rather than a list of alternative actions.

\section{Approach to guiding reinforcement learning agents by opinions}\label{sec:approach}

In this section, we present our approach to opinion-guided reinforcement learning, with an emphasis on modeling belief uncertainty for guiding purposes.
We argue that subjective logic is an appropriate formalism to encode opinions, and therefore, opinions should be captured in terms of subjective logic. However, as explained in \secref{sec:background}, subjective logic is not intuitive to human advisors, and therefore, we recommend supporting the human advisor with a domain-specific language (DSL)~\citep{schmidt2006model} to express their opinions.
Calibrating the uncertainty parameter \textit{u} of subjective logic is particularly challenging.

\begin{figure}[htb]
    \centering
    \includegraphics[width=\linewidth]{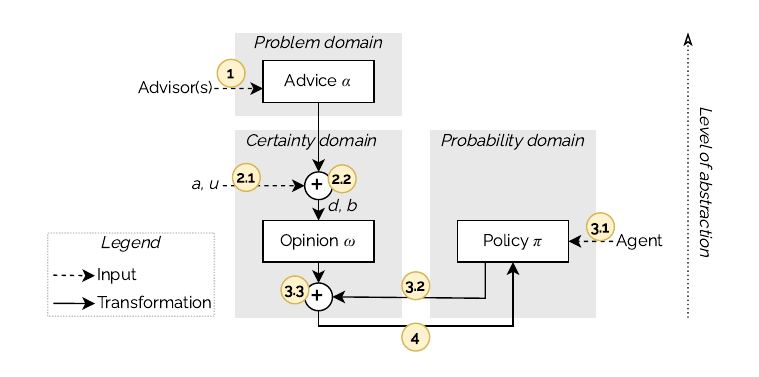}
    \caption{Overview of the approach}
    \label{fig:approach}
\end{figure}

Thus, as shown in \figref{fig:approach}, in our approach, we construct opinions in two steps and use them as the guiding advice in a subsequent step.
%
%
%
First, (\colornumber{1}) the advisor (or multiple advisors) provide(s) their advice through a suitable DSL, without the notion of uncertainty. Subsequently (\colornumber{2}), the advice is translated to an opinion by factoring in the level of uncertainty \textit{u} and base rate \textit{a} of the opinion. Uncertainty is calibrated by a suitable external function, and the base rate is obtained from the structure of the reinforcement learning problem at hand. Opinions are formulated in terms of subjective logic (certainty domain). To shape the agent's policy with the opinion, the policy is translated to subjective logic as well, and (\colornumber{3}) the advisor's opinion is fused into it within the mathematically sound framework of subjective logic.
Subjective logic provides an appropriate framework for our approach for several reasons. First, probabilities can be expressed as opinions without loss of information. As well, subjective logic defines several fusion operators that merge information. Finally, subjective logic captures uncertainty inherent to human advice. Thus, raising probabilities to the domain of subjective logic allows for the consolidation of the agent's policy and the advisor's knowledge, leading to a more informed and realistic measure of the ideal policy, while incorporating uncertainty.
Eventually (\colornumber{4}), the fused opinion is translated back to the probability domain as the shaped policy.


We now elaborate on each of these steps in detail, using the running example of \secref{sec:example} to illustrate concepts. (Subsections \ref{sec:approach-providing-advice}--\ref{sec:fused-to-policy} correspond to steps \colornumber{1}--\colornumber{4} in \figref{fig:approach}.)

\subsection{Providing advice}\label{sec:approach-providing-advice}

In the first step of the approach (\colornumber{1}), the advisor provides their \textit{advice} about the benefits of the agent being in a specific state, as perceived by the advisor. Advice is a subjective construct of the advisor and expresses beliefs about a statement, e.g., the reward that might be received for occupying a particular state. Advice can be expressed at any time, given sufficient support for interaction. In this work, we assume that advice is provided before the exploration begins; however, our approach can be extended to interactive RL scenarios too~\citep{celemin2022interactive}.

In our approach an advice $\alpha$ is given by a mapping $\alpha: s \in S' \mapsto v$, where
$S' \subseteq S$ is the known subset of the total problem space, $s \in S'$ is a specific state within the known problem space, and $v$ is a value that expresses the benefit of the particular state in a suitable language. Since it is not always feasible to survey the whole problem space, we only assume that the advisor provides a set of advice w.r.t. $S' \subseteq S$.
For example, in \secref{sec:example}, a partially informed advisor could formulate the advice ``\textit{Cell (1,1)} [a hole] \textit{is likely not beneficial}''. Here, $s$ refers to cell [1,1] in the grid world, and ``not beneficial'' is the value of occupying cell \cell{1}{1}.
This advice is then used to update the state-action pairs that lead to the advised cell from neighboring cells, thereby shaping the policy.

Unfortunately, the notion of value in the advice is not quantitative enough as of yet. 
Thus, advice needs to be formulated in a suitable domain-specific language (DSL)~\citep{schmidt2006model} that either allows for expressing quantitative advice intuitively or quantifies the advice in the background. DSLs are languages tailored to a specific problem and offer a restricted set of syntactic elements to express statements about the problem domain. This allows for working with domain concepts, e.g., expressing the value of occupying a cell in the frozen lake, rather than having to express the reward function in the usual terms of RL. Working with domain concepts, in turn, narrows the gap between RL and human cognition, substantially improving the usability of the approach.

\exampleheader{The GridWorld DSL for the Frozen Lake problem}

For the purposes of this article, we define the DSL in \lstref{lst:gridworld-dsl} to provide advice about the grid world of the running example (\figref{fig:frozenlake}). An example set of advice expressed in the DSL about the running example (\figref{fig:frozenlake}) is shown in \lstref{lst:gridworld-dsl-1}.

\begin{figure}[h]
  \begin{lstlisting}[mathescape, language={gridworld-dsl}, caption={GridWorld grammar in EBNF notation}, label={lst:gridworld-dsl}]
<AdviceList> ::= <Advice>+
<Advice> ::= <AdviceLocation> `,` <AdviceValue>
<AdviceLocation> ::= `[`[0-9]+, [0-9]+`]`
<AdviceValue> ::= ^-?[0-2]
  \end{lstlisting}
\end{figure}

\begin{figure}[h]
  \begin{lstinputlisting}[mathescape, caption={Example set of opinions for the problem in \figref{fig:frozenlake}}, label={lst:gridworld-dsl-1}]{files/lake-1.txt}
  \end{lstinputlisting}
\end{figure}

%
The value of the advice represents the advisor's idea about how beneficial the given cell can be for the agent. In our example DSL, the value is an integer that ranges between -2 and 2. In the example in \lstref{lst:gridworld-dsl-1}, the respective list of advice corresponds to the following cells of the running example (\figref{fig:frozenlake}): two holes on the grid, which are not beneficial (\parameter{1}{1}{-2}; and \parameter{1}{3}{-2}); the goal, which is beneficial (\parameter{3}{3}{+2}); and the frozen tile that is not as hazardous as a hole but still rather disadvantageous (\parameter{0}{3}{-1}) due to the number of holes in its vicinity.

\phantom{}

It is important to note that the above DSL is just an example, and alternative DSLs for different problems could be developed. However, our approach is general enough to accommodate any DSL.

Next, we show how advice is translated into the certainty domain as an opinion.

\subsection{Translating advice to opinion}\label{sec:advice-to-opinion}

To treat advice by formal means, we map them onto opinions of subjective logic. As shown in \figref{fig:approach} (\colornumber{2}), this requires calibrating the level of uncertainty \textit{u} at which the advice was communicated and setting base rate \textit{a} (\secref{sec:approach-u}); and subsequently, compiling the advice at the specific level of uncertainty into an opinion by calculating the remaining parameters of subjective logic: disbelief \textit{d} and belief \textit{b} (\secref{sec:approach-advice-compilation}).

\subsubsection{Calibrating the base rate (\textit{a}) and the level of uncertainty (\textit{u})}\label{sec:approach-u}

\paragraph{Calibrating base rate}

Base rate \textit{a} is the prior probability in the absence of belief or disbelief. In the context of RL, the base rate represents the default likelihood of the agent taking a specific action in a specific state. Thus, given a set of actions \textit{A}, the base rate is given as
\begin{align}
    a = \frac{1}{|A|}. \label{eq:a}
\end{align}

In the grid world of the running example, the base rate is $0.25$ in each state because the agent can choose from four actions to move to the next state. (If the agent is situated in an edge or corner cell and chooses an action that would lead out of the grid world, the agent stays in place.)

A similar approach is suggested by \citet[Sec 3.1]{zennaro2020using}.

\paragraph{Calibrating uncertainty}
As explained in \secref{sec:background}, there are two classes of methods to calibrate uncertainty \textit{u}: manual (deferred to the advisor) and automated (calibrated by an appropriate metric). Since human advisors might find it challenging to quantify the uncertainty of their advice, in our approach, we recommend automated derivation of the uncertainty metric.

Uncertainty can be derived, for example, from an appropriate distance metric, which assumes that distance discounts certainty, i.e., uncertainty increases with distance.
Distance can be a pure topological concept (e.g., a Manhattan distance in a grid world); or a more abstract concept, such as a trace distance between design models~\citep{dagenais2024driving,david2016towards} or a quantified notion of subject matter expertise.
%
%

There are two components to calibrating uncertainty by a distance measure: (i) choosing the specific distance measure, and (ii) choosing the discount function that progressively discounts certainty as distance increases.

In general the discount function is given as $\gamma: (u_{max}, \delta) \mapsto u \in (0, 1)$, where $\delta$ is a distance measured by a distance measure $\Delta$, and $u_{max}$ is the upper bound of uncertainty. From the identity of $u = 1 - (b+d)$ (\secref{sec:sl}), it follows that $u_{max} = 1 - (b+d)$.



In the most simple case, a liner discount function can be chosen to calibrate uncertainty as follows.
\begin{align}
    u = \frac{\delta}{\delta_{max}} \times u_{max} \label{eq:discount}
\end{align}
Here, $\frac{\delta}{\delta_{max}}$ is the distance relative to the maximum distance, that maps onto the (0,1) domain.

In some problems, uncertainty should not be scaled over the complete problem space, but to a subset of it.
For such cases, a threshold $0< \tau \in \Delta < 1$ is used to accelerate the discounting of certainty and reach $u_{max}$ in $\tau$. Generally, \equref{eq:discount} is adopted as follows.

\begin{equation}\label{eq:h2-consistency}
  u =
  \begin{cases}
    \frac{1}{\tau} \times \frac{\delta}{\delta_{max}} \times u_{max} & \mbox{if } \delta \leq (\tau\times\delta_{max}); \\
    u_{max} & \mbox{if } \delta>(\tau\times\delta_{max}).
  \end{cases}
\end{equation}

\subsubsection{Compiling advice into opinion by computing disbelief (\textit{d}) and belief (\textit{b})}\label{sec:approach-advice-compilation}

Once uncertainty $u$ has been calibrated, disbelief $d_i$ and belief $b_i$ of opinion $\omega_i$, for advice $\alpha_i$ can be computed.
Given $0 \leq u \leq 1$, the remainder of opinion weights has to be distributed between $b_i$ and $d_i$ in accordance with \equref{eq:bdu}.
The calculation method of belief $b_i$ and disbelief $d_i$ for the $j$th item in the $n$-point scale used for expressing advice values, in ascending order of confidence from least to most confident, are given as follows.
\begin{align}
    b_i &= \frac{j-1}{n-1} \times (1-u)~|~j \in \{1..n\} \label{eq:b} \\
    d_i &= (1-u) - b_i. \label{eq:d}
\end{align}
The corresponding algorithm is shown in \algoref{alg:advice2opinion}. Here, $\frac{j-1}{n-1}$ is the weight between \textit{b} and \textit{d} to split the remaining opinion weight of $1-u$.

Eventually, using Equations \ref{eq:a}--\ref{eq:d}, the opinion is given as
\begin{align*}
    \omega_i = \left(\frac{j-1}{n-1} \times (1-u), \frac{n-j}{n-1} \times (1-u), u, \frac{1}{|A|}\right).
\end{align*}

(Here, we also used that (\ref{eq:d})$\leftarrow$(\ref{eq:b}) reduces to $\frac{n-j}{n-1} \times (1-u)$.)

\begin{algorithm}
\caption{Compiling advice into opinion} 
\label{alg:advice2opinion}
\begin{algorithmic}
    \Require $Advice~\alpha$
    \Require $a$ \Comment{inferred base rate}
    \Require $u$ \Comment{automatically calibrated uncertainty}
    \Require $n$ \Comment{advice scale length}
    \State $Opinion~\omega$
        \State $\omega.a \gets a$
        \State $\omega.u \gets u$
        \State $\omega.b \gets (order(\alpha.value)-1)/(n-1) \times (1 - \omega.u)$
        \State $\omega.d \gets 1 - (\omega.b + \omega.u)$
    \State \Return $\omega$
\end{algorithmic}
\end{algorithm}

Three example mappings are shown in \tabref{tab:opinion-to-sl}. For example, advice value $+1$ is the element number $4$ in the advice scale. At uncertainty level $u=0.2$, the remaining opinion weight is $1-0.2 = 0.8$; which is distributed between \textit{b} and \textit{d} in a $\frac{j-1}{n-1} = \frac{3}{4}$ weight towards \textit{b}. That is, $b_{4} = 0.8 \times \frac{3}{4} = 0.6$; and consequently, $b_{4} = 0.2$.

\begin{table}[h]
\centering
\small
\caption{Example mappings of confidence levels onto the belief dimension of SL at different degrees of uncertainty $u$}
\label{tab:opinion-to-sl}
\begin{tabular}{@{}rcrcrrcrrcrrcrr@{}}
\toprule
\multicolumn{1}{c}{\multirow{2}{*}{\makecell[c]{\textbf{Advice}\\\textbf{value}}}} && \multicolumn{1}{c}{\multirow{2}{*}{\textbf{Order \textit{j}}}} && \multicolumn{2}{c}{\textit{u=0.0}} && \multicolumn{2}{c}{\textit{u=0.2}} && \multicolumn{2}{c}{\textit{u=0.5}} &&
\multicolumn{2}{c}{\textit{u=0.833}} \\ \cmidrule(l{0.75em}r{0.75em}){5-6} \cmidrule(l{0.75em}r{0.75em}){8-9} \cmidrule(l{0.5em}r{0.5em}){11-12} \cmidrule(l{0.5em}r{0.5em}){14-15}
&&&& \multicolumn{1}{c}{\textit{b}} & \multicolumn{1}{c}{\textit{d}} && \multicolumn{1}{c}{\textit{b}} & \multicolumn{1}{c}{\textit{d}} && \multicolumn{1}{c}{\textit{b}} & \multicolumn{1}{c}{\textit{d}} && \multicolumn{1}{c}{\textit{b}} & \multicolumn{1}{c}{\textit{d}}  \\
\cmidrule{1-1} \cmidrule{3-3} \cmidrule{5-6} \cmidrule{8-9} \cmidrule{11-12} \cmidrule{14-15}
-2 && 1 && 0.00 & 1.00 && 0.0 & 0.8 && 0.000 & 0.500 && 0.000 & 0.167 \\
-1 && 2 && 0.25 & 0.75 && 0.2 & 0.6 && 0.125 & 0.375
 && 0.042 & 0.125 \\
0 && 3 && 0.50 & 0.50 && 0.4 & 0.4 && 0.250 & 0.250 && 0.084 & 0.084 \\
+1 && 4 && 0.75 & 0.25 && 0.6 & 0.2 && 0.375 & 0.125 && 0.125 & 0.043 \\
+2 && 5 && 1.00 & 0.00 && 0.8 & 0.0 && 0.500 & 0.000 && 0.167 & 0.000 \\
\bottomrule
\end{tabular}
\end{table}

\exampleheader{From advice to opinions in the Frozen Lake problem}

\begin{figure}[h]
    \centering
    \includegraphics[width=0.9\linewidth]{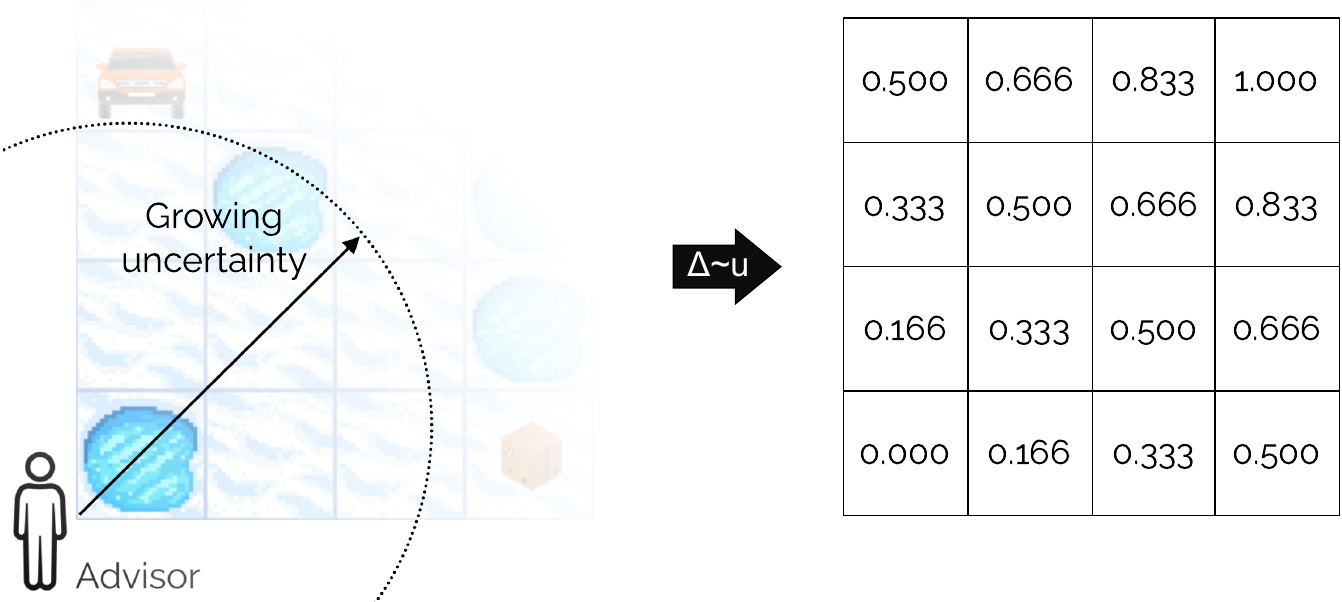}
    \caption{A visual intuition of the advisor's limited knowledge, subject to epistemic uncertainty (left), and the corresponding uncertainty levels of the cells in the grid world. Uncertainty grows with distance---in this specific example, with \textit{topological} distance. Using the distance between the \textit{Advisor} and the location the advice pertains to, the uncertainty of advice can be calibrated.}
    \label{fig:example-uncertainty}
\end{figure}

Given an advisor situated in the bottom-left corner of the field (\figref{fig:example-uncertainty}), we model the uncertainty of a piece of advice by the distance between the advisor and the cell the opinion pertains to. For simplicity, we use a normalized Manhattan distance and consider the bottom-left cell at $u_{(3,0)}=0.0$ and the top-right corner at $u_{(0,3)}=1.0$ uncertainty; and consider a linear relationship between uncertainty and distance. The Manhattan distance is a common distance measure in topological spaces, such as the running example illustrated in \secref{sec:example}. 
The Manhattan distance $D$ between points $X = (x_1,x_2,\hdots, x_n)$ and $Y = (y_1, y_2, \hdots, y_n)$ is defined as 
\begin{align*}
    D(X, Y) = \sum_{i=1}^n|x_i - y_i|.
\end{align*}

Here, by the Manhattan distance and $\tau=0$: $\delta_{max}=6$ (between the two corners), and corresponding to \equref{eq:discount}, $u = \frac{\delta}{6} \times 1.0$. Indeed, each unit of distance from the advisor increases the uncertainty by $1/6=0.166$. This progression constitutes the linear $\gamma$ discount function of certainty.


\begin{table}[h]
\centering
\captionsetup{width=\textwidth}
\caption{The four pieces of advice and their opinion equivalents of the running example}
\label{tab:opinion-distance-advice}
\begin{tabular}{@{}rcrcrrrr@{}}
\toprule
\multicolumn{1}{c}{\multirow{2}{*}{\textbf{Cell}}} && \multicolumn{1}{c}{\multirow{2}{*}{\textbf{Advice}}} && \multicolumn{4}{c}{\textbf{SL}} \\ 
&&&& \multicolumn{1}{c}{\textit{u}} & \multicolumn{1}{c}{\textit{b}} & \multicolumn{1}{c}{\textit{d}} & \multicolumn{1}{c}{\textit{a}} \\
\cmidrule{1-1} \cmidrule{3-3} \cmidrule{5-8}
[1,1] && -2 && 0.500 & 0.000 & 0.500 & 0.250\\

[1,3] && -2 && 0.833 & 0.000 & 0.167 & 0.250 \\ 

[0,3] && -1 && 1.000 & 0.000 & 0.000 & 0.250 \\ 

[3,3] && +2 && 0.500 & 0.500 & 0.000 & 0.250 \\
\bottomrule
\end{tabular}
\end{table}

\subsection{Policy shaping by opinion fusion}\label{sec:policy-shaping}

After having formed opinions from advice, the advisor's input is ready to be used to shape the agent's policy. As shown in \figref{fig:approach} (\colornumber{3}), this requires accessing the agent's policy (\secref{sec:policy}), transforming it from the probability domain to the certainty domain (\secref{sec:policy-prob-to-cert}), and within the framework of subjective logic, fusing it with the advisor's opinion (\secref{sec:fusion}).

\subsubsection{Policy}\label{sec:policy}

The product of the agent's exploration process in RL is the policy $\pi$ that maps states to actions. Formally, policy $\pi$ is given by a mapping $\pi = S \times A \rightarrow P$, where $S$ is the set of states accessible to the agent; $A$ is the set of actions the agent can take, and $P$ is a probability in the usual sense, i.e., $P = [0, 1]$. Policy $\pi(a|s)$ defines the conditional probability of choosing action $a\in A$ in state $s\in S$.
It follows that $\forall s \in S, a \in A: 0 \leq \pi(a|s) \leq 1$ always holds. Furthermore, the default policy is initialized as $\forall s \in S, a_i, a_j \in A, i \neq j: \pi(a_i|s) = \pi(a_j|s)$. That is, before exploration, the probability of choosing any action in a given state is equal.

\exampleheader{Default policy in the Frozen Lake problem} \tabref{tab:default-policy-prob} shows the default policy of the running example. As shown, in each state (here, each cell), there are four actions available for the agent (left, down, right, up). Before training, the default policy is uniform. Each action is equally as likely to be taken in each state. Here, this probability is $0.25$ as there are $4$ possible actions to take in each state. Through trial and error, the agent will modify these probabilities.

\begin{table}[h]
\centering
\captionsetup{width=\textwidth}
\caption{Default policy of the Frozen Lake running example}
\label{tab:default-policy-prob}
\begin{tabular}{@{}rccccc@{}}
\toprule
& Left & Down & Right & Up &
\\
\cmidrule(r){2-2}\cmidrule(r){3-3}\cmidrule(r){4-4}\cmidrule(r){5-5}
[0,0] & 0.25 & 0.25 & 0.25 & 0.25\\

[0,1] & 0.25 & 0.25 & 0.25 & 0.25\\

\multicolumn{5}{c}{...}\\

[3,2] & 0.25 & 0.25 & 0.25 & 0.25\\

[3,3] & 0.25 & 0.25 & 0.25 & 0.25\\
\bottomrule
\end{tabular}
\end{table}

\subsubsection{Transforming the policy into the certainty domain}\label{sec:policy-prob-to-cert}

To be able to shape the policy, we transform it from the \textit{probability} domain to the \textit{certainty} domain. That is, the probabilities of the policy (e.g., in \tabref{tab:default-policy-prob}) are translated to \textit{opinions} through the following mapping:
\begin{align*}
    f_{P\rightarrow \Omega}(\pi): S \times A \rightarrow \Omega~~|~~P.
\end{align*}

That is, mapping $f(\pi)$ assigns an opinion $\omega \in \Omega$ of choosing action $a \in A$ in state $s \in S$, given the probability $p \in P$ of choosing action $a \in A$ in state $s \in S$ under policy $\pi$.
As explained in \secref{sec:sl}, probability $p$ is translated to an opinion $\omega$ as follows:
\begin{align}
    \omega: p \mapsto (p, 1-p, 0, p). \label{eq:pbau}
\end{align}

This definition also shows that base rate $a$ (i.e., the last element of the $\omega$ tuple) is equal to the default probability of choosing an action, aligned with \equref{eq:a}.

The corresponding algorithm is shown in \algoref{alg:prob2cert}. For clarity, we refer to a policy expressed in the probability domain as $\pi_p$, and to a policy expressed in the certainty domain as $\pi_c$.

\begin{algorithm}[h]
\caption{Transforming a policy in the probability domain ($\pi_p$) to a policy in the certainty domain ($\pi_c$)}
\label{alg:prob2cert}
\begin{algorithmic}
    \Require $\pi_p$ \Comment{Policy in the probability domain}
    \State $\pi_c$ \Comment{Policy in the certainty domain}
    \For{\text{Each state} $s \in S$}
        \For{\text{Each action} $a \in A$}
            \State $p \gets \pi_p[s,a]$
            \State $ \pi_c[s,a] \gets (p, 1-p, 0, p)$
        \EndFor
    \EndFor
    \State \Return $\pi_c$
\end{algorithmic}
\end{algorithm}

\exampleheader{Default policy of the Frozen Lake problem in the certainty domain} \tabref{tab:default-policy-certainty} shows the default policy in the certainty domain transformed from the default policy in the probability domain shown in \tabref{tab:default-policy-prob}.

As shown, the previous probabilities of $p=0.25$ are now expressed as opinions, following \equref{eq:pbau} as $\omega = (0.25, 1-0.25, 0, 0.25) = (0.25, 0.75, 0, 0.25)$.

\begin{table}[h]
\centering
\footnotesize
\caption{Default policy of the Frozen Lake running example in the certainty domain}
\label{tab:default-policy-certainty}
\begin{tabular}{@{}rccccc@{}}
\toprule
& Left & Down & Right & Up &
\\
\cmidrule(r){2-2}\cmidrule(r){3-3}\cmidrule(r){4-4}\cmidrule(r){5-5}
[0,0] & (0.25, 0.75, 0.0, 0.25) & (0.25, 0.75, 0.0, 0.25) & (0.25, 0.75, 0.0, 0.25) & (0.25, 0.75, 0.0, 0.25)\\

[0,1] & (0.25, 0.75, 0.0, 0.25) & (0.25, 0.75, 0.0, 0.25) & (0.25, 0.75, 0.0, 0.25) & (0.25, 0.75, 0.0, 0.25)\\

\multicolumn{5}{c}{...}\\

[3,2] & (0.25, 0.75, 0.0, 0.25) & (0.25, 0.75, 0.0, 0.25) & (0.25, 0.75, 0.0, 0.25) & (0.25, 0.75, 0.0, 0.25)\\

[3,3] & (0.25, 0.75, 0.0, 0.25) & (0.25, 0.75, 0.0, 0.25) & (0.25, 0.75, 0.0, 0.25) & (0.25, 0.75, 0.0, 0.25)\\
\bottomrule
\end{tabular}
\end{table}

\subsubsection{Policy shaping by opinion fusion}\label{sec:fusion}

After the policy has been transformed to an opinion-based representation, the advisors' opinions can be combined with it. Combined (joined) opinions are obtained through \textit{fusing} single opinions. By fusion, we mean a mapping $\Omega \times \Omega  \rightarrow \Omega$ that produces an opinion from other opinions.

\paragraph{Locating the opinions to be fused}

In our approach, the advisor provides advice about the value of occupying a specific state. This information affects the agent's decision in states from which the advised state is reachable. To unambiguously formalize this step, we need some definitions.

\phantom{}

\begin{definition}[Neighboring states]
    States $s_i, s_j \in S$ are said to be neighbors if there exists a state-action pair in the policy for which $\pi(a | s_i) \mapsto s_j$, i.e., choosing action $a \in A$ in state $s_i$ will transition the agent to state $s_j$. \label{def:neighbors}
\end{definition}

\phantom{}

The neighborhood $N$ of a state is the set of all its neighbors.

\begin{definition}[Neighborhood (of a state)]
    $N (s_i \in S) = \bigcup_{s_j \in S} \exists a \in A: \pi(a | s_i) \mapsto s_j$. \label{def:neighborhood}
\end{definition}

In the running example, the state space is topological; thus, trivially, neighboring states are the neighboring cells of a given cell.

\phantom{}

Introducing advice $\alpha$ about state $s_i$ to policy $\pi$ is achieved by $\forall a \in N(s_i).A : \omega(\alpha) \odot \omega(a)$. \label{theo:shaping}
That is, the opinion formed from the advice ($\omega(\alpha)$) is fused with every opinion formed from actions ($\omega(a)$) that lead from its neighboring states ($N(s_i)$) to $s_i$.

\paragraph{Choosing a fusion operator}
The seminal work of \citet{josang2016subjective} defines an array of fusion operators and classifies them according to situational characteristics to facilitate the selection of the most appropriate operator. Here, we rely on the Belief Constraint Fusion (BCF) operator, which is an appropriate choice when agents and advisors have already made up their minds and will not seek compromise. Since the agent and advisors formulate opinions about the problem independently from each other, it is fair to assume that they will, indeed, not seek compromise.

As per \citet{josang2016subjective}, fused opinion $\bcf{\omega} = (\bcf{b}, \bcf{d}, \bcf{u}, \bcf{a})$ under Belief Constraint Fusion is calculated from the overlapping beliefs (called \textit{harmony}) and non-overlapping beliefs (\textit{conflict}) of individual opinions.
\begin{align}
    \textit{Harmony} = b_1u_2 + b_2u_1 + b_1b_2;\label{eq:harmony}\\
    \textit{Conflict} = b_1 d_2 + b_2 d_1.
\end{align}

Finally, the parameters of the fused opinion $\bcf{\omega}$ are calculated as follows.
\begin{align}
    \bcf{b} &= \frac{\textit{Harmony}}{(1 - \textit{Conflict})};\\
    \bcf{d} &= 1 - (\bcf{b} + \bcf{u});\\
    \bcf{u} &= \frac{u_1u_2}{(1 - \textit{Conflict})};\\
    \bcf{a} &= \frac{a_1(1-u_1) + a_2(1-u_2)}{2-u_1-u_2} \label{eq:a-joint}
\end{align}

Of course, different fusion operators can be chosen as well. A detailed account of fusion operators is given by \citet{josang2016subjective}.

\paragraph{The case of multiple advisors}

There is no limitation to the number of advisors in our approach. This is due to opinions forming a closed structure under fusion in subjective logic, i.e., fusing two opinions results in another opinion. By that and the commutative nature of fusion operators, arbitrary number of advisors can be involved.

By involving multiple advisors, the expectation is that the performance of guidance will increase. This is due to the fact that advisors might have complementary knowledge, i.e., they might be able to provide high-certainty advice about different parts of the problem space.

\exampleheader{Shaped policy of the Frozen Lake problem in the certainty domain}

\tabref{tab:fused-policy-certainty} shows the impact of a single advice \advice{1}{1}{-2} on the default policy shown in \tabref{tab:default-policy-certainty}. As defined in \theoref{theo:shaping}, shaping the policy impacts the neighbors of the advised state.

\subparagraph{Locating opinions to be fused}
From \defref{def:neighborhood}, $N[1,1] = \{[0,1], [1,0], [1,2], [2,1]\}$, with the respective actions of $\{Down, Right, Left, Up\}$ leading to \cell{1}{1}. The corresponding opinions will be fused with the opinion formed from the advice.

\subparagraph{Fusing opinions} As shown in \tabref{tab:opinion-distance-advice}, advice \advice{1}{1}{-2} translates to $\omega(\alpha) = (0.00, 0.50, 0.50, 0.25)$. In the default policy, each opinion is given by $\omega(a) = (0.25, 0.75, 0.00, 0.25)$.
The fused opinion under Belief Constraint Fusion is calculated as follows.
\begin{align*}
    \textit{Harmony} &= b_1u_2 + b_2u_1 + b_1b_2 = 0.00 \times 0.00 + 0.25 \times 0.50 = 0.125.\\
    \textit{Conflict} &= b_1 d_2 + b_2 d_1 = 0.00 \times 0.75 + 0.25 \times 0.50 = 0.125.\\\\
    \bcf{b} &= \frac{\textit{Harmony}}{(1 - \textit{Conflict})} = \frac{0.125}{1 - 0.125} = 0.143.\\
    \bcf{u} &= \frac{u_1u_2}{(1 - \textit{Conflict})} = \frac{0.50 \times 0.00}{1 - 0.125} = 0.\\
    \bcf{d} &= 1 - (\bcf{b} + \bcf{u}) = 1 - (0.143 + 0) = 0.857.\\
    \bcf{a} &= \frac{a_1(1-u_1) + a_2(1-u_2)}{2-u_1-u_2} = \frac{0.25 \times (1-0.50) + 0.25 \times (1-0.00)}{2-0.50-0.00} = 0.25.
\end{align*}

Thus, the resulting fused opinion $\bcf{\omega} = (0.143, 0.857, 0.000, 0.250)$, as highlighted in \tabref{tab:fused-policy-certainty}.

\begin{table}[h]
\centering
\footnotesize
\setlength{\tabcolsep}{2pt}
\captionsetup{width=\textwidth}
\caption{Advice \texttt{[1,1]$\rightarrow$-2} fused into the policy of the Frozen Lake running example}
\label{tab:fused-policy-certainty}
\begin{tabular}{@{}rccccc@{}}
\toprule
& Left & Down & Right & Up &
\\
\cmidrule(r){2-2}\cmidrule(r){3-3}\cmidrule(r){4-4}\cmidrule(r){5-5}
[0,0] & (0.25, 0.75, 0.0, 0.25) & (0.25, 0.75, 0.0, 0.25) & (0.25, 0.75, 0.0, 0.25) & (0.25, 0.75, 0.0, 0.25)\\

[0,1] & (0.25, 0.75, 0.0, 0.25) & \textbf{\hl{(0.143, 0.857, 0.0, 0.25)}} & (0.25, 0.75, 0.0, 0.25) & (0.25, 0.75, 0.0, 0.25)\\

\multicolumn{5}{c}{...}\\

[1,0] & (0.25, 0.75, 0.0, 0.25) & (0.25, 0.75, 0.0, 0.25) & \textbf{\hl{(0.143, 0.857, 0.0, 0.25)}} & (0.25, 0.75, 0.0, 0.25)\\

\multicolumn{5}{c}{...}\\

[1,2] & \textbf{\hl{(0.143, 0.857, 0.0, 0.25)}} & (0.25, 0.75, 0.0, 0.25) & (0.25, 0.75, 0.0, 0.25) & (0.25, 0.75, 0.0, 0.25)\\

\multicolumn{5}{c}{...}\\

[2,1] & (0.25, 0.75, 0.0, 0.25) & (0.25, 0.75, 0.0, 0.25) & (0.25, 0.75, 0.0, 0.25) & \textbf{\hl{(0.143, 0.857, 0.0, 0.25)}}\\

\multicolumn{5}{c}{...}\\

[3,3] & (0.25, 0.75, 0.0, 0.25) & (0.25, 0.75, 0.0, 0.25) & (0.25, 0.75, 0.0, 0.25) & (0.25, 0.75, 0.0, 0.25)\\
\bottomrule
\end{tabular}
\end{table}

\subsection{Transforming opinion-infused policies from the certainty domain to the probability domain}\label{sec:fused-to-policy}

As the final step of the approach (\figref{fig:approach} -- \colornumber{4}), the policy with the opinions fused into it, is transformed back from the certainty domain to the probability domain to be used by the agent. This is achieved in two steps: translating opinions to probabilities and, subsequently, normalizing probabilities in each state.

\paragraph{Transformation to the probability domain}

The transformation to the probability domain is the inverse transformation of what has been explained in \secref{sec:policy-prob-to-cert}.
As explained in \secref{sec:sl} and defined by \citet{josang2016subjective}, opinion $\omega$ is translated to a probability $p$ as follows:
\begin{align}
    p: \omega \mapsto b + au. \label{eq:pbau-inverse}
\end{align}

\paragraph{Normalization} Since the actions available for the agent in a given state form a complete probability space, the following invariant must always hold in a valid policy $\pi$.

\begin{align}
    \forall s \in S: \sum_{a \in A} p(a | s) = 1. \label{eq:p-invariant}
\end{align}

Thus, we apply the usual normalization to scale the sum of probabilities to 1 as follows.

\begin{align}
    \forall s \in S, a \in A: p(a | s) := p(a | s) \times \frac{1}{\sum_{a \in A} p(a|s)}. \label{eq:normalization}
\end{align}

\exampleheader{Shaped policy of the Frozen Lake problem in the probability domain}

\subparagraph{Calculating probabilities}
We take the fusion of the advice and the policy in \tabref{tab:fused-policy-certainty} and for each tuple, we apply \equref{eq:pbau-inverse}. \tabref{tab:fused-policy-prob} shows the result of this transformation and the new policy that considers advice \advice{1}{1}{-2}.
Consider, e.g., the \texttt{Down} action in state \cell{0}{1}. Here, $\omega = (0.143, 0.857, 0.0, 0.25))$, and from \equref{eq:pbau-inverse}, $p = b + au = 0.143 + 0.25 \times 0.0 = 0.143$ follows.

As seen by the highlighted values, the advice affects the actions of neighboring states that lead to state \cell{1}{1}, i.e., \cell{0}{1} -- \texttt{Down}, \cell{1}{0} -- \texttt{Right}, \cell{1}{2} -- \texttt{Left}, \cell{2}{1} -- \texttt{Up}. \tabref{tab:fused-policy-prob-nonnormalized} and \tabref{tab:fused-policy-prob-normalized} show the policy before and after normalization, respectively.

\subparagraph{Normalization}
The resulting probabilities in row \cell{0}{1} violate invariant \equref{eq:p-invariant} as $\sum_{a \in A} p(a | s=[0,1]) = 0.893$. Thus, we normalize by \equref{eq:normalization} and multiply each probability by $\frac{1}{\sum_{a \in A} p(a|s=[0,1])}$, resulting in the probability vector (0.28, 0.16, 0.28, 0.28), as shown in the \cell{0}{1} row of \tabref{tab:fused-policy-prob-normalized}.

\subparagraph{Full example}
To round out the running example, we show the full policy after considering every piece of advice in \lstref{lst:gridworld-dsl-1}. \tabref{tab:fused-policy-prob-full} shows the the new probabilities before (\tabref{tab:fused-policy-prob-full-nonnormalized}) and after (\tabref{tab:fused-policy-prob-full-normalized}) normalization. As seen, some advice does not change the probabilities of the policy much, e.g., in the \cell{0}{3} -- \texttt{Down} row of \tabref{tab:fused-policy-prob-full-nonnormalized} (0.217, previously 0.250); while other pieces of advice have substantial impact, e.g., in \cell{3}{2} -- \texttt{Right} (0348, previously 0.250). This is due to the uncertainty of the advice we modeled by the distance between the advisor and the advised state. In the running example, the advisor was situated in the bottom-left corner, and therefore, as a consequence, their advice about the distant \cell{1}{3} cell impacted the \texttt{Down} action of cell \cell{0}{3} less than the advice about the closer \cell{3}{3} cell that impacted the \texttt{Right} action of cell \cell{3}{2}.
\clearpage

\begin{table*}
    \captionsetup[subtable]{position=top}
    \captionsetup[table]{position=top}
    \caption{Shaped policy of the Frozen Lake running example in the probability domain after advice \advice{1}{1}{-2} affecting the actions of neighboring states \cell{0}{1}, \cell{1}{0}, \cell{1}{2}, \cell{2}{1} that lead to state \cell{1}{1}}
    \label{tab:fused-policy-prob}
    \begin{subtable}{0.5\linewidth}
        \caption{Before normalization}
        \label{tab:fused-policy-prob-nonnormalized}
        \begin{tabular}{@{}rccccc@{}}
    \toprule
    & Left & Down & Right & Up &
    \\
    \cmidrule(r){2-2}\cmidrule(r){3-3}\cmidrule(r){4-4}\cmidrule(r){5-5}
    [0,0] & 0.25 & 0.25 & 0.25 & 0.25\\
    
    [0,1] & 0.25 & {\hl{0.143}} & 0.25 & 0.25\\
    
    \multicolumn{5}{c}{...}\\
    
    [1,0] & 0.25 & 0.25 & {\hl{0.143}} & 0.25\\
    
    \multicolumn{5}{c}{...}\\
    
    [1,2] & {\hl{0.143}} & 0.25 & 0.25 & 0.25\\
    
    \multicolumn{5}{c}{...}\\
    
    [2,1] & 0.25 & 0.25 & 0.25 & {\hl{0.143}}\\
    
    \multicolumn{5}{c}{...}\\
    
    [3,3] & 0.25 & 0.25 & 0.25 & 0.25\\
    \bottomrule
\end{tabular}
    \end{subtable}%
    \hfill
    \begin{subtable}{0.5\linewidth}
        \caption{After normalization}
        \label{tab:fused-policy-prob-normalized}
        \begin{tabular}{@{}rccccc@{}}
    \toprule
    & Left & Down & Right & Up &
    \\
    \cmidrule(r){2-2}\cmidrule(r){3-3}\cmidrule(r){4-4}\cmidrule(r){5-5}
    [0,0] & 0.25 & 0.25 & 0.25 & 0.25\\
    
    [0,1] & \hlgreen{0.28} & \hlred{0.16} & \hlgreen{0.28} & \hlgreen{0.28}\\
    
    \multicolumn{5}{c}{...}\\
    
    [1,0] & \hlgreen{0.28} & \hlgreen{0.28} & \hlred{0.16} & \hlgreen{0.28}\\
    
    \multicolumn{5}{c}{...}\\
    
    [1,2] & \hlred{0.16} & \hlgreen{0.28} & \hlgreen{0.28} & \hlgreen{0.28}\\
    
    \multicolumn{5}{c}{...}\\
    
    [2,1] & \hlgreen{0.28} & \hlgreen{0.28} & \hlgreen{0.28} & \hlred{0.16}\\
    
    \multicolumn{5}{c}{...}\\
    
    [3,3] & 0.25 & 0.25 & 0.25 & 0.25\\
    \bottomrule
\end{tabular}
   \end{subtable}
\end{table*}

\begin{table*}
    \captionsetup[subtable]{position=top}
    \captionsetup[table]{position=top}
    \setlength{\tabcolsep}{6pt}
    \caption{Shaped policy of the Frozen Lake running example}
    \label{tab:fused-policy-prob-full}
    \begin{subtable}{0.5\linewidth}
        \caption{Before normalization}
        \label{tab:fused-policy-prob-full-nonnormalized}
        \begin{tabular}{@{}rccccc@{}}
    \toprule
    & Left & Down & Right & Up &
    \\
    \cmidrule(r){2-2}\cmidrule(r){3-3}\cmidrule(r){4-4}\cmidrule(r){5-5}
    [0,0] & 0.25 & 0.25 & 0.25 & 0.25\\
    
    [0,1] & 0.25 & \hl{0.143} & 0.25 & 0.25\\

    [0,2] & 0.25 & 0.25 & \hl{0.250} & 0.25\\

    [0,3] & 0.25 & \hl{0.217} & 0.25 & 0.25\\
    
    \multicolumn{5}{c}{...}\\
    
    [1,0] & 0.25 & 0.25 & \hl{0.143} & 0.25\\
    
    \multicolumn{5}{c}{...}\\
    
    [1,2] & \hl{0.143} & 0.25 & \hl{0.217} & 0.25\\

    [1,3] & 0.25 & \hl{0.250} & 0.25 & 0.25\\
    
    \multicolumn{5}{c}{...}\\
    
    [2,1] & 0.25 & 0.25 & 0.25 & \hl{0.143}\\
    
    \multicolumn{5}{c}{...}\\

    [2,3] & 0.25 & \hl{0.400} & 0.25 & \hl{0.217}\\

    \multicolumn{5}{c}{...}\\

    [3,2] & 0.25 & 0.25 & \hl{0.400} & 0.25\\
    
    [3,3] & 0.25 & 0.25 & 0.25 & 0.25\\
    \bottomrule
\end{tabular}
    \end{subtable}%
    \hfill
    \setlength{\tabcolsep}{3.5pt}
    \begin{subtable}{0.5\linewidth}
        \caption{After normalization}
        \label{tab:fused-policy-prob-full-normalized}
        \begin{tabular}{@{}rccccc@{}}
    \toprule
    & Left & Down & Right & Up &
    \\
    \cmidrule(r){2-2}\cmidrule(r){3-3}\cmidrule(r){4-4}\cmidrule(r){5-5}
    [0,0] & 0.25 & 0.25 & 0.25 & 0.25\\
    
    [0,1] & \hlgreen{0.28} & \hlred{0.16} & \hlgreen{0.28} & \hlgreen{0.28}\\

    [0,2] & 0.25 & 0.25 & 0.25 & 0.25 \\

    [0,3] & \hlgreen{0.259} & \hlred{0.223} & \hlgreen{0.259} & \hlgreen{0.259}\\
    
    \multicolumn{5}{c}{...}\\
    
    [1,0] & \hlgreen{0.28} & \hlgreen{0.28} & \hlred{0.16} & \hlgreen{0.28}\\
    
    \multicolumn{5}{c}{...}\\
    
    [1,2] & \hlred{0.166} & \hlgreen{0.29} & \hlgreen{0.252} & \hlgreen{0.29}\\

    [1,3] & 0.25 & 0.25 & 0.25 & 0.25 \\
    
    \multicolumn{5}{c}{...}\\
    
    [2,1] & \hlgreen{0.28} & \hlgreen{0.28} & \hlgreen{0.28} & \hlred{0.16}\\

    \multicolumn{5}{c}{...}\\
    
    [2,3] & \hlred{0.224} & \hlgreen{0.358} & \hlred{0.224} & \hlred{0.194}\\
    
    \multicolumn{5}{c}{...}\\

    [3,2] & \hlred{0.217} & \hlred{0.217} & \hlgreen{0.348} & \hlred{0.217}\\
    
    [3,3] & 0.25 & 0.25 & 0.25 & 0.25\\
    \bottomrule
\end{tabular}
   \end{subtable}
\end{table*}

\begin{figure}[htb]
    \centering
    \includegraphics[width=0.33\linewidth]{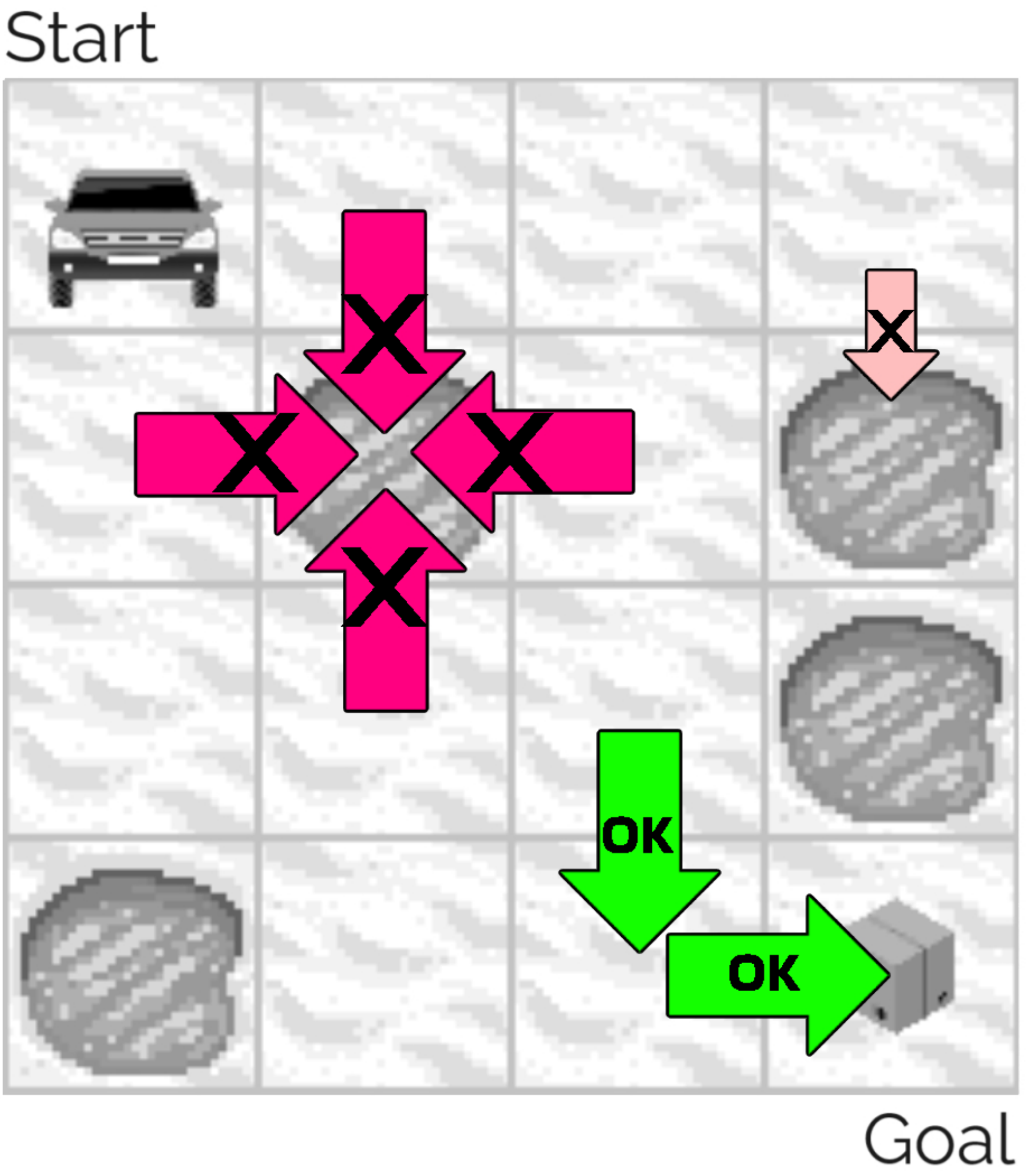}
    \caption{The result of policy shaping in the running example with the major changes highlighted. Red: decreased probability; green: increased probability.}
    \label{fig:frozenlake-shaped-policy}
\end{figure}

\clearpage


\subparagraph{Visualization of results}
The final policy in \tabref{tab:fused-policy-prob-full-normalized} is visualized in \figref{fig:frozenlake-shaped-policy}, with the major changes highlighted.
As seen, the negative advice about the hole in \cell{1}{1} resulted in a strong decrease in the corresponding probabilities of actions from neighboring cells, while the negative advice about the hole in \cell{1}{3} resulted in slightly smaller decrease due to higher uncertainty of the advice.
Conversely, the positive advice about the goal in \cell{3}{3} coupled with high certainty resulted in substantial positive bias that will guide the agent around the hole to the goal. Note that this policy is only the input to the agent's learning phase, and the preferences in each state have yet to be reinforced. For example, in state \cell{0}{2}, there is no change based on the advice, but surely, the agent will learn not to move down toward a hole but rather, to the right.

\subparagraph{Reflection}
The agent's policy is affected by the advice. Positive advice by opinions with belief being greater than base rate ($b > a$) causes the shaped policy to bias towards more belief to take actions that lead to the state that has been advised as a beneficial one. This bias manifests in an increased probability of the agent taking corresponding actions that lead to the advised state from its neighboring states. Such cases can be observed around the goal in the running example.
In contrast, negative advice ($b < a$) shapes the policy in a way that it drives the agent way from advised states by decreasing the probability of the agent taking corresponding actions that to the advised state from its neighboring states. Such cases can be observed around the holes in the running example.
The level of uncertainty has a profound impact on the shaped policy, as demonstrated by the difference between the weight of negative advice about state \cell{1}{1} and \cell{1}{3}.

\subsection*{Conclusion of RQ1}

In \secref{sec:introduction}, we posed our first research question: \textit{How can one use opinions to guide reinforcement learning agents?}
In this section, we presented a method for guiding RL agents through opinions.

\begin{conclusionframe}{RQ1}
Our method shows that through a combination of subjective logic, appropriate domain-specific languages, and a suitable distance function, opinions can be introduced into reinforcement learning as the means of guidance. The sound mathematical framework of subjective logic allows for multiple sources of opinions to be uniformly fuse with the agent's original policy, effectively achieving policy shaping through opinions.
Decoupling the calibration of uncertainty allows for more streamlined and intuitive advising by human experts.
\end{conclusionframe}
\section{Evaluation}\label{sec:evaluation}

In this section, we empirically evaluate our approach and investigate the effects of opinion-based human advice on the performance of reinforcement learning agents.

\subsection{Study design}

We conduct the study in a series of experiments on a sufficiently scaled-up version of the Frozen Lake running example (\secref{sec:example}). To find the most appropriate configuration of the running example that is challenging enough but still feasible for the unadvised agent, we first manually experiment with our framework (\secref{sec:manual-exp}), and based on our experiences we finalize the scenario (\secref{sec:scenario}) and the setup (\secref{sec:setup}).

\subsubsection{Manual experimentation}\label{sec:manual-exp}

Before we run extensive, long-running experiments, we first manually investigate the effect of map configuration on the minimum number of episodes required for the vanilla RL agent to learn appropriately. In general, the larger the map, the harder it is to explore it, and therefore, more episodes are required for the agent to start functioning appropriately. The placement of termination points (holes) is another factor that might increase the minimum number of required episodes.

To experiment with various map configurations in a reproducible way, we developed a map generator that is able to generate maps of arbitrary size and hole density and disperse holes on the map randomly, controlled by a user-defined seed. The map generator is available in the replication package.

Eventually, we choose a $12\times12$ map with 20\% hole density (which is the default density in OpenAI Gymnasium) under seed 63. The final map is shown in \appref{sec:appendix-studysetup}.
On this map, we observe that the behavior of the unadvised RL agent is appropriate after $5\,000$ episodes. Thus, we will choose a sufficiently higher number of episodes in our experiments, as explained below.

\subsubsection{Scenario}\label{sec:scenario}

The scenario is a scaled-up version of the running example.
We use the frozen lake environment, as described in detail in \secref{sec:example}.
The size of the lake in our experiments is 12$\times$12 with 20\% hole ratio (which is the default value in Gym's Frozen Lake environment\footnote{\url{https://gymnasium.farama.org/environments/toy\_text/frozen\_lake/\#arguments}}). We also ensure that the goal is reachable from the start in the map.

\paragraph{Map generation} To generate such a large map, we developed a map generator in Python that allows for generating maps of arbitrary size and hole ratio. Holes are distributed randomly on the lake. For replicability, the generator can be parameterized with a seed. The generator outputs an Excel file for human inspection and the standard graphical rendering of the generated environment. The Excel file is then used as input for each experiment execution.
The map generator is available in the replication package.

\subsubsection{Experimental setup}\label{sec:setup}

Each experiment is executed using the same setup as discussed below.

\paragraph{Parameters}

The list of parameters is shown in \tabref{tab:params}.

\begin{table}[h]
\centering
\caption{Parameters of the experiments}
\label{tab:params}
\begin{tabular}{@{}ll@{}}
\toprule
\textbf{Parameter} & \multicolumn{1}{c}{\textbf{Value}} \\\midrule

\multicolumn{2}{c}{\textbf{Fixed parameters}} \\

RL method & Discrete policy gradient \\
Number of episodes & 10\,000\ \\
Learning rate & 0.9\\
Discount factor & 1.0\\
Advice strategy & Once at the beginning of the experiment \\
Fusion operator & BCF\\
Grid size & 12\\
\midrule
\multicolumn{2}{c}{\textbf{Variables}} \\

Agent type & \{Random, Unadvised, Advised\}\\
Source of advice & \makecell[l]{\{Oracle, Single human,\\ Cooperating humans\}}\\
Advice quota -- oracle & \{100\% (``All''), 20\% (``Holes\&Goal'')\}\\
Advice quota -- single human & \{10\%, 5\%\}\\
Advice quota -- coop. human & \{10\% each, 5\% each\}\\
Uncertainty (oracle and single human) & \{$0.2k~|~k \in {0..4}$\}\\
Uncertainty (cooperative humans) & Calculated dynamically from a distance metric \\
Cooperative advice type & \{Sequential cooperation, parallel cooperation\}\\
\bottomrule
\end{tabular}
\end{table}

\subparagraph{Fixed parameters}
In this work, we choose discrete policy gradient as the \textbf{reinforcement learning method}. Policy-based methods represent the policy explicitly, which allows for directly investigating the effects of advice on the policy. 
To mitigate threats to validity, we use meaningful hyper parameters of the RL algorithm in order to allow the unadvised agent to perform as well as possible. Specifically, we set the learning rate to 0.9, and the discount factor to 1.0. 

Each experiment runs for a number of episodes, defined by the \textbf{number of episodes} parameter.
An episode starts with a reset environment and the agent in the initial position. Within the same experiment, the policy is updated and not reset. An episode lasts until the agent either finds the goal or steps into a hole. The higher the maximum number of episodes, the more opportunities agents have to find the goal and reinforce a beneficial policy.
After we observed in our manual experiments that the behavior of the unadvised RL agents is appropriate after 5\,000 episodes, we decided to run our experiments for a comfortable 10\,000 episodes.

The \textbf{advice strategy} stipulates advice to be provided \textit{once} and that it is done at the beginning of the experiment. Alternatively, more interactive advice strategies could be explored, but this is outside the scope of the current study. 

The \textbf{fusion operator} is also fixed. We used the Belief Constraint Fusion (BCF) operator for simplicity. BCF is an appropriate choice when agents have already made up their minds and will not seek compromise~\citep{josang2016subjective}. Since the source of the advice formulates their opinions about the map independently from the RL agent, it is fair to assume that compromise will not be sought.

Finally, we fix the \textbf{problem size} to a $12\times12$, as explained in \secref{sec:manual-exp}.

\phantom{}

\subparagraph{Variables}
We execute experiments for each \textbf{agent type} to compare their performance. The random agent randomly samples from the available actions. Unadvised agents maintain a policy. Advised agents maintain a policy, and before their first training episode, advice is fused into their initial policy.
In each case, the initial policy is the one in which the probability of choosing an action is uniformly distributed.

We conduct experiments with different \textbf{sources of advice}.
To form a ground truth, we first experiment with an idealized oracle with full information about the problem space. The oracle provides synthetic advice based on pre-defined rules as follows. If the cell is a goal, its value is +2. If the cell if a hole, the value is -2. If the cell has no neighboring holes, its value is +1. If the cell has one neighboring hole, its value is 0. If the cell has more than one neighboring hole, its value is -1.
In the \textit{single human} mode, a human advisor provides advice, and we modulate the uncertainty of their advice, similar to the oracle's advice. In the \textit{cooperative} mode, advice is provided by two human advisors who have \textit{partial} and \textit{complementary} information about the problem. Thus, this experimental mode is highly realistic. The human advisor labels hole and goal cells similarly to the oracle, but for other cells, may use values $\pm1, 0$  differently than the oracle.

We use different \textbf{advice quota}s for different advisor types.
We evaluate the oracle by two advice quotas: first, by providing the agent with advice about \textit{all} the cells, i.e., 100\% quota; second, by providing the agent with advice about the \textit{holes and the goal} on the map, which amounts to about 20\% of cells (see \secref{sec:scenario}). 
We evaluate the single human advisor by two advice quotas: first, by providing the agent with advice about \textit{10\%} of the cells; second, by providing the agent with advice about \textit{5\%} of the cells.
We evaluate the cooperating human advisors by two human advice quotas: first, by allowing \textit{each advisor} \textit{10\%} advice quota (thus, 20\% in total); second, by allowing \textit{each advisor} \textit{5\%} advice quota (thus, 10\% in total).

We synthetically modulate the \textbf{degree of uncertainty} of the advice of the oracle and the single human by sweeping through the [0.0, 1.0)  interval in 0.2 increments. Here, $u=0.0$ reduces to classic probability, and $u=1$ represents a uniform distribution~\citep{josang2016subjective}. This choice allows us to directly compare the oracle with the single human advisor at different levels of uncertainty.

In the cooperative mode, \textbf{uncertainty} is calculated from the distance between the advisor and the location of the advice, in accordance with \secref{sec:approach-u}.
%
We use the two dimensional Manhattan distance, defined as follows.
\begin{align*}
    D((x_1, y_1), (x_2, y_2)) = |x_1-x_2|+|y_1-y_2|.
\end{align*}
Here, $(x_1,y_1)$ is the location of the advisor.
We assume a linear certainty discount function with $\tau=1$, as explained in \equref{eq:discount}. That is, uncertainty of an advisor in one of the corners of the grid reaches its maximum in the opposing corner. Thus, uncertainty at any point $(x_2, y_2)$ is defined as follows.
\begin{align*}
    u (x_2,y_2) =  D((x_1, y_1), (x_2, y_2)) \cdot \frac{1}{D((x_1, y_1), (x_n, y_n))}.
\end{align*}




Finally, we define two types of \textbf{cooperative advice with partial information}. In \textit{sequential cooperation}, one human advises the agent in the first half of its mission, and subsequently, the other human advises the agent in the second half of its mission. To achieve this, we place the first human advisor in the top-left corner of the grid (start), and the second human advisor in the bottom-right corner of the grid (goal). As the agent follows the main diagonal of the grid world, the first advisor's input will gradually lose its influence due to the distance explained above; and the second advisor's input will gradually gain more influence.
In \textit{parallel collaboration}, both humans provide advice throughout the agent's entire mission. To achieve this, we place one human advisor in the top-right corner of the grid, and the other human advisor in the bottom-left corner of the grid. By that, both advisors are able to provide advice throughout the entire start-goal trajectory of the agent. However, due to the distance metric defined above, both advisors will focus on their half of the grid world, i.e., the triangle under and above the diagonal, respectively. Clearly, this advice mode is the most \textit{realistic one} of the three as in real settings, uncertainty pertains to individual pieces of advice, and cooperation among stakeholders and experts is the usual way to solve complex problems.

%

\paragraph{Performance metric and analysis methods}

Following community best practices, we use the cumulative reward as the metric of performance, i.e., the total reward $R_T = \sum_{i=0}^{T} r_{t+1}$ collected by the agent throughout an episode.
We observe the amount of cumulative reward and its dynamics, i.e., how early and rapidly the cumulative reward increases under different settings.

\paragraph{Execution}

We execute experiments using each combination of parameters defined above. We run each experiment 30 times to achieve sufficient statistical power. The results of experiments are saved as csv files. These data files are available in the replication package.

We run the experiments on standard office equipment with the following parameters:
Apple Mac mini, equipped with a 3.6 GHz Quad-Core Intel Core i3 CPU, running Mac OS Sonoma 14.14.1 with 8 GB 2667 MHz DDR4 of memory, using Python 3.11.5. This setup is satisfactory for our purposes as we are not interested in runtime performance.

\subsection{Results}

In the following, we report the results of our experiments by reviewing the performance of guidance by the oracle (\secref{sec:results-synthetic}), single human advisor (\secref{sec:results-single-human}), and cooperative human advisors (\secref{sec:results-coop-human}). In each case, we first investigate the performance in terms of the cumulative reward, and then we look at the final policy.


\subparagraph{Main takeaways}

\begin{figure}[H]
    \centering

    \caption*{Linear scale}
    
    \begin{subfigure}{0.16\linewidth}
        \includegraphics[width=\linewidth]{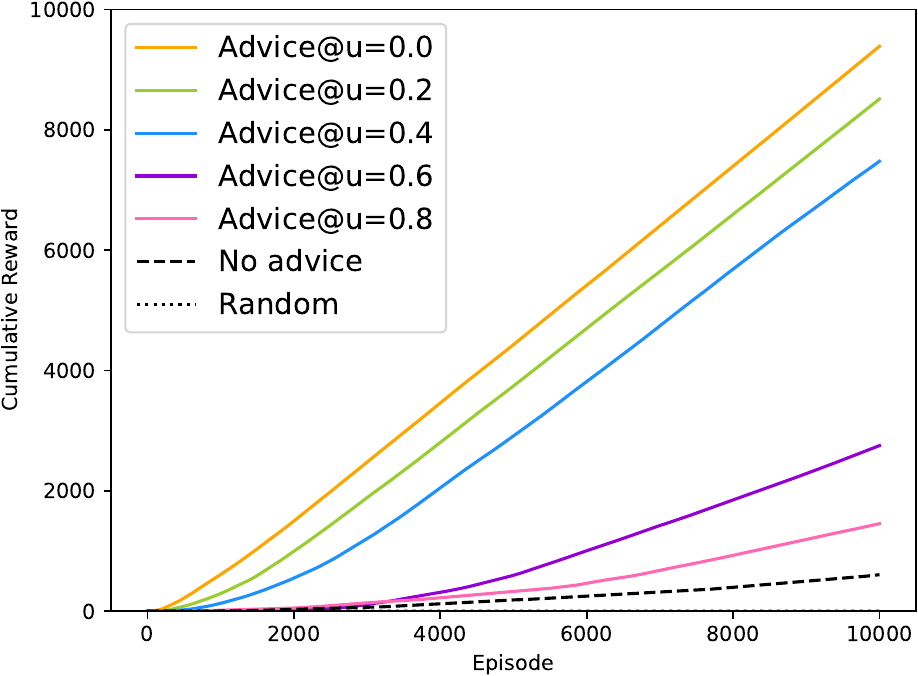}
        \caption{Oracle 100\%}
        \label{fig:results-cumulativereward-synthetic-all}
    \end{subfigure}
    \hfil
    \begin{subfigure}{0.16\linewidth}
        \includegraphics[width=\linewidth]{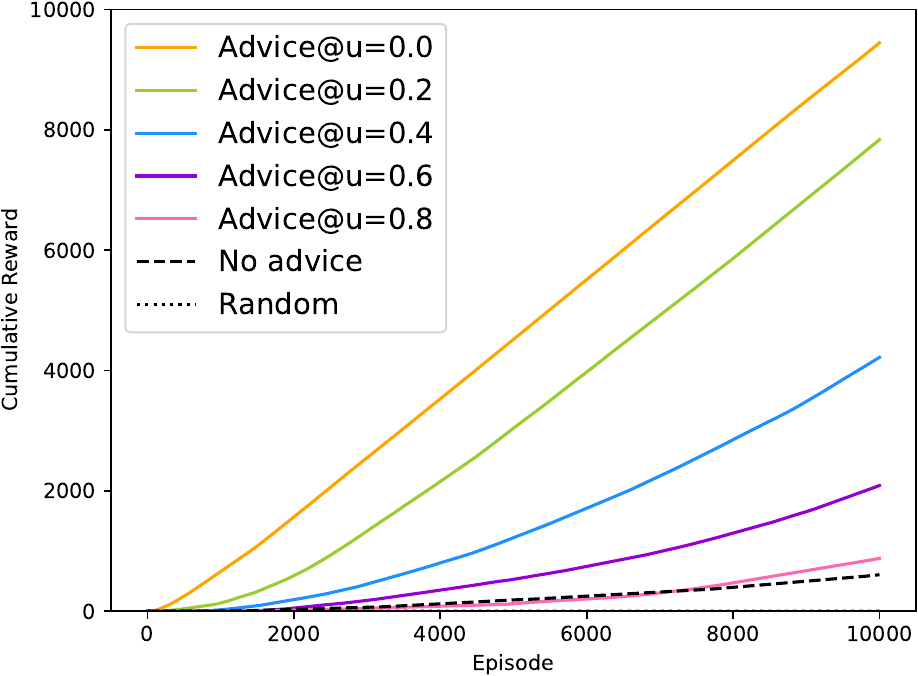}
        \caption{Oracle 20\%}
        \label{fig:results-cumulativereward-synthetic-holes}
    \end{subfigure}
    \hfil
    \begin{subfigure}{0.16\linewidth}
        \includegraphics[width=\linewidth]{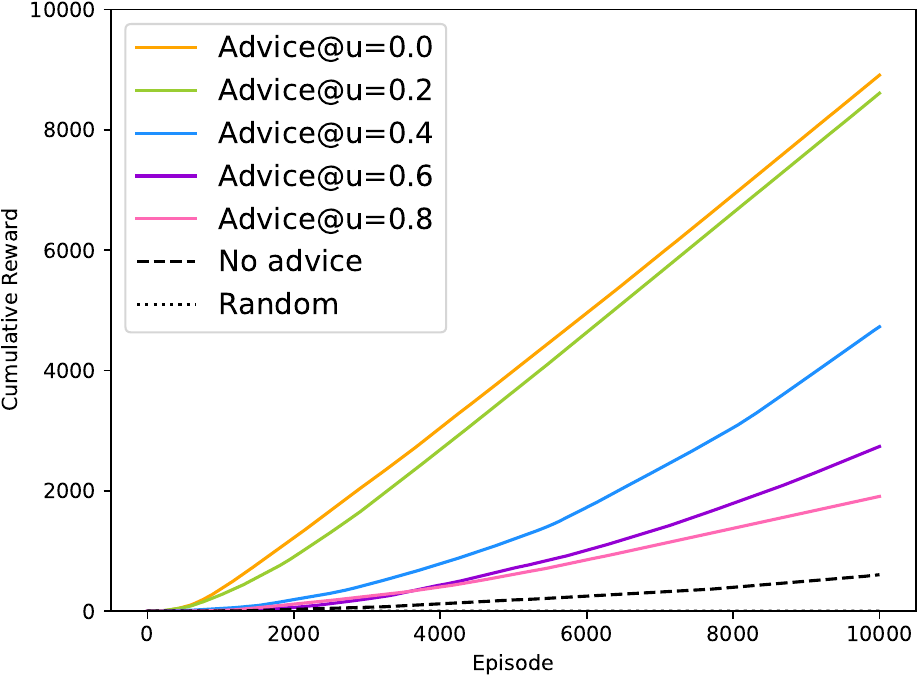}
        \caption{Human 10\%}
    \end{subfigure}
    \hfil
    \begin{subfigure}{0.16\linewidth}
        \includegraphics[width=\linewidth]{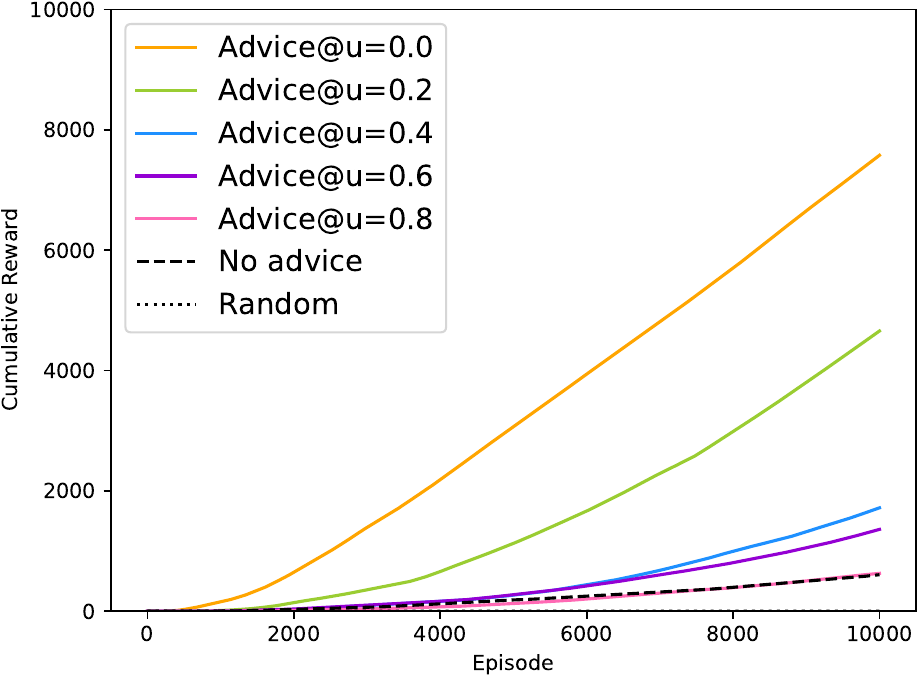}
        \caption{Human 5\%}
    \end{subfigure}
    \hfil
    \begin{subfigure}{0.16\linewidth}
        \includegraphics[width=\linewidth]{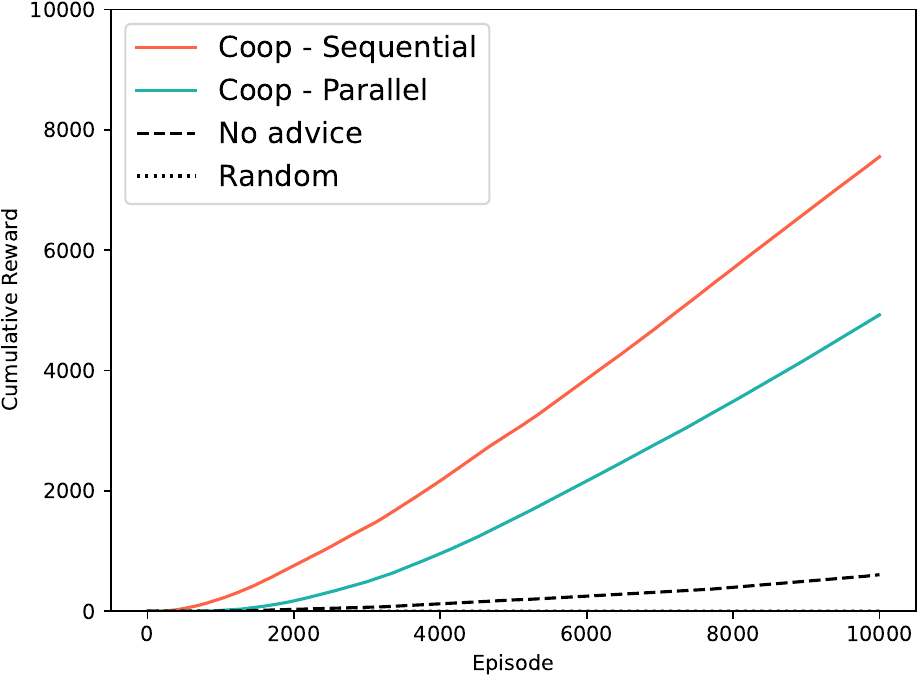}
        \caption{Coop $10\%$}
        \label{fig:results-cumulativereward-synthetic-all}
    \end{subfigure}
    \hfil
    \begin{subfigure}{0.16\linewidth}
        \includegraphics[width=\linewidth]{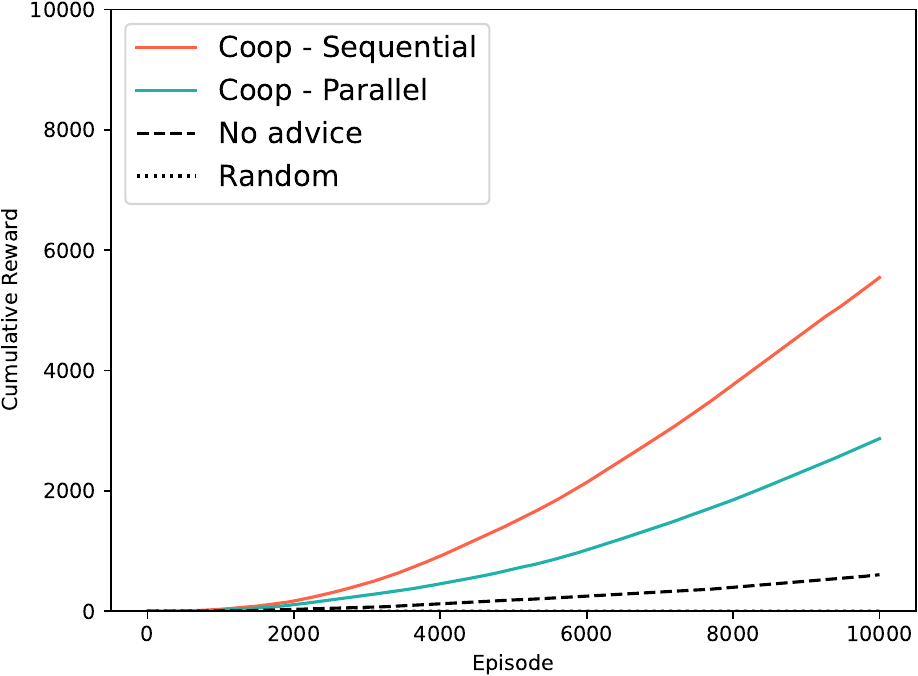}
        \caption{Coop $5\%$}
        \label{fig:results-cumulativereward-synthetic-holes}
    \end{subfigure}

    \caption*{Log scale}

    \begin{subfigure}{0.16\linewidth}
        \includegraphics[width=\linewidth]{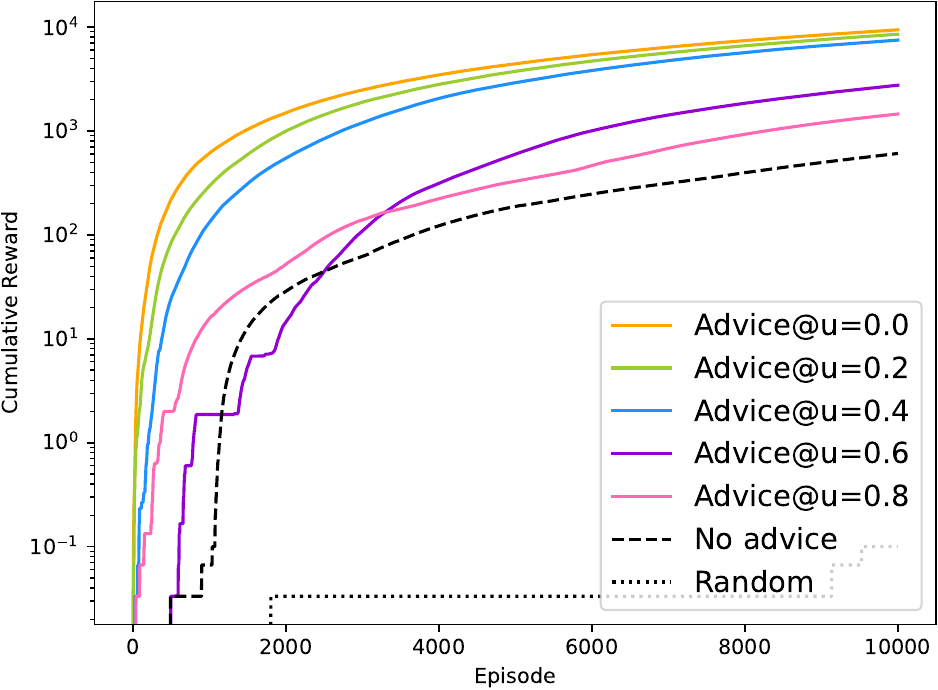}
        \caption{Oracle 100\%}
    \end{subfigure}
    \hfil
    \begin{subfigure}{0.16\linewidth}
        \includegraphics[width=\linewidth]{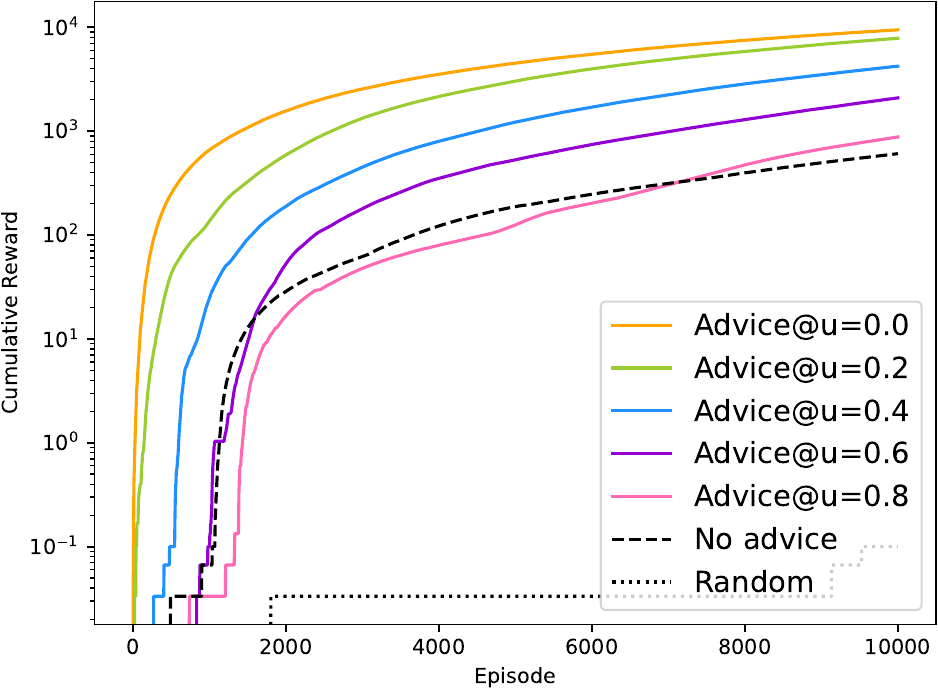}
        \caption{Oracle 20\%}
    \end{subfigure}
    \hfil
    \begin{subfigure}{0.16\linewidth}
        \includegraphics[width=\linewidth]{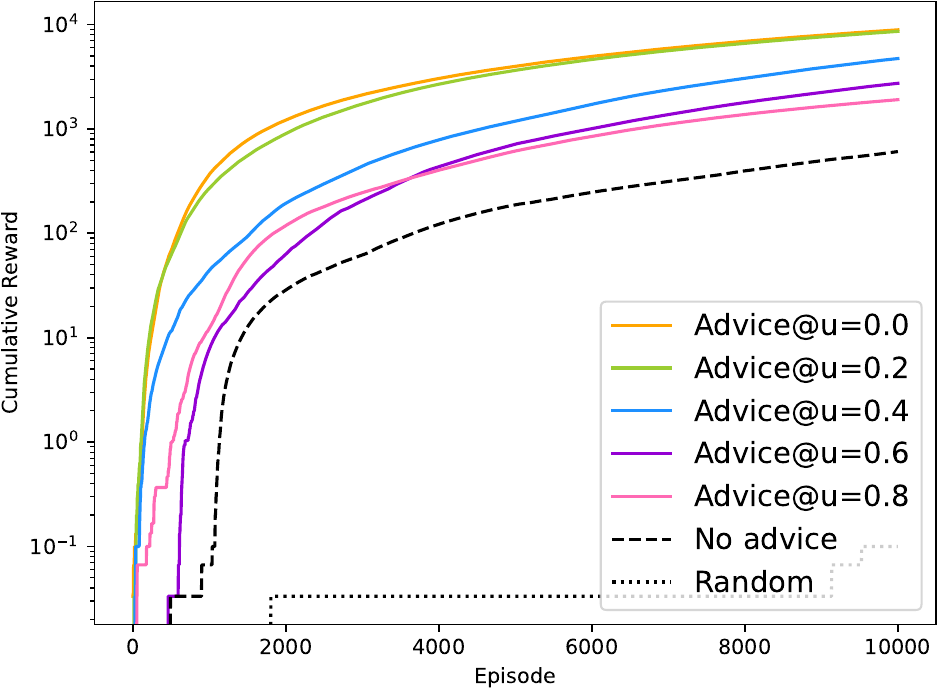}
        \caption{Human 10\%}
    \end{subfigure}
    \hfil
    \begin{subfigure}{0.16\linewidth}
        \includegraphics[width=\linewidth]{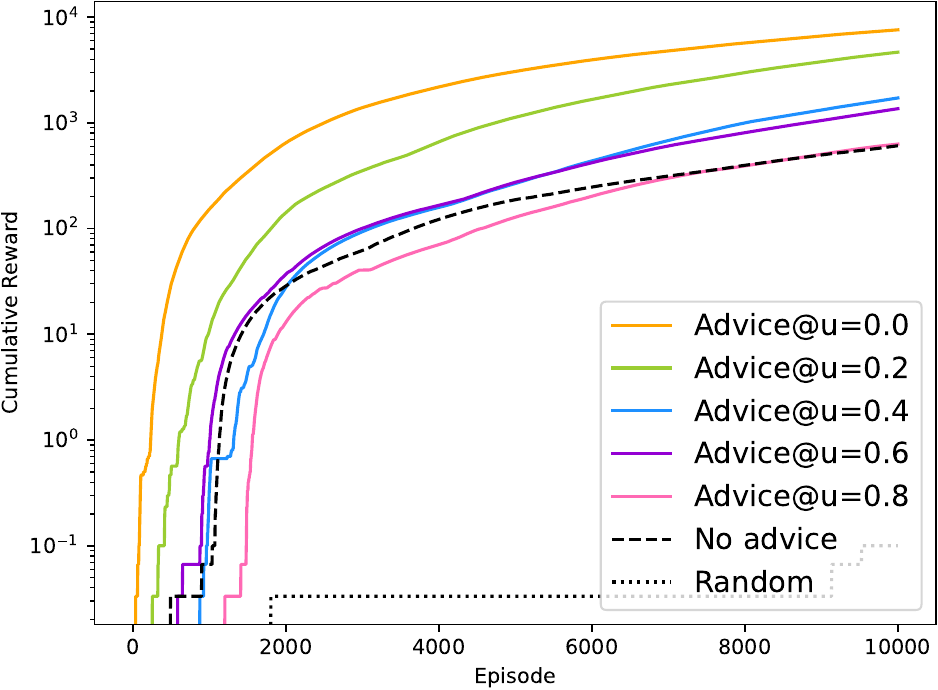}
        \caption{Human 5\%}
    \end{subfigure}
    \hfil
    \begin{subfigure}{0.16\linewidth}
        \includegraphics[width=\linewidth]{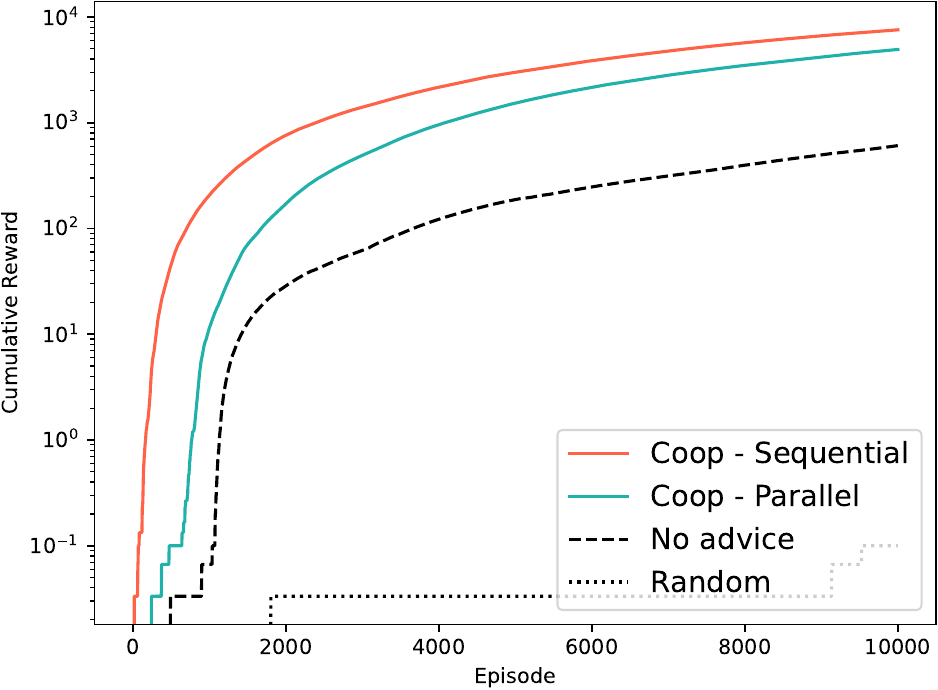}
        \caption{Coop: $10\%$}
        \label{fig:results-cumulativereward-synthetic-all-log}
    \end{subfigure}
    \hfil
    \begin{subfigure}{0.16\linewidth}
        \includegraphics[width=\linewidth]{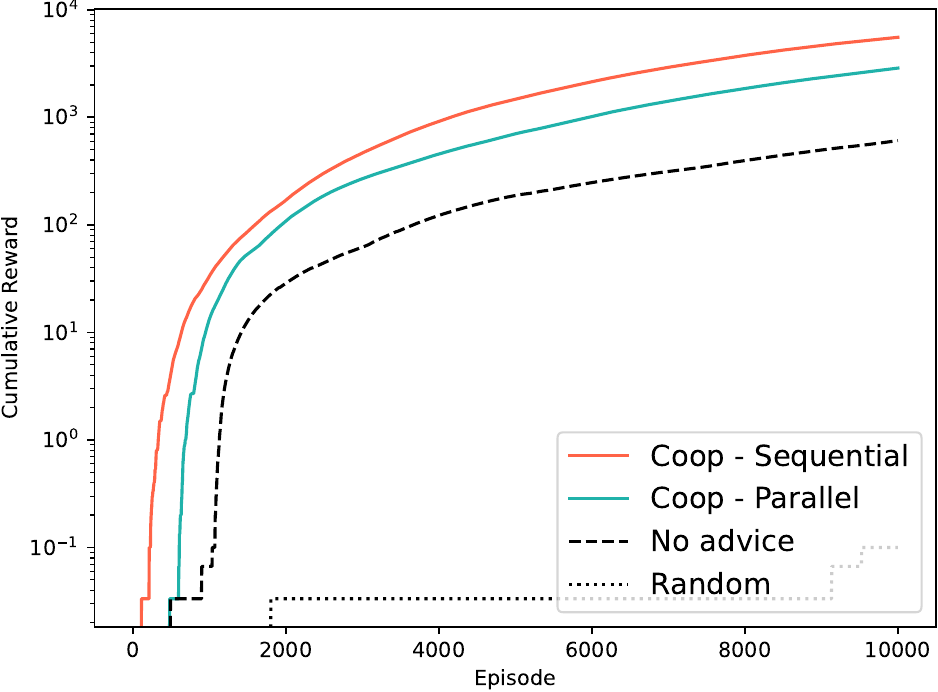}
        \caption{Coop $5\%$}
        \label{fig:results-cumulativereward-synthetic-holes-log}
    \end{subfigure}
    
    \caption{Performance comparison of every experiment in terms of cumulative reward}
    \label{fig:cumulative-rewards-all}
\end{figure}

\tabref{tab:cumulative-rewards} and \figref{fig:cumulative-rewards-all} report the cumulative rewards in different evaluation modes and highlights the key takeaways of our work.

We note that every advised agent performs better than the unadvised agent (607.267 mean cumulative reward), even at high levels of uncertainty.

We observe comparable results between the fully informed (100\% quota) oracle and the 10\% quota single human case at low levels of uncertainty. In fact, at $u=0.2$, the human advice with 10\% quota outperforms the oracle, even with 100\% quota (8\,607.500 vs 8\,511.367).

Finally, to compare cooperative performance with other experiments, we show the results of the four realistic cooperative experiments aligned with the comparable oracle or single human experiments with the same advice quota. For example, sequential cooperation with 2$\times$10\% quota is comparable to the oracle with 20\% quota. In this comparison, performances align best at $u=0.2$, where cooperating humans with partial information perform nearly as well as the oracle with complete information about the problem space (7\,551.567 vs 7\,835.267).
In general, sequential cooperation outperforms parallel cooperation; and parallel cooperation at 5\% outperforms the fully-informed single human advisor at 10\%.

\begin{table}[!h]
\centering
\captionsetup{width=\textwidth}
\caption{Cumulative rewards by experiment. Bold is best at the given level of certainty. Unadvised = 607.267. Random = 0.100. Cooperative modes aligned with the best matching uncertainty level of the comparable oracle or single human mode.}
\label{tab:cumulative-rewards}
\begin{tabular}{@{}lrrrrcrrrr@{}}
\cmidrule[\heavyrulewidth]{1-5}\cmidrule[\heavyrulewidth]{7-10}
 & \multicolumn{2}{c}{Oracle} & \multicolumn{2}{c}{Single human} & & \multicolumn{2}{c}{Coop. -- Sequential} & \multicolumn{2}{c}{Coop. -- Parallel} \\ \cmidrule(r){2-3}\cmidrule(r){4-5}\cmidrule(r){7-8}\cmidrule{9-10}
 u & \multicolumn{1}{c}{100\% (All)} & \multicolumn{1}{c}{20\% (H\&G)} & \multicolumn{1}{c}{10\%} & \multicolumn{1}{c}{5\%} & & \multicolumn{1}{c}{10\%} & \multicolumn{1}{c}{5\%} & \multicolumn{1}{c}{10\%} & \multicolumn{1}{c}{5\%} \\ \cmidrule(r){1-1} \cmidrule(r){2-2}\cmidrule(r){3-3} \cmidrule(r){4-4}\cmidrule(r){5-5} \cmidrule(r){7-7}\cmidrule(r){8-8}\cmidrule(r){9-9}\cmidrule{10-10}
0.0  & 9\,386.467 & \textbf{9\,443.733} & 8\,907.733 & 7\,576.067 && &&& \\
0.2  & 8\,511.367 & 7\,835.267 & \textbf{8\,607.500} & 4\,656.633 && 7\,551.567 &&& \\
0.4  & \textbf{7\,476.633} & 4\,218.000 & 4\,727.433 & 1\,718.300 &&& 5\,544.100 & 4\,924.967 & \\
0.6  & \textbf{2\,751.833} & 2\,089.433 & 2\,737.600 & 1\,360.267 &&&&& 2\,867.633 \\
0.8  & 1\,454.400 & 878.467 & \textbf{1\,907.933} & 629.367 &&&&& \\
\cmidrule[\heavyrulewidth]{1-5}\cmidrule[\heavyrulewidth]{7-10}
\end{tabular}
\end{table}

In the following, we elaborate on the experimental results in detail.

\subsubsection{Performance of guidance by an oracle}\label{sec:results-synthetic}


\subparagraph{Advisor type} In this experiment, we use an oracle, i.e., synthetic advice that assigns values to states based on pre-defined rules (as explained in \secref{sec:setup}).
\subparagraph{Advice quota} We evaluate two advice quotas: 100\% (the agent is provided with advice about \textit{all} the cells by an idealized oracle) and 20\% (the agent is provided with advice about the terminating states, i.e., the \textit{holes and goal}).
\subparagraph{Uncertainty calibration} We execute each experiment at multiple characteristic levels of uncertainty, sweeping through the \{0.0, 0.2, 0.4, 0.6, 0.8\} range. The case of complete uncertainty $u=1.0$ is not tested as these opinions carry no information that can be used by the agent.\footnote{It can also be shown that in the fusion operator of choice in this work (BCF), $u=1.0$ results in an idempotent transformation, which follows trivially from Equations \ref{eq:harmony}--\ref{eq:a-joint}. The proof is left as an exercise for the reader.}

\paragraph{Cumulative reward}

\begin{figure}
    \centering

	\caption*{Linear scale}
    
    \begin{subfigure}{0.48\linewidth}
        \includegraphics[width=\linewidth]{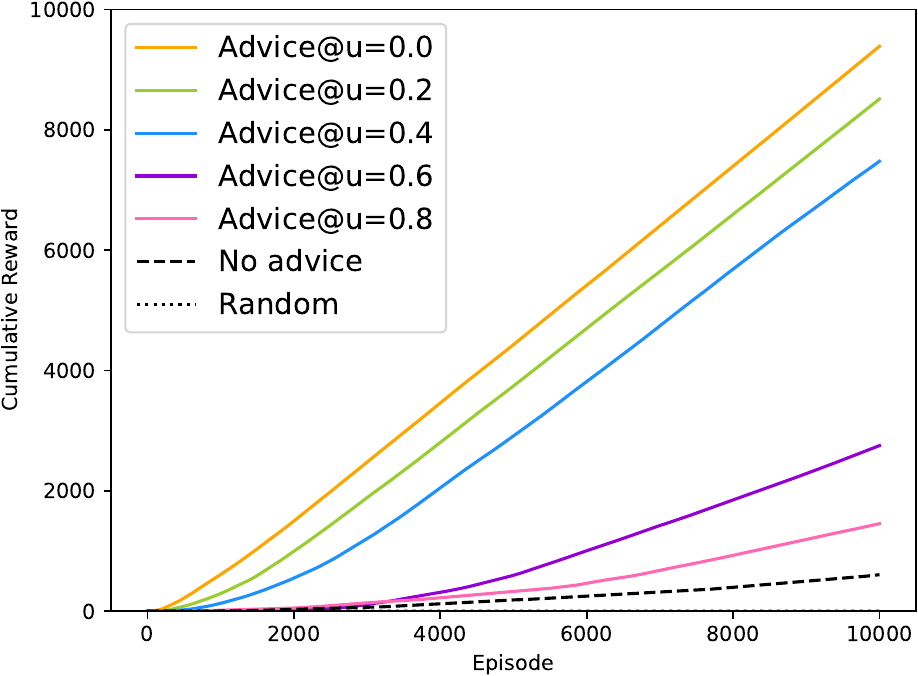}
		\caption{Advice quota: 100\% (All)}
        \label{fig:results-cumulativereward-synthetic-all}
    \end{subfigure}
    \hfil
    \begin{subfigure}{0.48\linewidth}
        \includegraphics[width=\linewidth]{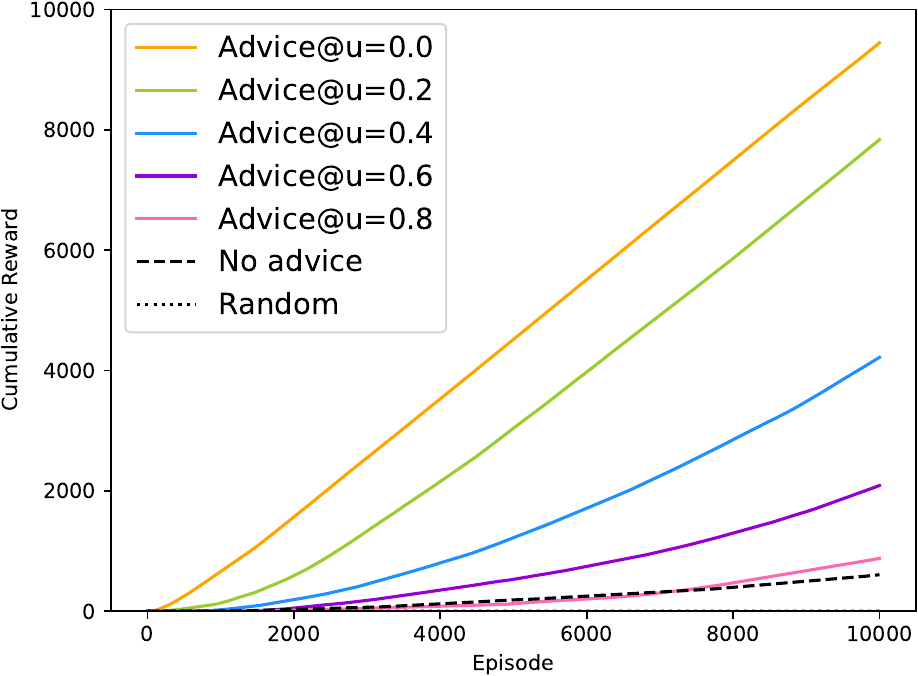}
        \caption{Advice quota: 20\% (Holes\&Goal)}
        \label{fig:results-cumulativereward-synthetic-holes}
    \end{subfigure}

    \caption*{Log scale}

    \begin{subfigure}{0.48\linewidth}
        \includegraphics[width=\linewidth]{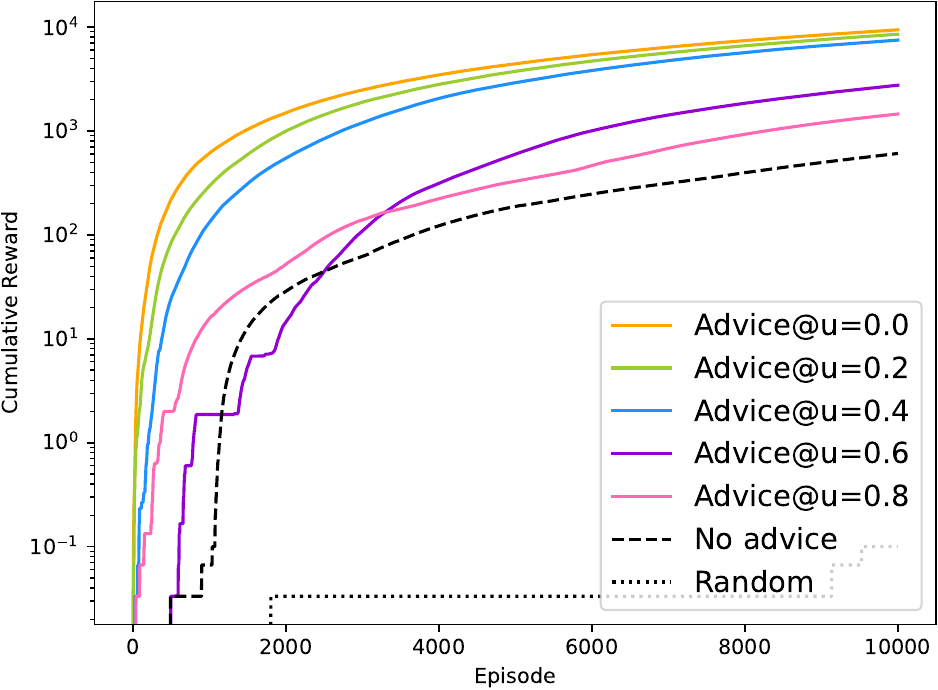}
        \caption{Advice quota: 100\% (All)}
    \end{subfigure}
    \hfil
    \begin{subfigure}{0.48\linewidth}
        \includegraphics[width=\linewidth]{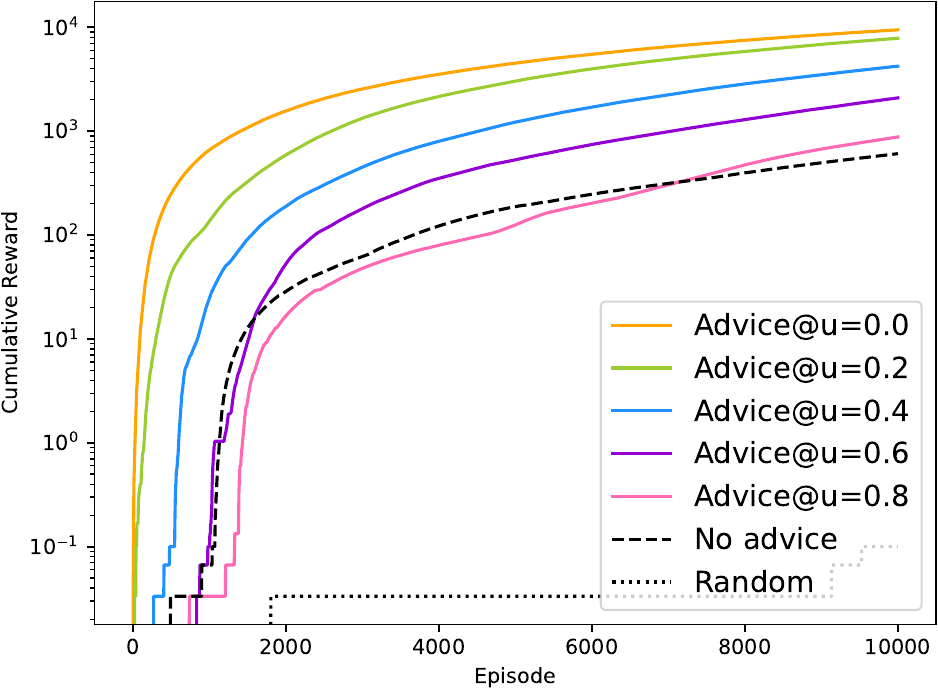}
        \caption{Advice quota: 20\% (Holes\&Goal)}
    \end{subfigure}
    
    \caption{Cumulative reward with an oracle as the advisor}
    \label{fig:results-cumulativereward-synthetic}
\end{figure}

\figref{fig:results-cumulativereward-synthetic} shows the cumulative rewards of the agent with synthetic advice from the oracle under the two advice quotas.

\subparagraph{Clear performance improvement in advised agents} As the most important takeaways, we observe that (i) advised agents clearly outperform the unadvised agent and that (ii) performance improves with the level of certainty of the advice. This can be seen from the lines corresponding to lower uncertainty consistently being over lines corresponding to higher uncertainty.
As expected, the random agent hardly accumulates any reward as it follows a random trajectory without learning.

Improved performance manifests in two forms. First, the cumulative reward is higher and increases at a consistently higher pace in advised and certain agents, as indicated by the advised lines over others in the linear and log scales, respectively. Second, cumulative reward increases earlier and reaches a gradual slope sooner in advised and certain agents than in the unadvised agent, indicated by the earlier rise of advised lines in the log scale charts.

\subparagraph{Even substantially uncertain advice makes a difference} Demonstrated in moderate ($u=0.6$) and high ($u=0.8$) uncertainty cases, advised agents still outperform the unadvised agent. The 100\% advice quota compensates for low certainty and renders the advised agent in the $u=0.8$ case more performant than the unadvised agent. This advantage is lost at 20\% quota as shown in \figref{fig:results-cumulativereward-synthetic-holes} by the similarity of the 0.8 and ``No advice'' lines.

\subparagraph{Learning dynamics: more rapid learning} 
The log scale charts further show the two main phases of the learning process. In the first phase, approximately between episodes 0--2\,000, we see a steep increase, indicating fast learning. The learning phase is substantially steeper as uncertainty decreases. In the second phase, approximately after episode 2\,000, the slopes become more gradual, indicating a slowed-down collection of reward and a likely convergence to the maximum potential performance. Performance differences remain consistent throughout the experiment.
The effect of advice, thus, is shifting the exponentially growing initial phase earlier in the learning process.

\subparagraph{Advice quotas: difference shows only in low-certainty cases}
There is not much difference between the two advice quotas in high-certainty cases, as shown by the very similar lines of $u=0.0$ and $u=0.2$ in \figref{fig:results-cumulativereward-synthetic-all} and \figref{fig:results-cumulativereward-synthetic-holes}. In fact, as shown in \tabref{tab:cumulative-rewards}, the oracle with 20\% quota slightly outperforms the oracle with 100\% quota at $u=0.0$. (+0.61\%) The most noticeable performance deterioration is observed at $u = 0.4$, when performance drops by 43.6\% between the 100\% quota and 20\% quota.

\paragraph{Effects on the policy}

\begin{figure}[H]
    \centering

    \begin{subfigure}[c]{0.32\linewidth}
        \includegraphics[width=\linewidth]{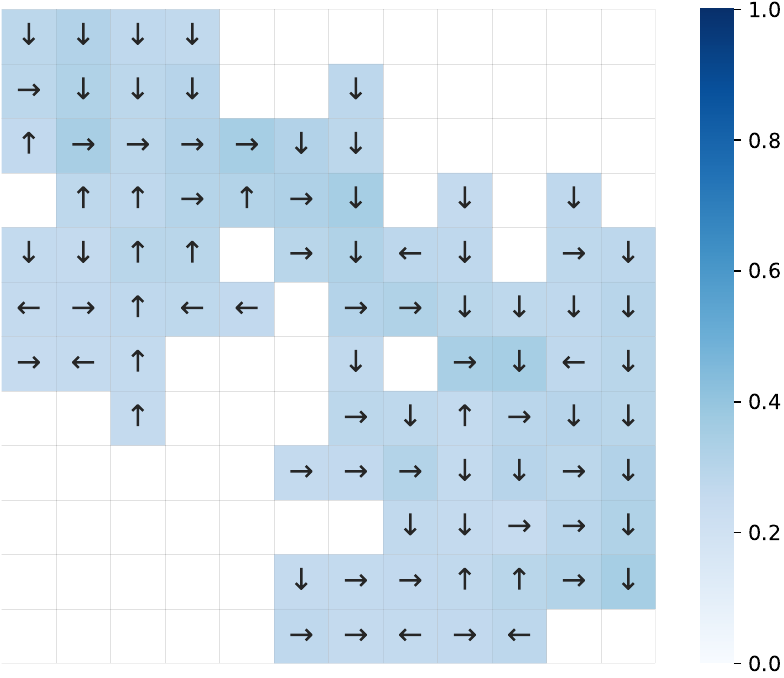}
        \caption{No advice}
    \end{subfigure}
    \hfil
    \begin{subfigure}[c]{0.64\linewidth}
        \caption*{Advice quota: 100\% (\texttt{ALL})}
        
        \begin{subfigure}{0.49\linewidth}
            \includegraphics[width=\linewidth]{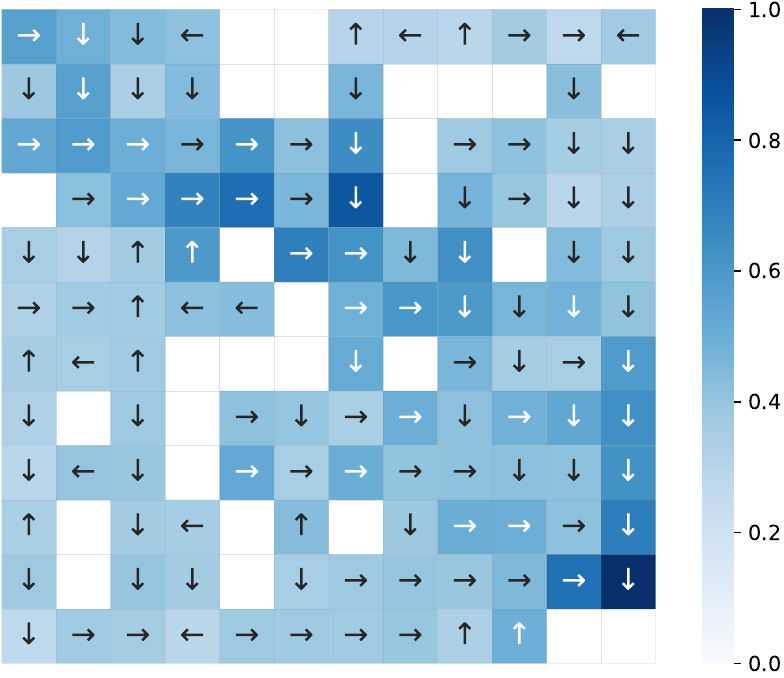}
            \caption{$u=0.4$}
            \label{fig:heatmap-synthetic-all-04}
        \end{subfigure}
        \hfil
        \begin{subfigure}{0.49\linewidth}
            \includegraphics[width=\linewidth]{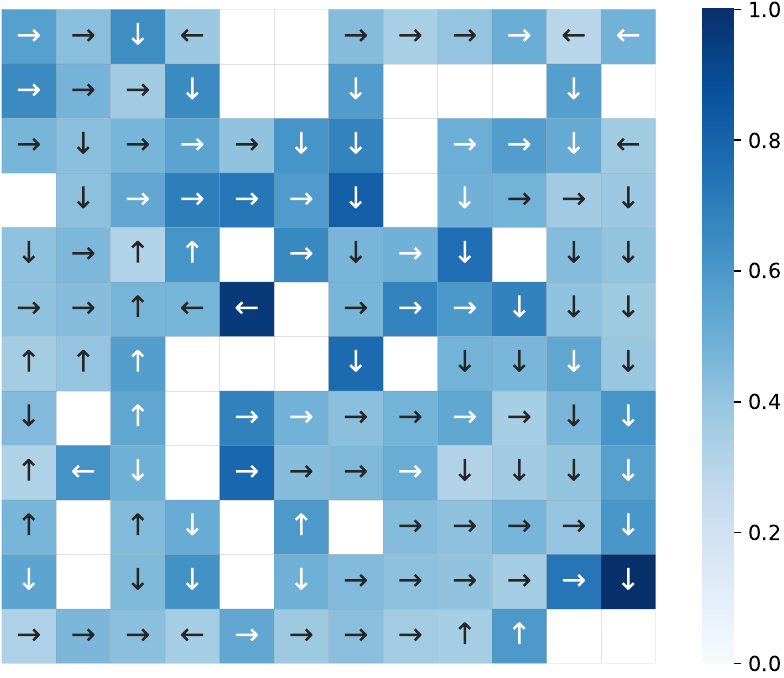}
            \caption{$u=0.0$}
            \label{fig:heatmap-synthetic-all-00}
        \end{subfigure}
    
        \caption*{Advice quota: 20\% (\texttt{HOLES\&GOAL})}
    
        \begin{subfigure}{0.49\linewidth}
            \includegraphics[width=\linewidth]{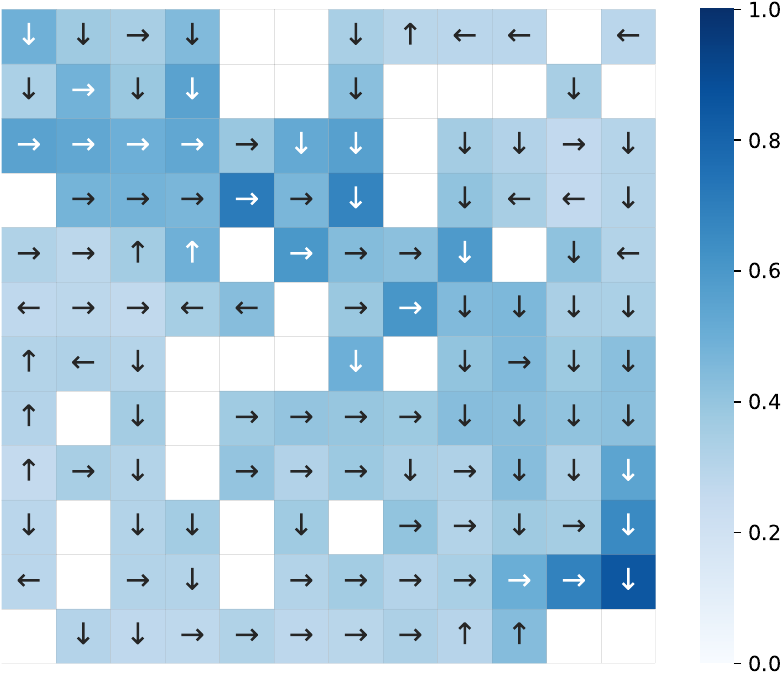}
            \caption{$u=0.4$}
            \label{fig:heatmap-synthetic-holes-04}
        \end{subfigure}
        \hfil
        \begin{subfigure}{0.49\linewidth}
            \includegraphics[width=\linewidth]{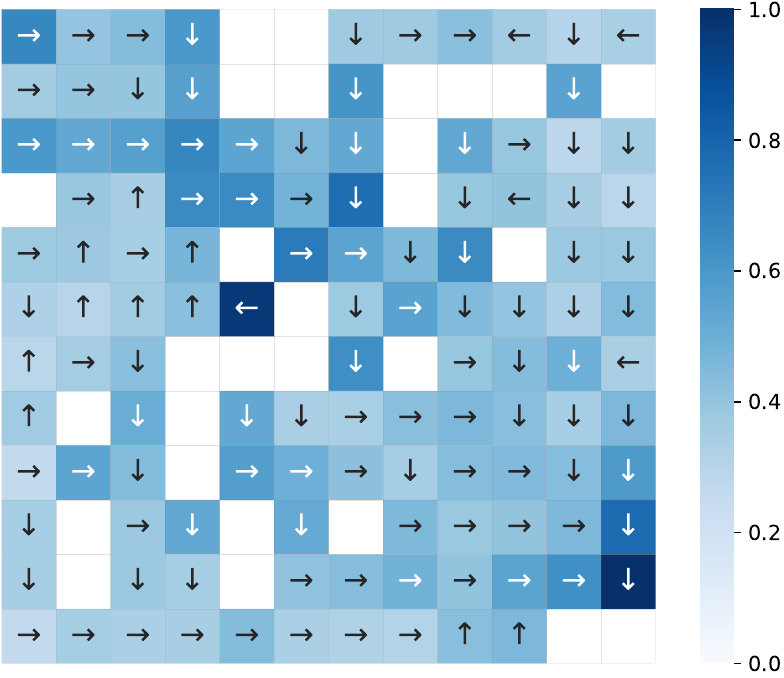}
            \caption{$u=0.0$}
            \label{fig:heatmap-synthetic-holes-00}
        \end{subfigure}
    \end{subfigure}
    
    \caption{Heatmap visualization of the final learned policy with an oracle as the advisor}
    \label{fig:results-heatmaps-synthetic-2x2}
\end{figure}

\figref{fig:results-heatmaps-synthetic-2x2} visualizes the policy as a heatmap. The cells of each heatmap correspond to the cells of the Frozen Lake problem; arrows represent the action with the highest probability; and shading corresponds to the probability. Empty cells show unexplored states.

\subparagraph{Better reinforced policy} One of the two key improvements is indicated by the darker shades of cells compared to the unadvised agent. Darker shades correspond to higher probabilities of the indicated action in a cell and an overall better reinforced beneficial policy. By providing advice, the agent navigated through the state space more efficiently and had time (episodes) to explore crucial cells, e.g., the ones around holes and the goal. The agent with 20\% advice quota reinforces some key decisions better than the agent with 100\% advice quota, e.g., around the goal and around the starting point.

\subparagraph{More thorough exploration} The other key improvement is shown by the higher number of cells explored by advised agents, as both advice quotas result in more cells being covered, compared to the unadvised agent.
The 20\% advice quota allows to explore more states at higher levels of certainty than the 100\% advice quota (cf. \figref{fig:heatmap-synthetic-holes-00} and \figref{fig:heatmap-synthetic-all-00}); however, this relation turns around in the lower-certainty case (cf. \figref{fig:heatmap-synthetic-all-04} and \figref{fig:heatmap-synthetic-holes-04}).

\subparagraph{Effects clear even at relatively low certainty} At $u=0.4$, we still see demonstrated improvements compared to the unadvised agent. That is, even fairly uncertain advice has a clear positive impact on the policy.

\subsubsection{Performance of guidance by a single human advisor}\label{sec:results-single-human}

\subparagraph{Advisor type} In this experiment, we use the advice of a single human advisor.
\subparagraph{Advice quota} We evaluate two advice quotas. In the first case, the human provides advice about 10\% of the cells at their discretion. In the second case, the human provides advice about 5\% of the cells at their discretion.
\subparagraph{Uncertainty calibration} We execute each experiment at multiple characteristic levels of uncertainty, sweeping through the \{0.0, 0.2, 0.4, 0.6, 0.8\} range. The case of complete uncertainty $u=1.0$ is not tested as these opinions carry no information that can be used by the agent.

\paragraph{Cumulative reward}

\begin{figure}[H]
    \centering

    \caption*{Linear scale}
    
    \begin{subfigure}{0.48\linewidth}
        \includegraphics[width=\linewidth]{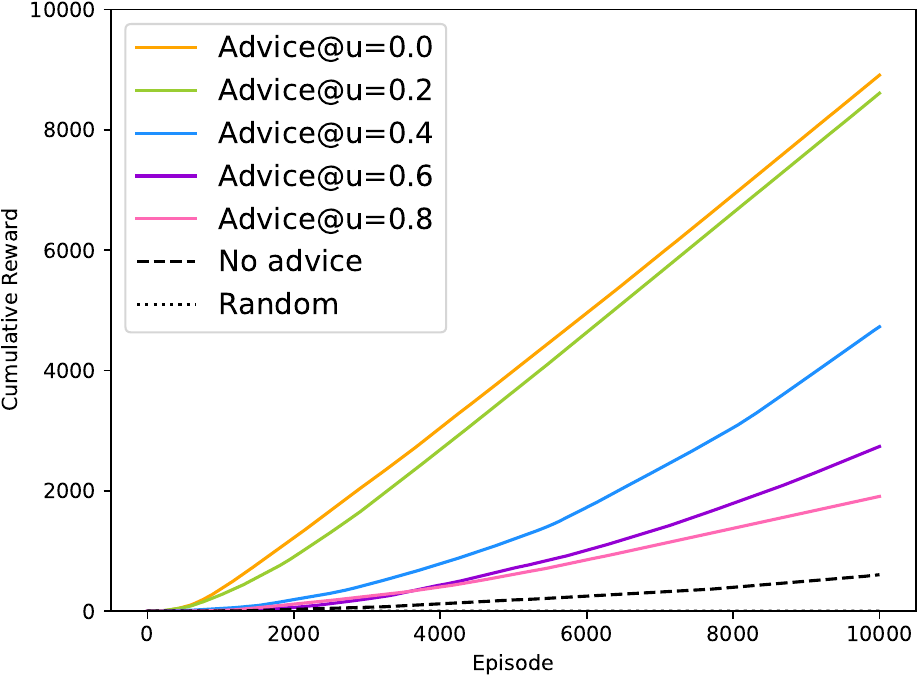}
		\caption{Advice quota: 10\%}
    \end{subfigure}
    \hfil
	\begin{subfigure}{0.48\linewidth}
        \includegraphics[width=\linewidth]{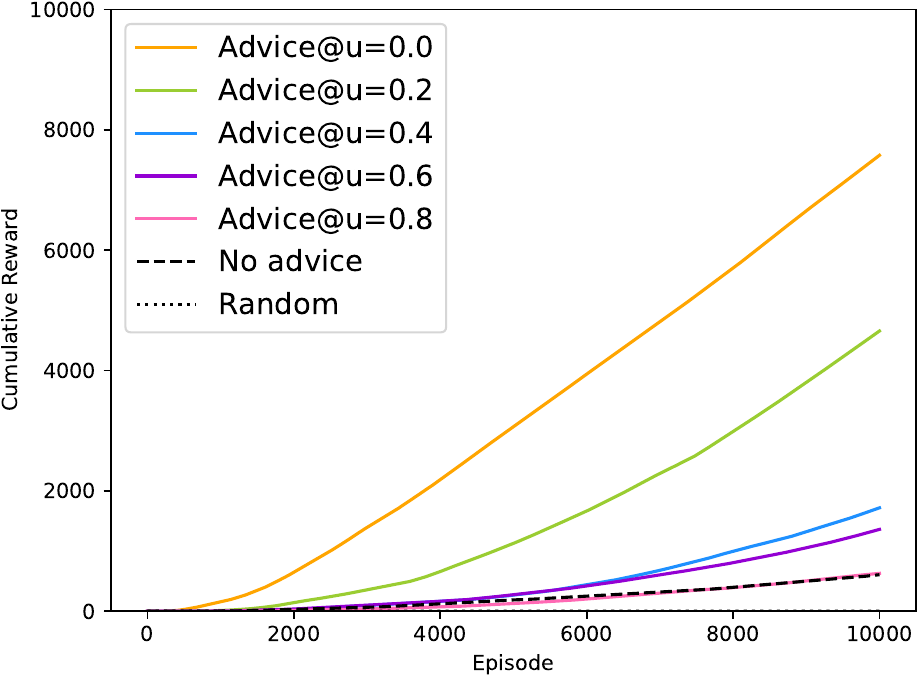}
        \caption{Advice quota: 5\%}
        \label{fig:cumulative-reward-human-5}
    \end{subfigure}

    \caption*{Log scale}

    \begin{subfigure}{0.48\linewidth}
        \includegraphics[width=\linewidth]{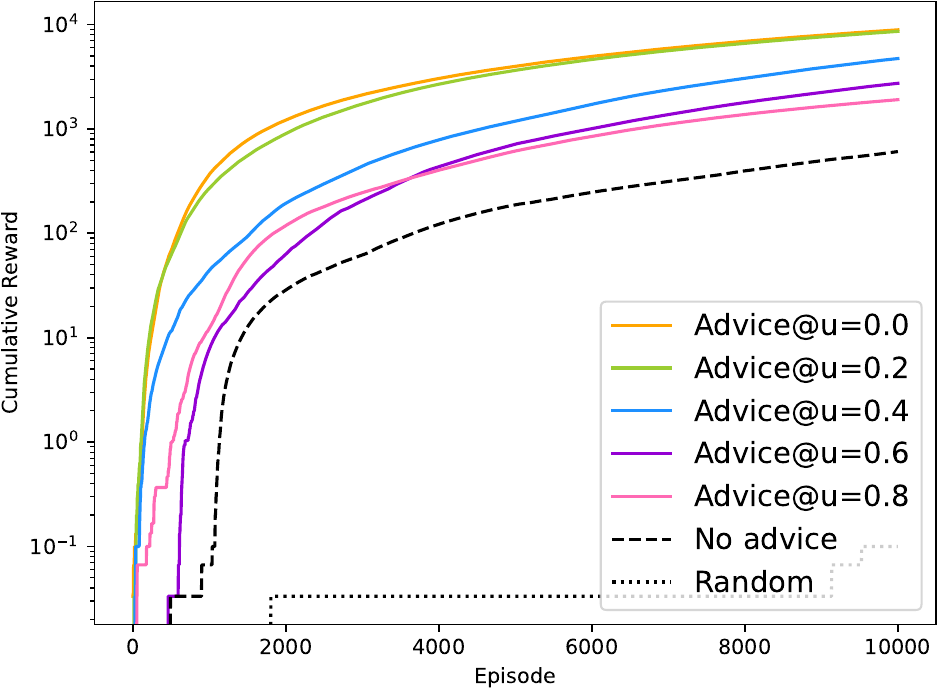}
        \caption{Advice quota: 10\%}
    \end{subfigure}
    \hfil
    \begin{subfigure}{0.48\linewidth}
        \includegraphics[width=\linewidth]{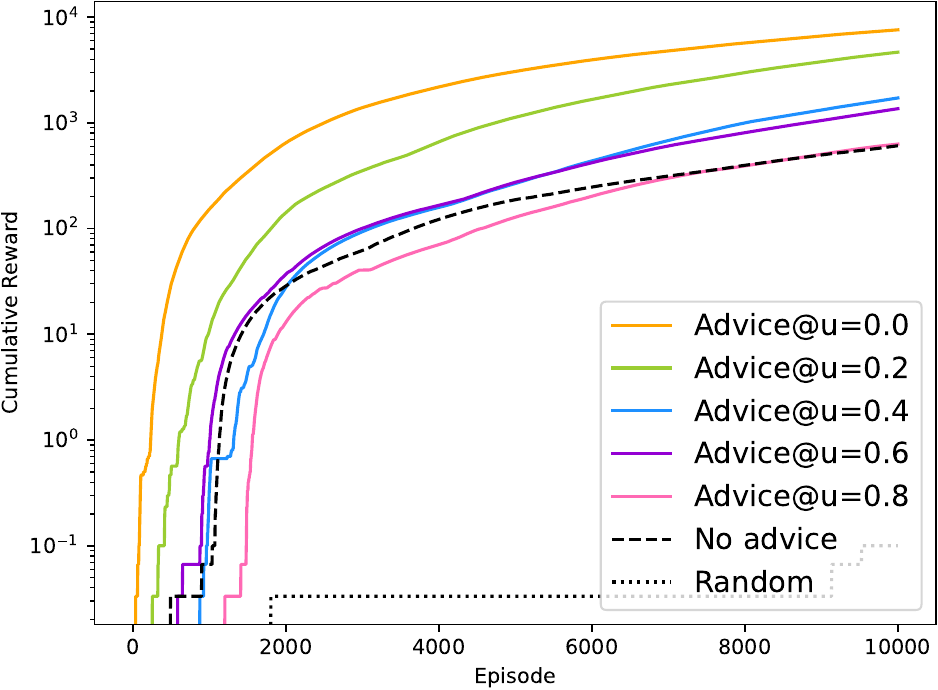}
        \caption{Advice quota: 5\%}
    \end{subfigure}
    
    \caption{Cumulative reward with a human advisor}
    \label{fig:results-cumulativereward-human}
\end{figure}

\figref{fig:results-cumulativereward-human} shows the cumulative rewards of the agent with a human advisor under the two advice quotas.

\subparagraph{Key performance improvement trends are retained}
We see similar trends to those in \secref{sec:results-synthetic}. That is, advised agents outperform the unadvised agent even with a severely reduced advice quota. The human operating with 10\% and 5\% quotas is still able to improve the agent's performance. With 10\% quota, the performance typically reaches that of the oracle. Moreover, at $u=0.2$ and $u=0.8$, the human with 10\% quota performs better than any oracle. At $u=0.4$ and $u=0.6$, the human advisor with 10\% advice quota performs better than the oracle with 20\% advice quota. The only case in which the human does not outperform at least one of the oracles is $u=0.0$; but even in this case, the human's performance is only about 5\% below of that of the oracles.

\subparagraph{The role of certainty increases as the advice quota is reduced} This trend is indicated by the distances between the lines both in the linear and log scales between 10\% and 5\%. At 10\% advice quota, the performance differences between different levels of certainty are similar; at 5\% advice quota, performance drops more prominently as certainty decreases.

\subparagraph{Advice quotas: differences more pronounced in low-certainty cases}
The difference between the two quotas is more obvious as certainty decreases. In the 5\% case, even the fully certain advice $u=0.0$ performs worse than the somewhat certain advice of $u=0.2$ in the 10\% case. Starting at $u=0.2$, the 5\% quota deteriorates substantially; but still outperforming the unadvised agent.

\paragraph{Effects on the policy}

\begin{figure}[H]
    \centering

    \begin{subfigure}[c]{0.32\linewidth}
        \includegraphics[width=\linewidth]{figures/results/heatmaps/heatmap-no_advice-10000.pdf}
        \caption{No advice}
    \end{subfigure}
    \hfil
    \begin{subfigure}[c]{0.64\linewidth}
        \caption*{Advice quota: 10\%}
        
        \begin{subfigure}{0.49\linewidth}
            \includegraphics[width=\linewidth]{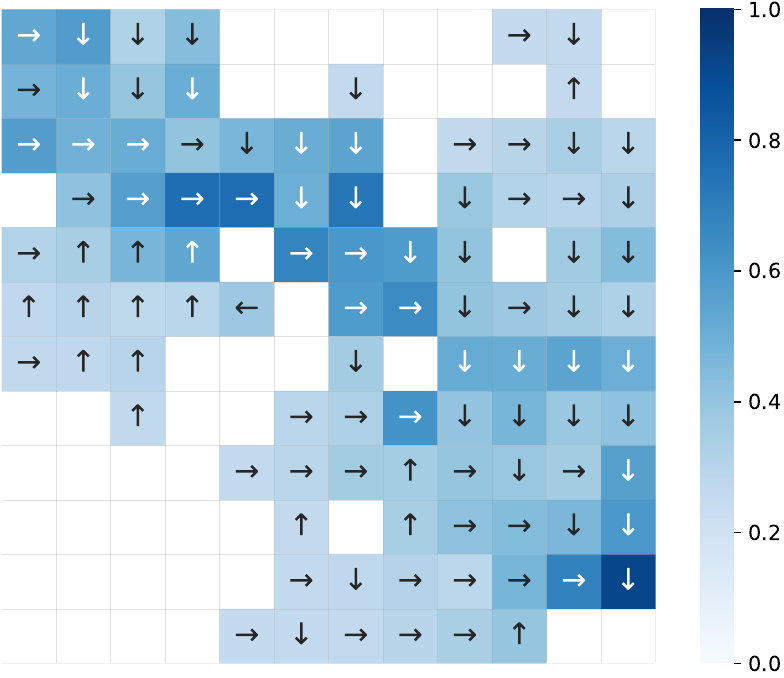}
            \caption{$u=0.4$}
            \label{fig:results-heatmaps-human-10-u04}
        \end{subfigure}
        \hfil
        \begin{subfigure}{0.49\linewidth}
            \includegraphics[width=\linewidth]{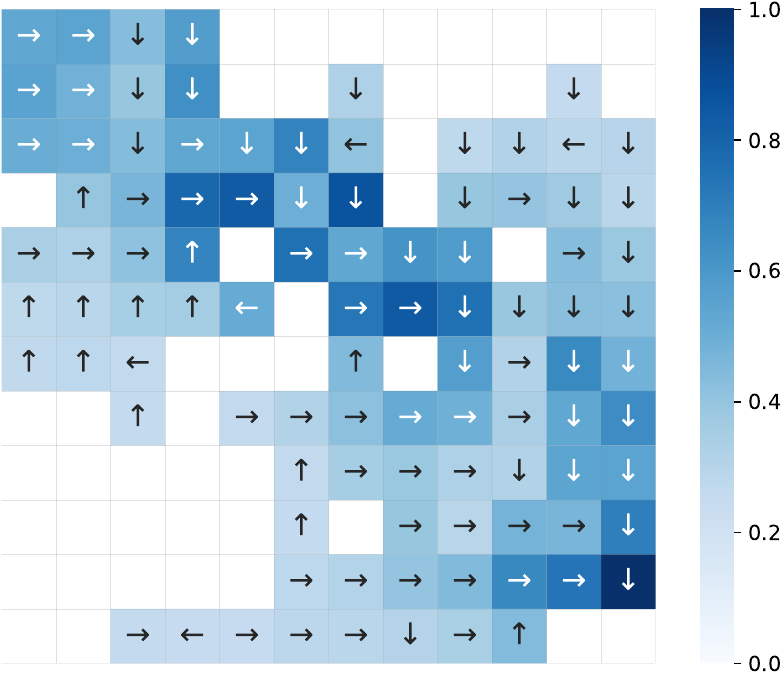}
            \caption{$u=0.0$}
            \label{fig:heatmap-human-10-u00}
        \end{subfigure}
    
        \caption*{Advice quota: 5\%}
    
        \begin{subfigure}{0.49\linewidth}
            \includegraphics[width=\linewidth]{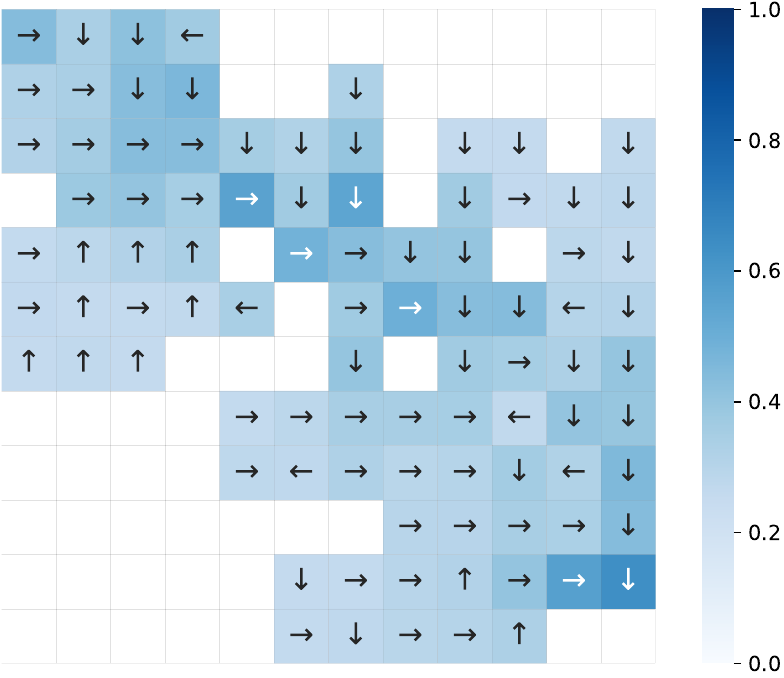}
            \caption{$u=0.4$}
            \label{fig:results-heatmaps-human-5-u04}
        \end{subfigure}
        \hfil
        \begin{subfigure}{0.49\linewidth}
            \includegraphics[width=\linewidth]{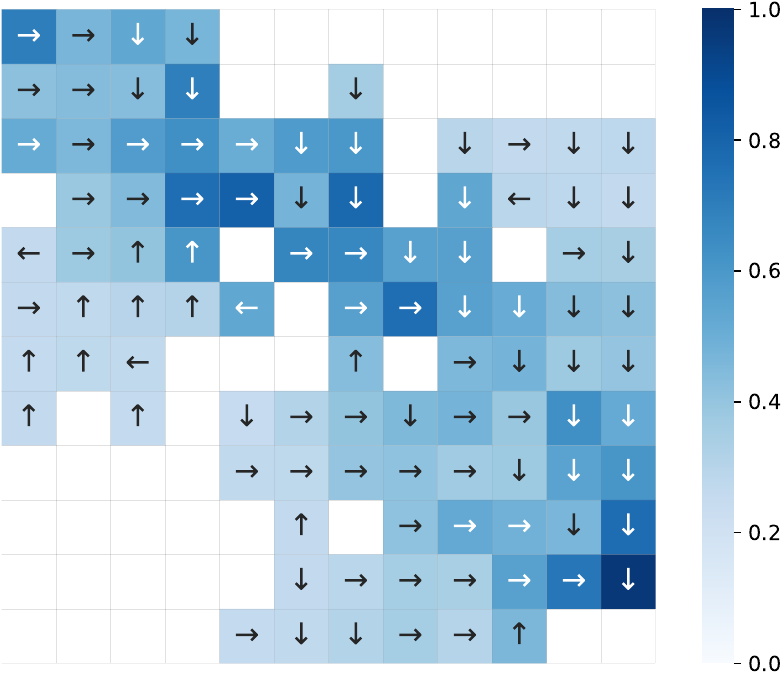}
            \caption{$u=0.0$}
            \label{fig:heatmap-human-5-u00}
        \end{subfigure}
    \end{subfigure}
    
    \caption{Heatmap visualization of the final learned policy with a human advisor}
    \label{fig:results-heatmaps-human-2x2}
\end{figure}

\figref{fig:results-heatmaps-human-2x2} visualizes the policy as a heatmap. The cells of each heatmap correspond to the cells of the Frozen Lake problem; arrows represent the action with the highest probability; and shading corresponds to the probability. Empty cells show unexplored states.

\subparagraph{Key improvements are retained} We see similar trends to those in \secref{sec:results-synthetic}. The policy in advised agents is better reinforced, as indicated by the darker shades of blue, especially around holes and the goal. We see more explored states by advised agents, though the difference between moderately certain advice with low advice quota, and no advice is less substantial.

\subparagraph{Certainty becomes more of a factor} In line with the observations about the cumulative reward in the human-advised mode, certainty becomes an important factor in advice. The policy under 5\% advice quota and at $u=0.4$ shows minor improvements over the unadvised mode, while fully certain advice at $u=0.0$ still shows substantial differences.

\subsubsection{Performance of guidance by cooperating human advisors with partial information}\label{sec:results-coop-human}

\subparagraph{Advisor type} In this experiment, we use the advice of two human advisors who have partial knowledge about the problem. Partial information is modeled by the advisors not giving advice about cells that are beyond a given distance, as the increasing uncertainty would discount these pieces of advice, rendering their effect on the policy virtually nil.

\subparagraph{Advice quota} We evaluate two advice quotas. In the first case, both human advisors provide advice about 10\% of the cells at their own discretion. In the second case, both human advisors provide advice about 5\% of the cells at their own discretion.

\subparagraph{Uncertainty calibration} As opposed to the previous experiments, where we evaluated guidance at multiple levels of uncertainty, here, we calibrate uncertainty for each advice, based on the distance between the advisor and the location of the advice.

\subparagraph{Cooperative advice type} We evaluate two types of cooperation: sequential cooperation (one human in the top-left corner and on in the bottom-right corner), and parallel cooperation (one human in the top-right corner and one in the bottom-left corner).

\paragraph{Cumulative reward}

\begin{figure}[H]
    \centering

    \caption*{Linear scale}
    
    \begin{subfigure}{0.48\linewidth}
        \includegraphics[width=\linewidth]{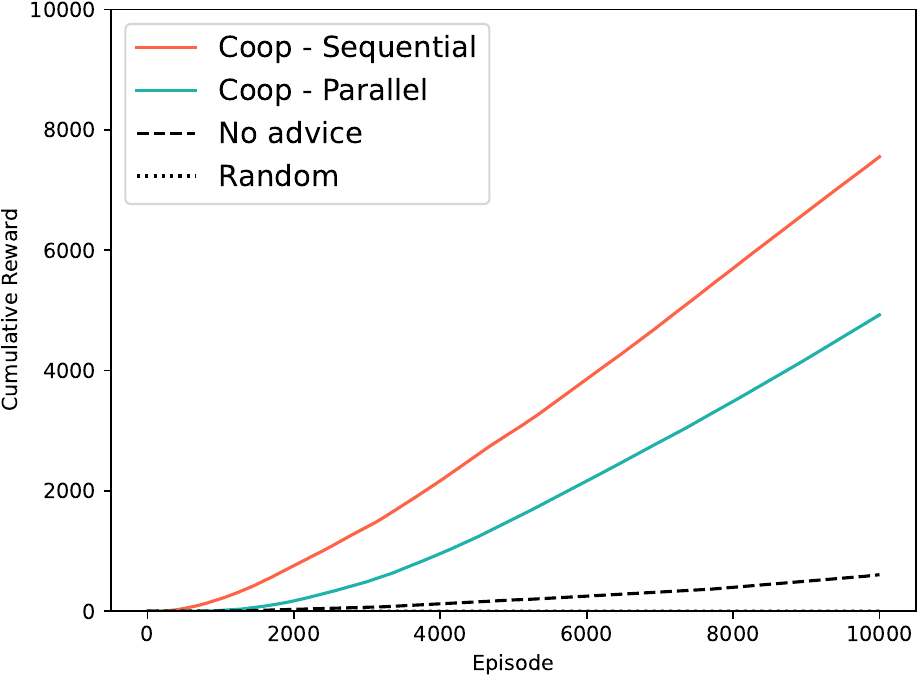}
        \caption{Advice quota: $10\%$}
        \label{fig:results-cumulativereward-coop-10}
    \end{subfigure}
    \hfil
    \begin{subfigure}{0.48\linewidth}
        \includegraphics[width=\linewidth]{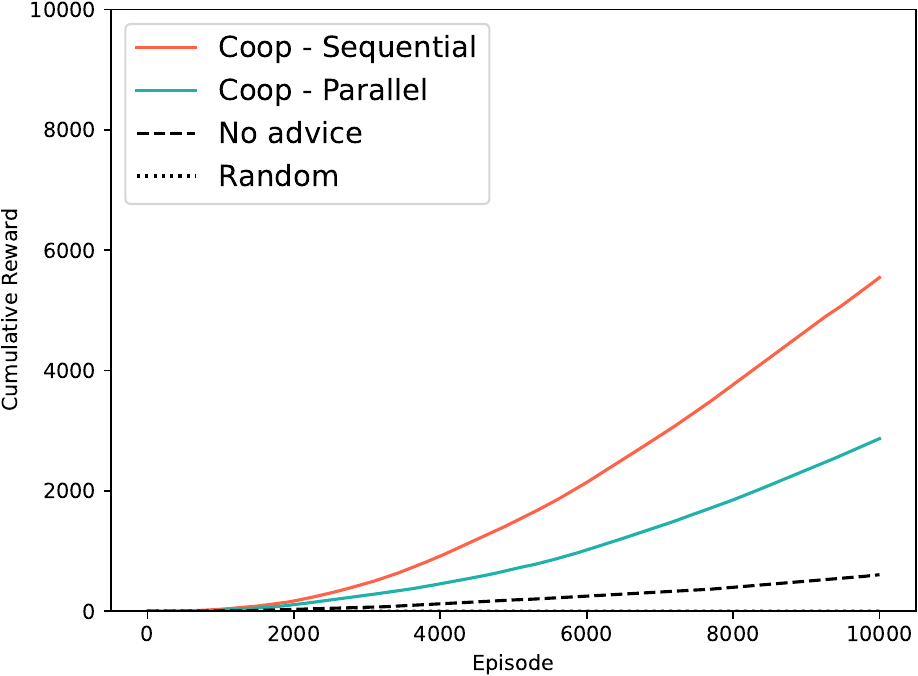}
        \caption{Advice quota: $5\%$}
        \label{fig:results-cumulativereward-coop-5}
    \end{subfigure}

    \caption*{Log scale}
    
    \begin{subfigure}{0.48\linewidth}
        \includegraphics[width=\linewidth]{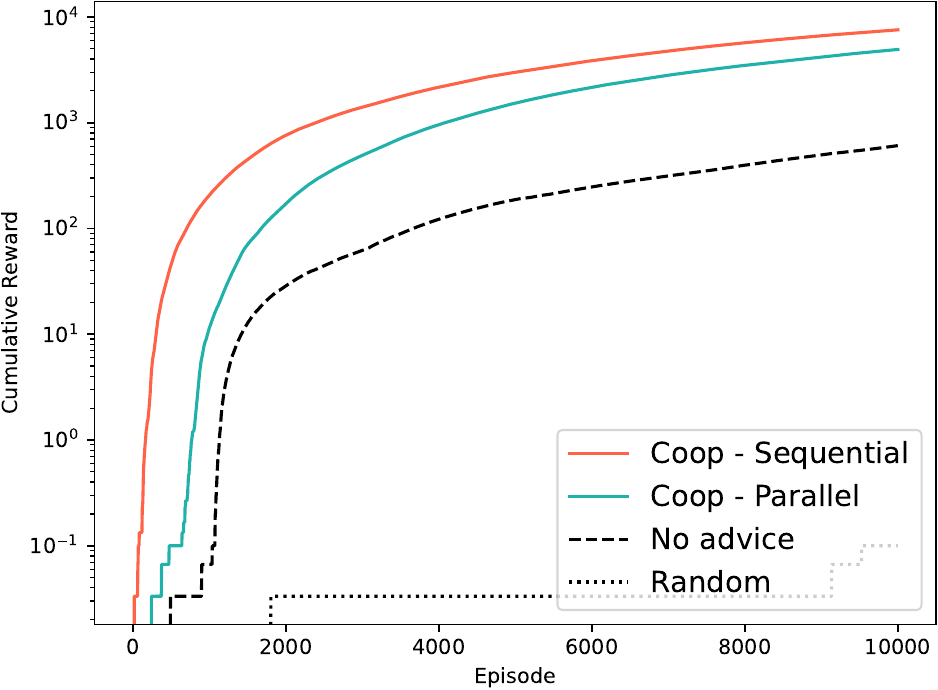}
        \caption{Advice quota: $10\%$}
        \label{fig:results-cumulativereward-coop-10-log}
    \end{subfigure}
    \hfil
    \begin{subfigure}{0.48\linewidth}
        \includegraphics[width=\linewidth]{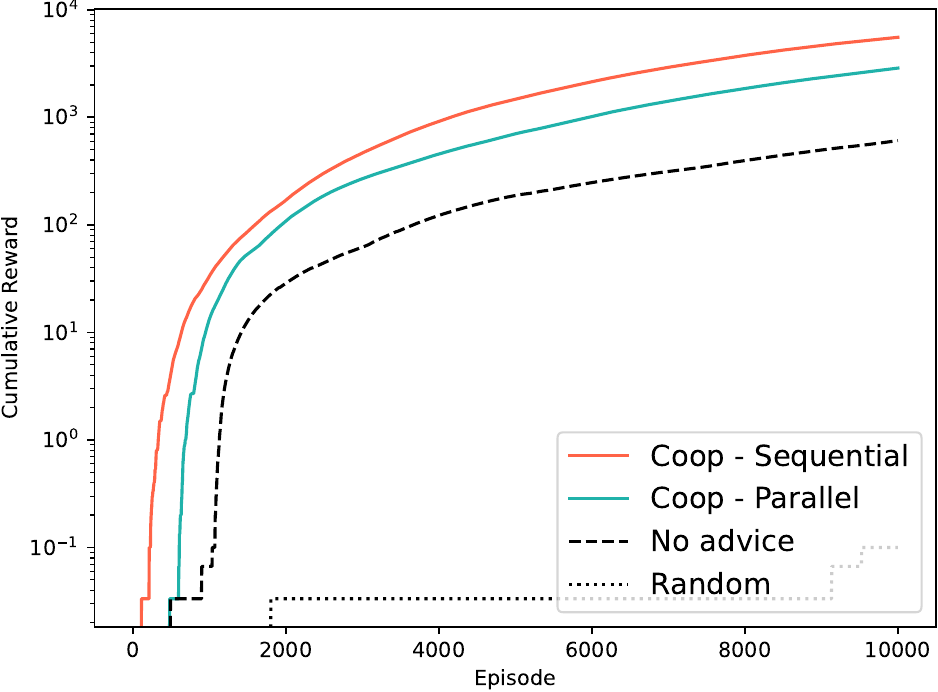}
        \caption{Advice quota: $5\%$}
        \label{fig:results-cumulativereward-coop-5-log}
    \end{subfigure}
    
    \caption{Cumulative reward with advice from two cooperating human advisors}
    \label{fig:results-cumulativereward-coop}
\end{figure}

\figref{fig:results-cumulativereward-coop} shows the cumulative reward of the agent with two cooperating advisors under two different advice quotas.

\subparagraph{Performance improvement over the unadvised agent is retained} Similar to the previous experiments, advised agents outperform the unadvised agent, despite the severely reduced advice quota, and the advisor's partial information about the problem space.

\subparagraph{Sequential cooperation outperforms parallel cooperation} This performance difference is consistent both with 10\% and 5\% advice quota. Parallel cooperation at 10\% shows similar performance to sequential cooperation at 5\%.

\subparagraph{Cooperation with partial information can perform comparably to fully informed humans} Comparing the cooperative results with single human experiments reveals that allowing 10\% quota to each of the two humans in a sequential cooperation (\figref{fig:results-cumulativereward-synthetic-all}) performs similarly to a single human advisor with 10\% quota at almost complete certainty of $u=0.0$--$0.2$ and outperforms the single human advisor with 5\% quota at very high levels of certainty at $u=0.2$ (\figref{fig:results-cumulativereward-human}).

\subparagraph{Learning dynamics}

Learning dynamics are affected in both cooperative cases, but the performance deterioration is more pronounced in the 5\% case. The learning curve is both delayed and less steep as compared to the previous experiments. Still, the pace of learning is significantly more rapid than that of the unadvised agent.

\paragraph{Effects on the policy}

\begin{figure}[H]
    \centering

    \begin{subfigure}[c]{0.32\linewidth}
        \includegraphics[width=\linewidth]{figures/results/heatmaps/heatmap-no_advice-10000.pdf}
        \caption{No advice}
    \end{subfigure}
    \hfil
    \begin{subfigure}[c]{0.64\linewidth}
        \caption*{Advice quota: 10\%}
        
        \begin{subfigure}{0.49\linewidth}
            \includegraphics[width=\linewidth]{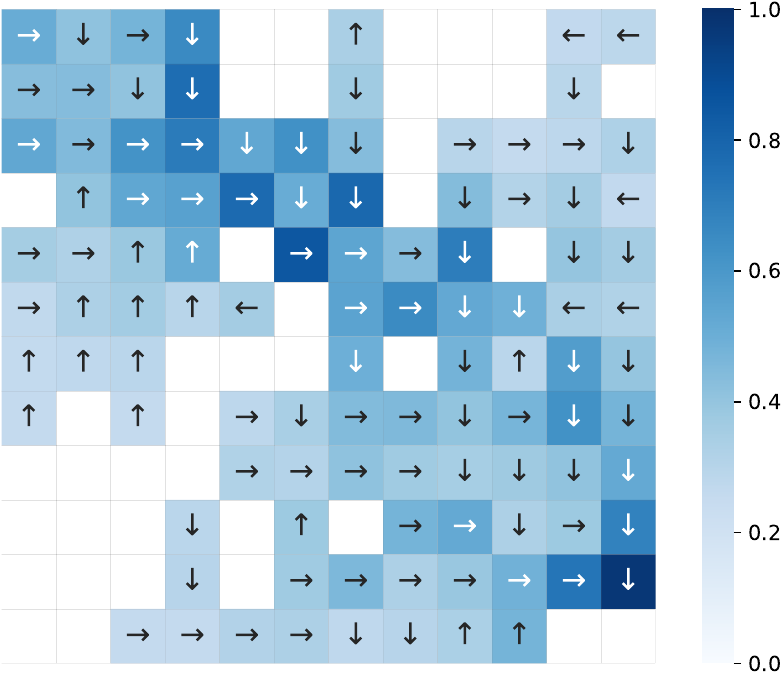}
            \caption{Sequential}
            \label{fig:results-heatmaps-coop-sequential-10}
        \end{subfigure}
        \hfil
        \begin{subfigure}{0.49\linewidth}
            \includegraphics[width=\linewidth]{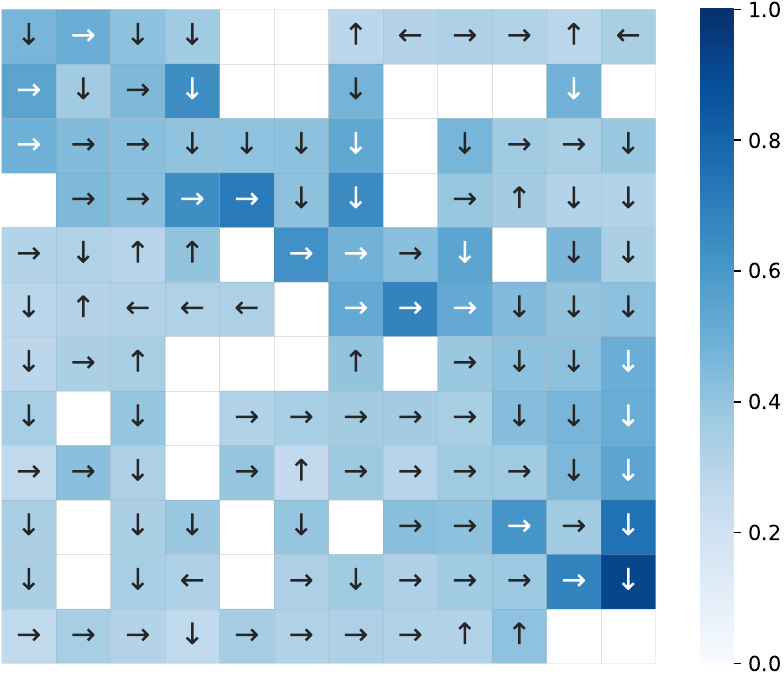}
            \caption{Parallel}
            \label{fig:results-heatmaps-coop-parallel-10}
        \end{subfigure}
    
        \caption*{Advice quota: 5\%}
    
        \begin{subfigure}{0.49\linewidth}
            \includegraphics[width=\linewidth]{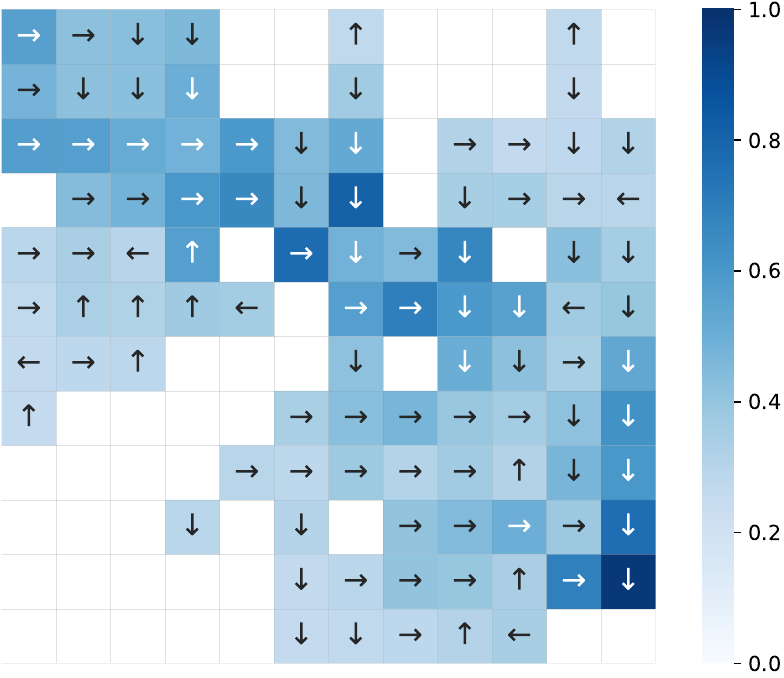}
            \caption{Sequential}
            \label{fig:results-heatmaps-coop-sequential-5}
        \end{subfigure}
        \hfil
        \begin{subfigure}{0.49\linewidth}
            \includegraphics[width=\linewidth]{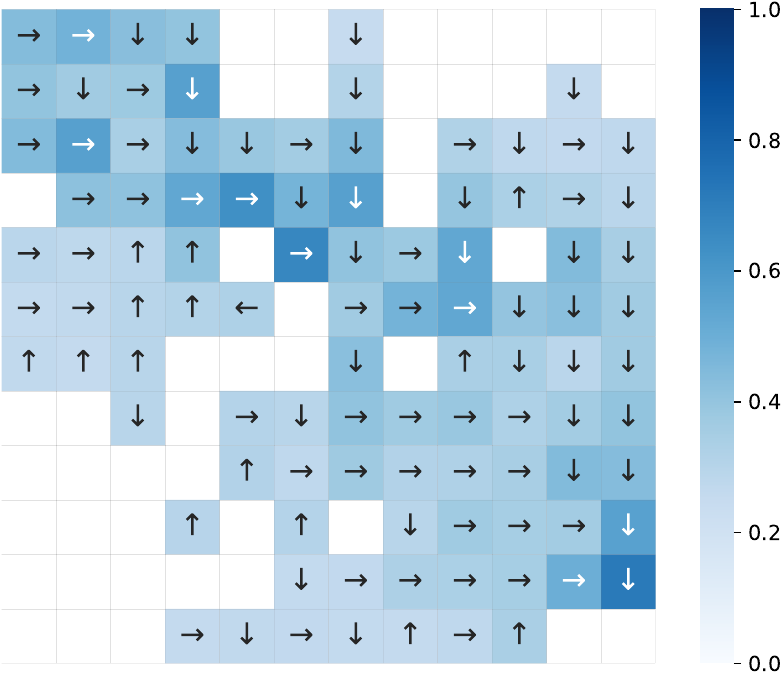}
            \caption{Parallel}
            \label{fig:results-heatmaps-coop-parallel-5}
        \end{subfigure}
    \end{subfigure}
    
    \caption{Heatmap visualization of the final learned policy with two human advisors}
    \label{fig:results-heatmaps-coop-2x2}
\end{figure}

\figref{fig:results-heatmaps-coop-2x2} visualizes the policy as a heatmap. The cells of each heatmap correspond to the cells of the Frozen Lake problem; arrows represent the action with the highest probability; and shading corresponds to the probability. Empty cells show unexplored states.

\subparagraph{Cooperative advice outperforms single human advice} Specifically, sequential cooperation results in a better-reinforced policy and allows for more explored states in more uncertain situations. This is can be inferred from the difference between \figref{fig:results-heatmaps-coop-2x2} and \figref{fig:results-heatmaps-human-2x2}: we encounter more darker cells (better reinforced local decisions) and less empty cells (less unexplored states).
The effect on the policy is weaker to that of the oracle.

\subparagraph{Sequential cooperation is better in reinforcement and parallel cooperation is better in exploration performance} In pattern unique to cooperative guidance, we observe that heatmaps visualizing sequential cooperation show stronger reinforced local decisions (\figref{fig:results-heatmaps-coop-sequential-10} over \figref{fig:results-heatmaps-coop-parallel-10}, and \figref{fig:results-heatmaps-coop-sequential-5} over \figref{fig:results-heatmaps-coop-parallel-5}); while heatmaps visualizing parallel cooperation show better coverage of the problem (\figref{fig:results-heatmaps-coop-parallel-10} over \figref{fig:results-heatmaps-coop-sequential-10}, and \figref{fig:results-heatmaps-coop-parallel-5} over \figref{fig:results-heatmaps-coop-sequential-5}). In fact, parallel cooperative guidance at 10\% (\figref{fig:results-heatmaps-coop-parallel-10}) explores as many states as the oracle at 100\% quota and $u=0.0$ uncertainty (\figref{fig:heatmap-synthetic-all-00}), i.e., the fully informed and fully certain oracle.

\subsection*{Conclusion of RQ2}

In \secref{sec:introduction}, we asked \textit{How does opinion-based guidance affect the performance of reinforcement learning agents?} as our second research question.
In this section, we designed and executed a study to investigate this research question.

\begin{conclusionframe}{RQ2}
Our observations indicate that opinion-based guidance contributes to significant performance improvement in reinforcement learning agents, even at moderate-to-high uncertainty.
Human advisors, even with substantially lower advice quotas, can perform at the level of an oracle, and might even outperform them. Uncertainty becomes more of a factor as advice quota decreases. In realistic settings of cooperating humans with partial and complementary information about the problem, we find that guidance, even at low advice quotas, still results in significant improvements over unadvised agents.
\end{conclusionframe}
\section{Discussion}\label{sec:discussion}

In this section, we discuss the results and some of the important implications.

\subsection{The impact of advice -- Discussion of the results}

\paragraph{Even uncertain advice is useful for guidance}

The key takeaway of this work is that \textbf{even uncertain opinions improve the performance} of the RL agent.
The only case in which an opinion-guided RL agent did not outperform the unadvised agent is in the case with a single human advisor with 5\% quota at uncertainty level $u=0.8$ (\figref{fig:cumulative-reward-human-5}).

To provide an \textbf{intuition of what uncertainty at the level of $u=0.8$ means}, e.g., in terms of the Frozen Lake problem, consider the two edge cases when an advisor is sure about a state being (i) disadvantageous and (ii) advantageous. Given that $u=0.8$, the remaining opinion weight is $0.2$, which has to be distributed over disbelief \textit{d} and belief \textit{b}. Case (i) translates to an opinion $\omega_i = (0.2, 0.0, 0.8, 0.25)$, and the second to $\omega_{ii} = (0.0, 0.2, 0.8, 0.25)$. Using the formulas from \secref{sec:sl}, the two opinions translate to $P_i= 0.2 + 0.25 \times 0.8 = 0.6$ and $P_{ii} = 0.0 + 0.25 \times 0.8 = 0.4$, respectively. That is, the advisor can express they believe the probability of a particular state being advantageous should be between 0.4--0.6 probability---good for a meager 20\% wiggle room. In other words, the advisor cannot express anything below 40\% and above 60\% which, evidently, is the descriptive power an unadvised agent has with 5\% advice quota (\figref{fig:cumulative-reward-human-5}). (At higher advice quotas, the same 0.4--0.6 advice range, i.e., the same uncertainty of $u=0.8$ is used more efficiently by the advisors, resulting in performance improvements over the unadvised agent.)

Considering that opinions emerge earlier than hard evidence can be produced~\citep{dagenais2024driving}, opinion-based guidance positions to be an important improvement over traditional guidance mechanisms. To leverage the benefits of opinion-based guidance, opinions should be elicited efficiently (\secref{sec:discussion-directions}).

\paragraph{Effectiveness of guidance is impacted by certainty and exhaustiveness}

Two key trends we observe are that \textbf{performance deteriorates as uncertainty increases} and that \textbf{performance deteriorates as advice quota decreases}. We observe these trends in each of our experiments (see \tabref{tab:cumulative-rewards} and \figref{fig:cumulative-rewards-all}).
The performance drop between advice quotas is more pronounced at higher levels of uncertainty. Specifically, at $u=0.8$, advice loses its guidance value at 5\% quota as the advised agent does not perform better than the unadvised one (single human case, \figref{fig:cumulative-reward-human-5}). This combination of parameters seems to be the \textbf{lower bound of utility} of our approach.

Another interesting trend to observe is that at high levels of certainty ($u=0.0$) the human advisor with 5\% advice quota reinforces as many key decisions as with 10\% quota (cf. \figref{fig:heatmap-human-5-u00} and \figref{fig:heatmap-human-10-u00}).
Furthermore, two cooperating human advisors, each with 5\% advice quota, reinforce the agent's policy better than a single advisor with 10\% quota (cf., e.g., \figref{fig:results-heatmaps-coop-sequential-5} and \figref{fig:heatmap-human-10-u00}).

Thus, while higher certainty and higher advice quota (exhaustiveness) improve the effectiveness of guidance, the cost of a thorough advice strategy might not be always justified.
We recommend adopters of our approach to seek trade-offs between certainty, exhaustiveness, and the costs of providing advice.

\paragraph{Human advisors are as effective as idealized oracle advisors}
Comparing the performance of the oracles and human advisors (\tabref{tab:cumulative-rewards}), we do not observe substantial differences. The single human advisor with sufficient quota achieves a performance similar to that of the oracle.
The performance is within 5\% in three of five evaluated cases between oracles and human advisors ($u=0.0, 0.2, 0.6$), with one of these cases showing an improvement of the human over the oracle ($u=0.2$). In two of five evaluated cases, the difference is more pronounced, once in favor of the oracle ($u=0.4$) and once in favor of the human advisor ($u=0.8$). Improvements due to human advice are particularly apparent in highly uncertain situations.

We attribute these observations to human creativity in providing advice. Thus, ours is another one in the long line of works that highlight the role of human cognition in computer-automated methods, such as machine learning~\cite{jarrahi2018artificial,bansal2019accuracy}.

Since the human advisor can quickly become the bottleneck in RL, adopters of our approach should consider implementing efficient advice strategies, particularly asynchronous ones in which human and computer-automated agents can work independently.

\paragraph{Cooperative guidance with partial information performs well compared to advisors with complete information}

Our experiments show that the performance of cooperative guidance with partial information about the problem is only slightly below the performance of oracles and single human advisors with complete information about the problem.
At $u=0.4$, sequential cooperation with $2\times5$\% quota outperforms the comparable single human advisor with 10\% quota (5\,544.100 vs 4\,727.433; +17.28\%) and the oracle with twice as much quota of 20\% (5\,544.100 vs 4\,218.000; +31.44\%). Related to this latter comparison, at $u=0.2$ sequential cooperation with $2\times10$\% quota performs close to the oracle with comparable quota (20\%) -- 7\,551.567 vs 7\,835.267 (-3.62\%).
These results demonstrate that partial information equates to moderate levels of uncertainty in fully-informed situations. The cooperative experiments we conducted emulate a highly realistic setting. Thus, it is plausible to expect that \textbf{the benefits of opinion-based guidance can be retained in real applications}.

Comparing cooperation modes, we observe that sequential cooperation gains more (\figref{fig:results-cumulativereward-coop-10} and \figref{fig:results-cumulativereward-coop-5}) cumulative reward and earlier (\figref{fig:results-cumulativereward-coop-10-log} and \figref{fig:results-cumulativereward-coop-5-log}) than parallel cooperation.
In sequential cooperative guidance, uncertainty is lower in critical areas of map, i.e., around start, goal, and along path (\figref{fig:results-heatmaps-coop-sequential-10} vs \figref{fig:results-heatmaps-coop-parallel-10}; and \figref{fig:results-heatmaps-coop-sequential-5} vs \figref{fig:results-heatmaps-coop-parallel-5}). This means advice about critical areas has more of an impact on the agent's policy, which leads to higher cumulative reward. In particular, having an advisor located at the agent's start state where uncertainty is low helps the agent navigate the early parts of the map better. Thus, the agent is able to avoid holes early and take more steps in early episodes (more exploration), leading to learning the optimal policy earlier (as demonstrated in the log charts \figref{fig:results-cumulativereward-coop-10-log} and \figref{fig:results-cumulativereward-coop-5-log}).

\subsection{Open challenges and research opportunities}\label{sec:discussion-directions}

Our approach opens up new challenges and research opportunities for prospective researchers.

\paragraph{Interaction patterns for optimal guidance}

We relied on one-time advice provided before exploration. However, more sophisticated advice dynamics have been used in guided RL~\citep{najar2021reinforcement,scherf2022learning}. We recommend researching more \textbf{interactive patterns of advising}. A known drawback of placing humans into the learning loop is the high cost of human reward signals, which are caused by the human stakeholders quickly becoming performance bottlenecks~\citep{knox2012reinforcement}. We see opportunities in opinion-based guidance in this aspect as opinions might be easier to formulate than hard evidence, alleviating the load on the human. Another interesting research opportunity our approach motivates is the investigation of \textbf{push and pull mechanisms} in advising RL agents, in which the human can push advice toward the agent, and the agent can actively query the human for advice at any time. Such mechanisms would allow for human advisors to provide advice under limited quotas when advice matters the most.

\paragraph{Languages for expressing advice}

A key advantage of opinion-based guidance is the rapid pace at which opinions emerge. To leverage this advantage to its fullest, intuitive languages are needed to express opinions. Domain-specific languages (DLSs) \citep{schmidt2006model} have been widely used to capture expert opinions at high levels of abstraction, narrowing the gap between the modeling language and human cognition. However, DSLs typically operate with reduced syntax to remain truly specific to the problem domain, and therefore, \textbf{automated methods for DSL engineering} are in high demand~\citep{bucchiarone2020grand}. This is true for guided RL as well. We envision language workbenches that allow for the automated construction of DSLs through which advice can be provided by domain concepts. For example, in the Frozen Lake example, one might want to use descriptive quantifiers, such as ``hazardous'' or ``beneficial'' instead of relying on an integer scale.
Unfortunately, there is little research on \textbf{domain-specific languages for RL}, and existing languages mostly focus on programming RL procedures~\citep{molderez2019marlon} rather than expressing opinions at high cognitive levels. We see plenty of room for research targeting domain-specific modeling languages for RL, inspired by the advancements in the model-driven engineering community~\citep{david2013ontology,kulagin2022ontology,wu2005automated}.

\paragraph{Opinion-based guiding in value-based and deep reinforcement learning}

In our approach, we opted for a policy-based RL algorithm because it allows for a better investigation of the impact of opinion on the explicitly represented policy. We are fairly confident that our method translates to value-based techniques as well, but validation is left for future work. A fairly more complicated challenge is introducing guidance into deep reinforcement learning, where the policy is encoded in a neural network, and the lack of explainability might hinder the development of a sound guiding framework. With the demonstrated utility of deep reinforcement learning in numerous problems, such as the management of engineering systems~\citep{andriotis2019managing} and simulator inference~\citep{david2022devs}, its support by opinion-based guidance seems like a natural next step.
\section{Threats to validity and limitations}\label{sec:threats}

\paragraph{Construct validity}
Observations of improved exploration performance may be artifacts of fortunately aligned problems (here: maps) or a few fortunate pieces of advice rather than artifacts of systematic effects. To mitigate this threat to validity, first, we experimented with different map layouts and observed the same performance trends. To facilitate consistent manual experimentation, we developed a map generator that generates map layouts in a reproducible way. Second, we ran our scaled-up experiments with a variety of settings, including both human and oracle advisors.
A negligible threat to validity in cooperative experiments stems from the human advisors' strategy of considering their own level of certainty when giving advice. Human advisors prefer not to waste their limited advice quota and thus, will be likely biased towards giving advice with higher certainty. Thus, the positive results of cooperative guidance are artifacts of opinion-based guidance and human behavior. For the purposes of our research questions, human behavior does not threaten validity.

\paragraph{Internal validity}
Evaluating the approach with human advisors inevitably poses threats to internal validity. Specifically, maturation and testing might have influenced our results as the authors experimented with advising strategies and advice values. For example, cells that are not holes or goals are labeled freely with -1, 0, or 1, and a human might learn how to misuse these values to achieve a specific behavior by the agent. We mitigated these threats by defining clear advise score rules, which are also used by the oracle advisor.

\paragraph{External validity}
We chose only one flavor of RL---a policy-based method---to demonstrate our approach, and we did not provide evidence that our approach can be applied to other RL methods. Although we are reasonably confident that our methods translate to RL methods where the policy is not explicitly represented (e.g., value-based methods), future work will investigate this question in more detail.

\paragraph{Limitations}
We used a relatively simple guidance strategy, in which advice is provided once before the exploration. It is reasonable to assume that more interactive guidance modes might shed light on additional benefits and challenges. However, we are reasonably confident that the scope of this article was sufficient to draw key takeaways that hold in more interactive reinforcement learning setups as well. Nonetheless, we will experiment with interactive guidance modes in future work.
\section{Conclusion}\label{sec:conclusion}

In this article, we presented a method to guiding reinforcement learning agents by opinions---cognitive constructs subject to uncertainty. We have devised a formal, mathematically sound method for using opinions for policy shaping, based on subjective logic, and demonstrated that opinions, even at moderate-to-substantial levels of uncertainty help achieve better performance in reinforcement learning. Specifically, we observe improved cumulative rewards and faster convergence to the theoretical maximum performance in opinion-guided reinforcement learning agents.

Our method is particularly useful in situations where hard evidence is impractical or infeasible to produce, e.g., in problems with a scarce reward structure, where human creativity plays an important role. Such situations are, for example, problems with infeasible, inaccessible, hazardous, or costly states, which, consequently, cannot be efficiently explored by the reinforcement learning agent. In addition, problems, where the expertise of multiple distinct domain experts is required, will benefit from our approach as well, thanks to the sound fusion semantics provided by subjective logic.

Although demonstrated in a topological problem, our approach translates naturally to problems where the notion of distance is more abstract, e.g., in guided design-space exploration. Such directions are left for future work. Additional future work will focus on extending the method to different flavors of reinforcement learning, supporting more interactive advice modes, and evaluating our method in real industry settings.

\backmatter

\clearpage
\begin{appendices}
\section{Study setup}\label{sec:appendix-studysetup}

\begin{figure}[htb]
    \centering
    \includegraphics[trim={0 0.3cm 0.3cm 0},clip,width=0.5\linewidth]{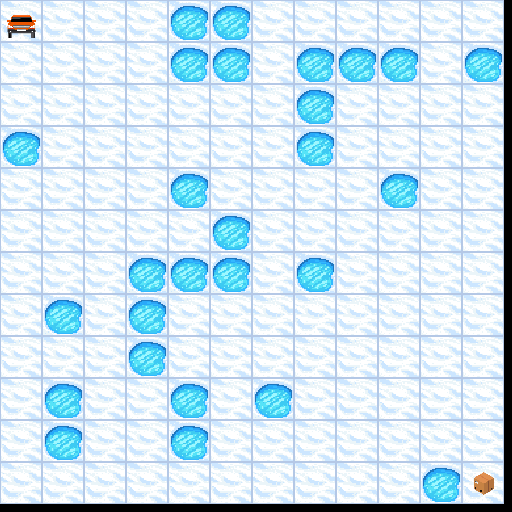}
    \caption{The map used in the experiments. (\texttt{12x12 seed 63})}
    \label{fig:experiment-map}
\end{figure}

\clearpage
\section{Input artifacts}\label{sec:appendix-artifacts}

\subsection{Oracle}

\subsubsection{Advice -- 100\% quota, oracle}

\begin{figure}[H]
    \begin{multicols}{3}
      \begin{lstinputlisting}[frame=none,mathescape,
      caption={Advice of the oracle with 100\% quota, with an opinion about every cell of \figref{fig:experiment-map}.}, 
      label={lst:advice-file-synth-100}]{sections/appendices/synthetic/advice-12x12-seed63-all.txt}
      \end{lstinputlisting}
    \end{multicols}
\end{figure}

\subsubsection{Advice -- 20\% quota, oracle}

\begin{figure}[H]
  \begin{lstinputlisting}[mathescape,
  caption={Advice of the oracle with 20\% quota, with an opinion about the holes and the goal cell of \figref{fig:experiment-map}.}, 
  label={lst:advice-file-synth-20}]{sections/appendices/synthetic/advice-12x12-seed63-holes.txt}
  \end{lstinputlisting}
\end{figure}

\subsection{Single human experiments}

\subsubsection{Advice -- 10\% quota, single human advisor}

\begin{figure}[H]
    \begin{subfigure}{0.48\linewidth}
        \begin{lstinputlisting}[mathescape]{sections/appendices/human/advice-12x12-seed63-human10.txt}
        \end{lstinputlisting}
    \end{subfigure}
    \hfil
    \begin{subfigure}{0.48\linewidth}
        \centering
        \includegraphics[width=0.66\linewidth]{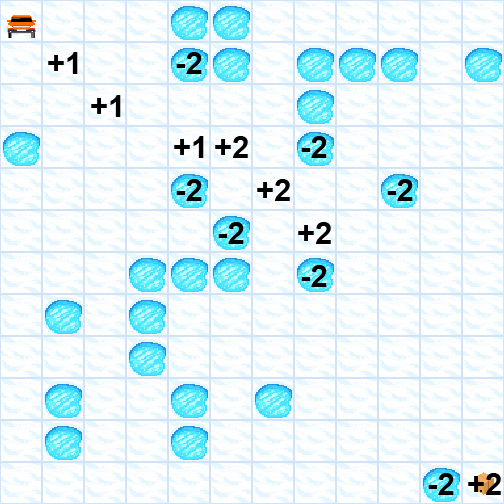}
    \end{subfigure}
    \caption{Advice of the single human advisor with $10\%$ quota, and its visualization.}
    \label{fig:advice-human-10}
\end{figure}

\subsubsection{Advice -- 5\% quota, single human advisor}

\begin{figure}[H]
    \begin{subfigure}{0.48\linewidth}
        \begin{lstinputlisting}[mathescape]{sections/appendices/human/advice-12x12-seed63-human5.txt}
        \end{lstinputlisting}
    \end{subfigure}
    \hfil
    \begin{subfigure}{0.48\linewidth}
        \centering
        \includegraphics[width=0.66\linewidth]{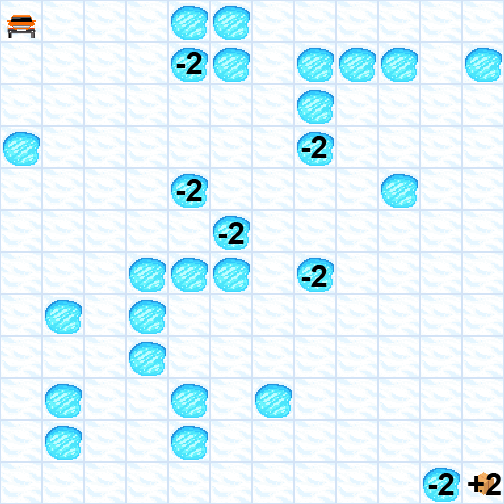}
    \end{subfigure}
    \caption{Advice of the single human advisor with $5\%$ quota, and its visualization.}
    \label{fig:advice-human-5}
\end{figure}

\subsection{Cooperative human experiments}

\subsubsection{Sequential cooperation, 10\% quota each}

\begin{figure}[H]
    \begin{subfigure}{0.48\linewidth}
        \begin{lstinputlisting}[mathescape]{sections/appendices/coop/advice-12x12-seed63-coop10-A1-topleft.txt}
        \end{lstinputlisting}
    \end{subfigure}
    \hfil
    \begin{subfigure}{0.48\linewidth}
        \centering
        \includegraphics[width=0.66\linewidth]{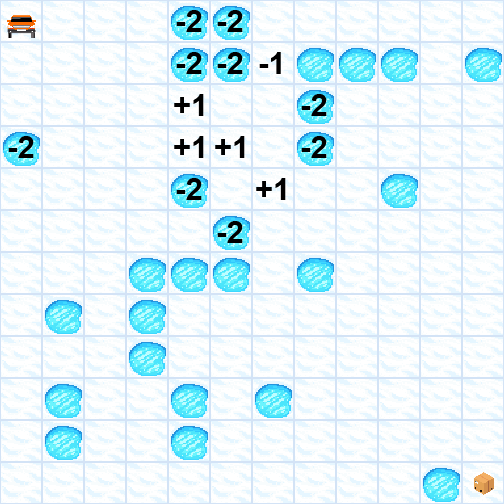}
    \end{subfigure}
    \caption{Advice from the \textit{top-left} corner in the sequential cooperative human experiments, with $10\%$ quota, and its visualization. The counterpart of the cooperative advice is shown in \figref{fig:advice-file-coop-seq-10-BR}.}
    \label{fig:advice-file-coop-seq-10-TL}
\end{figure}

\begin{figure}[H]
    \begin{subfigure}{0.48\linewidth}
        \begin{lstinputlisting}[mathescape]{sections/appendices/coop/advice-12x12-seed63-coop10-A2-bottomright.txt}
        \end{lstinputlisting}
    \end{subfigure}
    \hfil
    \begin{subfigure}{0.48\linewidth}
        \centering
        \includegraphics[width=0.66\linewidth]{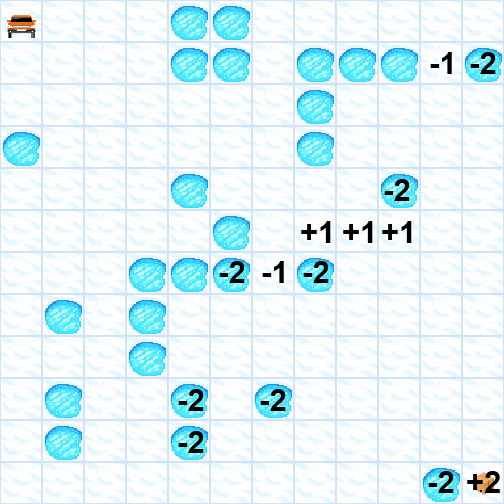}
    \end{subfigure}
    \caption{Advice from the \textit{bottom-right} corner in the sequential cooperative human experiments, with $10\%$ quota, and its visualization. The counterpart of the cooperative advice is shown in \figref{fig:advice-file-coop-seq-10-TL}.}
    \label{fig:advice-file-coop-seq-10-BR}
\end{figure}

\subsubsection{Sequential cooperation, 5\% quota each}

\begin{figure}[H]
    \begin{subfigure}{0.48\linewidth}
        \begin{lstinputlisting}[mathescape]{sections/appendices/coop/advice-12x12-seed63-coop5-A1-topleft.txt}
        \end{lstinputlisting}
    \end{subfigure}
    \hfil
    \begin{subfigure}{0.48\linewidth}
        \centering
        \includegraphics[width=0.66\linewidth]{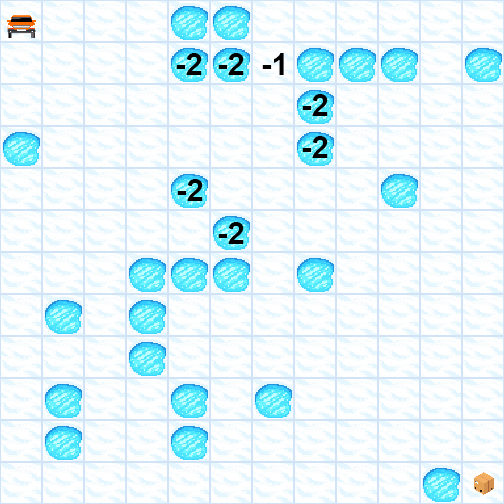}
    \end{subfigure}
    \caption{Advice from the \textit{top-left} corner in the sequential cooperative human experiments, with $5\%$ quota, and its visualization. The counterpart of the cooperative advice is shown in \figref{fig:advice-file-coop-seq-5-BR}.}
    \label{fig:advice-file-coop-seq-5-TL}
\end{figure}

\begin{figure}[H]
    \begin{subfigure}{0.48\linewidth}
        \begin{lstinputlisting}[mathescape]{sections/appendices/coop/advice-12x12-seed63-coop5-A2-bottomright.txt}
        \end{lstinputlisting}
    \end{subfigure}
    \hfil
    \begin{subfigure}{0.48\linewidth}
        \centering
        \includegraphics[width=0.66\linewidth]{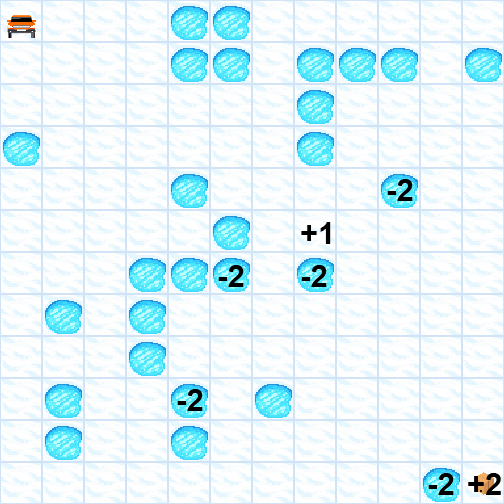}
    \end{subfigure}
    \caption{Advice from the \textit{bottom-right} corner in the sequential cooperative human experiments, with $5\%$ quota, and its visualization. The counterpart of the cooperative advice is shown in \figref{fig:advice-file-coop-seq-5-TL}.}
    \label{fig:advice-file-coop-seq-5-BR}
\end{figure}

\subsubsection{Parallel cooperation, 10\% quota each}

\begin{figure}[H]
    \begin{subfigure}{0.48\linewidth}
        \begin{lstinputlisting}[mathescape]{sections/appendices/coop/advice-12x12-seed63-coop10-A1-topright.txt}
        \end{lstinputlisting}
    \end{subfigure}
    \hfil
    \begin{subfigure}{0.48\linewidth}
        \centering
        \includegraphics[width=0.66\linewidth]{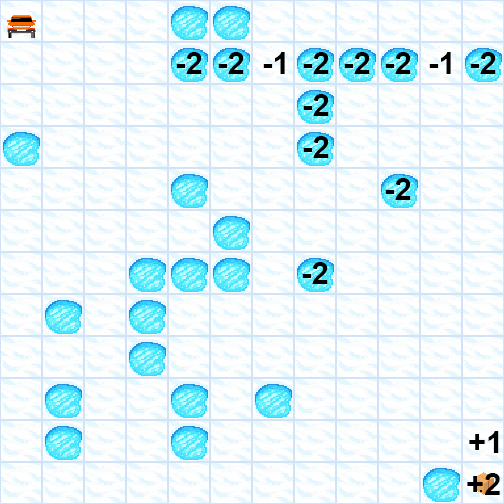}
    \end{subfigure}
    \caption{Advice from the \textit{top-right} corner in the parallel cooperative human experiments, with $10\%$ quota, and its visualization. The counterpart of the cooperative advice is shown in \figref{fig:advice-file-coop-par-10-BL}.}
    \label{fig:advice-file-coop-par-10-TR}
\end{figure}

\begin{figure}[H]
    \begin{subfigure}{0.48\linewidth}
        \begin{lstinputlisting}[mathescape]{sections/appendices/coop/advice-12x12-seed63-coop10-A2-bottomleft.txt}
        \end{lstinputlisting}
    \end{subfigure}
    \hfil
    \begin{subfigure}{0.48\linewidth}
        \centering
        \includegraphics[width=0.66\linewidth]{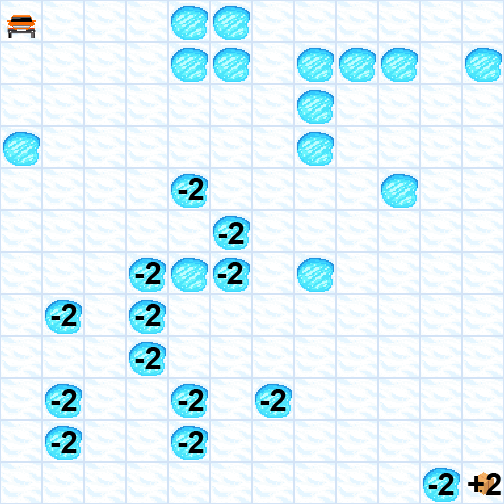}
    \end{subfigure}
    \caption{Advice from the \textit{bottom-left} corner in the parallel cooperative human experiments, with $10\%$ quota, and its visualization. The counterpart of the cooperative advice is shown in \figref{fig:advice-file-coop-par-10-TR}.}
    \label{fig:advice-file-coop-par-10-BL}
\end{figure}

\subsubsection{Parallel cooperation, 5\% quota each}

\begin{figure}[H]
    \begin{subfigure}{0.48\linewidth}
        \begin{lstinputlisting}[mathescape]{sections/appendices/coop/advice-12x12-seed63-coop5-A1-topright.txt}
        \end{lstinputlisting}
    \end{subfigure}
    \hfil
    \begin{subfigure}{0.48\linewidth}
        \centering
        \includegraphics[width=0.66\linewidth]{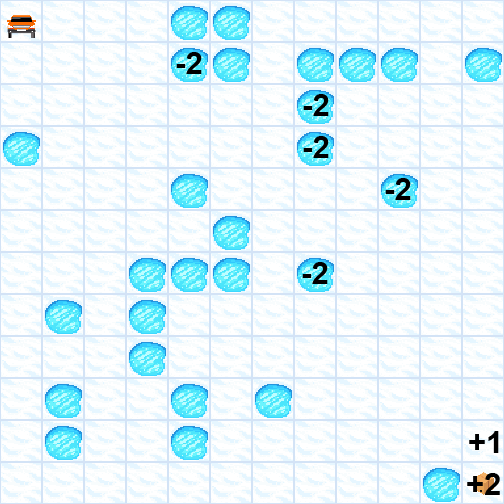}
    \end{subfigure}
    \caption{Advice from the \textit{top-right} corner in the parallel cooperative human experiments, with $5\%$ quota, and its visualization. The counterpart of the cooperative advice is shown in \figref{fig:advice-file-coop-par-5-BL}.}
    \label{fig:advice-file-coop-par-5-TR}
\end{figure}

\begin{figure}[H]
    \begin{subfigure}{0.48\linewidth}
        \begin{lstinputlisting}[mathescape]{sections/appendices/coop/advice-12x12-seed63-coop5-A2-bottomleft.txt}
        \end{lstinputlisting}
    \end{subfigure}
    \hfil
    \begin{subfigure}{0.48\linewidth}
        \centering
        \includegraphics[width=0.66\linewidth]{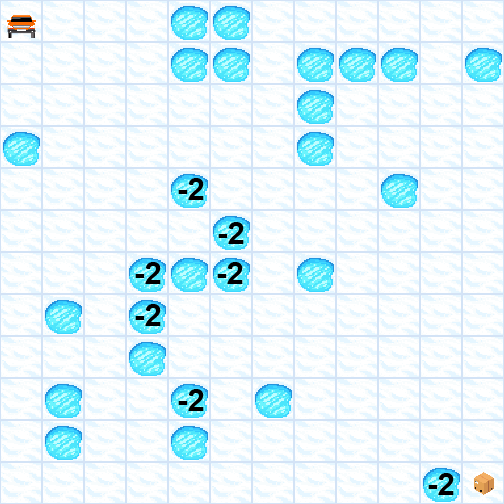}
    \end{subfigure}
    \caption{Advice from the \textit{bottom-left} corner in the parallel cooperative human experiments, with $5\%$ quota, and its visualization. The counterpart of the cooperative advice is shown in \figref{fig:advice-file-coop-par-5-TR}.}
    \label{fig:advice-file-coop-par-5-BL}
\end{figure}

\end{appendices}

\clearpage
\bibliography{bib/references}

\end{document}